\UseRawInputEncoding
\documentclass[11pt]{article}

\usepackage[final]{acl}

\usepackage{times}
\usepackage{latexsym}
\usepackage[T1]{fontenc}
\usepackage[utf8]{inputenc}
\usepackage{microtype}
\usepackage{inconsolata}
\usepackage{graphicx}
\usepackage{cuted}
\usepackage{listings}
\usepackage{mdframed}
\usepackage[dvipsnames,table]{xcolor}
\usepackage{bm}
\usepackage{amssymb}

\usepackage{subcaption}
\usepackage{booktabs}
\usepackage{rotating}
\usepackage{fancyvrb} 
\usepackage{amsmath}
\usepackage{amssymb}
\usepackage{mathtools}
\usepackage{amsthm}

\usepackage{tcolorbox}
\tcbuselibrary{breakable} % Essential for long appendices!
\usepackage{enumitem}  % for nosep option
\usepackage{array}
\usepackage{tabularx}
\usepackage{multirow}
\usepackage{makecell}
\usepackage{adjustbox}
\usepackage{colortbl}
\usepackage{placeins}
\usepackage{lipsum}

\usepackage{pifont}              % for \ding
\usepackage{fontawesome}         % icons

\usepackage[capitalize,noabbrev]{cleveref}

\newcommand{\worst}[1]{%
  {\setlength{\fboxsep}{0.6pt}\colorbox{red!25}{\hspace{1pt}#1\hspace{1pt}}}%
}
\newcommand{\cmark}{\textcolor{ForestGreen}{\ding{51}}}
\newcommand{\xmark}{\textcolor{BrickRed}{\ding{55}}}
\newcommand{\high}[1]{\textbf{#1}}

\theoremstyle{plain}

\theoremstyle{definition}

\theoremstyle{remark}

\title{SONIC-O1: A Real-World Benchmark for Evaluating Multimodal Large Language Models on Audio-Video Understanding}

\author{
  \textbf{Ahmed Y. Radwan\textsuperscript{1}},
  \textbf{Christos Emmanouilidis\textsuperscript{2}},
  \textbf{Hina Tabassum\textsuperscript{3}},
\\
  \textbf{Deval Pandya\textsuperscript{1}},
  \textbf{Shaina Raza\textsuperscript{1,*}}
\\
\\
  \textsuperscript{1}Vector Institute for Artificial Intelligence, MaRS Centre, Toronto, ON M5G 1L7, Canada \\
  \textsuperscript{2}University of Groningen, Nijenborgh 4, 9747 AG Groningen, Netherlands \\
  \textsuperscript{3}York University, 4700 Keele Street, Toronto, ON M3J 1P3, Canada
\\
\small{
  \texttt{\{ahmed.radwan, shaina.raza\}@vectorinstitute.ai}
  }
\\
\small{
  \textsuperscript{*}\textbf{Correspondence}
}
}

\begin{document}
\maketitle
\begin{abstract}
Multimodal Large Language Models (MLLMs) are a major focus of recent AI research. However, most prior work focuses on static image understanding, while their ability to process audio–video data remains underexplored. This gap highlights the need for a high-quality benchmark to systematically evaluate MLLM performance in a real-world setting. We introduce \textbf{SONIC-O1}, a comprehensive, fully human-verified benchmark of $\approx$60 hours (231 clips) spanning 13 real-world conversational domains with 4,958 annotations and perceived demographic metadata. SONIC-O1 evaluates three capabilities: open-ended summarization, multiple-choice question (MCQ) answering, and temporal localization with supporting rationales (reasoning). Across closed- and open-source models, we find that the MCQ accuracy shows the smallest gap between model families, under their recommended settings, the best closed-source model outperforms the best open-source model by \textbf{22.6\%} on temporal localization. We further observe accuracy gaps of up to 21.4\% on temporal localization across demographic groups, indicating diagnostic disparities in model behaviour. SONIC-O1 provides an open evaluation suite for temporally grounded and demographically informed multimodal understanding. SONIC-O1 is publicly available for research: 

\noindent
\href{https://vectorinstitute.github.io/sonic-o1/}{\faGlobe~Project page} \quad
\href{https://huggingface.co/datasets/vector-institute/sonic-o1}{\raisebox{-0.2ex}{\includegraphics[height=2ex]{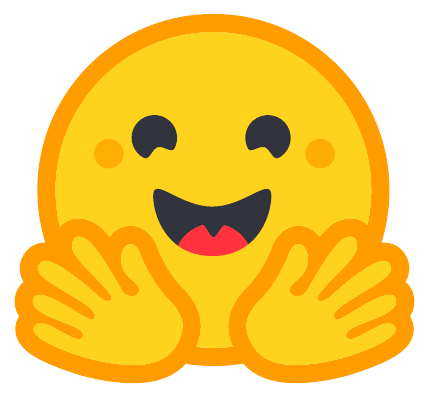}}~Dataset} \quad
\href{https://github.com/vectorinstitute/sonic-o1}{\raisebox{-0.2ex}{\includegraphics[height=2ex]{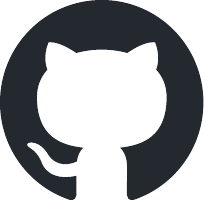}}~Github} \quad
\href{https://huggingface.co/spaces/vector-institute/sonic-o1-leaderboard}{\faTrophy~Leaderboard}

%UNCOMMENT it during preprint, should be after title  \begin{center}
% \begin{tabular}{cll}
% % \webpage & \textbf{Project:} & {\small\url{https://vectorinstitute.github.io/SONIC-O1/}} \\ <--- Ahmed this one uncomment during preprint
% \huggingface & \textbf{Data:} & {\small\href{https://huggingface.co/datasets/vector-institute/SONIC-O1}{\nolinkurl{https://huggingface.co/datasets/vector-institute/SONIC-O1}}} \\
% \github & \textbf{Code:} & {\small\url{https://github.com/VectorInstitute/SONIC-O1}} \\
% \end{tabular}
% {\small We fully comply with the double-blind review process. All code and data are anonymized in the submission.}
% \end{center}

\end{abstract}

\section{Introduction}
MLLMs have advanced from static image captioning to general-purpose perception-and-reasoning systems capable of processing video and audio inputs~\citep{fu2024mme}. These models are increasingly deployed to assist real-world decisions through conversational interactions. As MLLMs move into high-stakes such as healthcare, education, and public safety capability alone is insufficient. We must also evaluate whether these systems behave \textit{accurately}, \textit{fairly across demographics}, and \textit{transparently} across diverse users and situations.

\begin{figure}[t]
    \centering
    \includegraphics[width=1.02\linewidth]{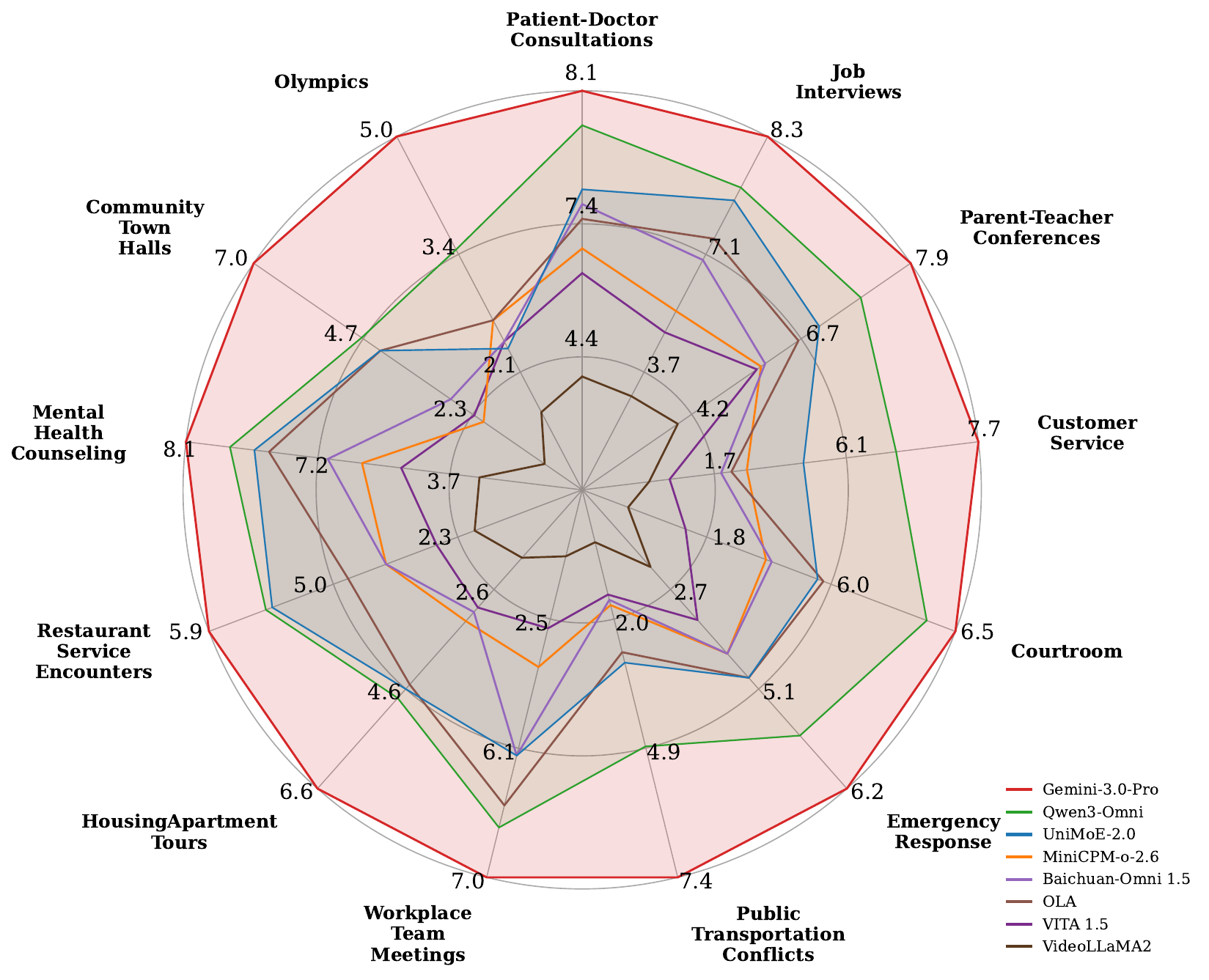}
    \caption{\textbf{Performance comparison across 13 conversational domains}. We compare closed-source and open-source MLLMs across 13 conversational domains using LLM-judge scores (0–10) for video summarization task. Gemini 3.0 Pro consistently outperforms open-source models, and high-stakes domains (e.g., Emergency Response, Mental Health) remain more challenging.}
       \vspace{-1em}
    \label{fig:topic_performance}
\end{figure}

\begin{figure*}
    \centering
    \includegraphics[width=0.9\textwidth]{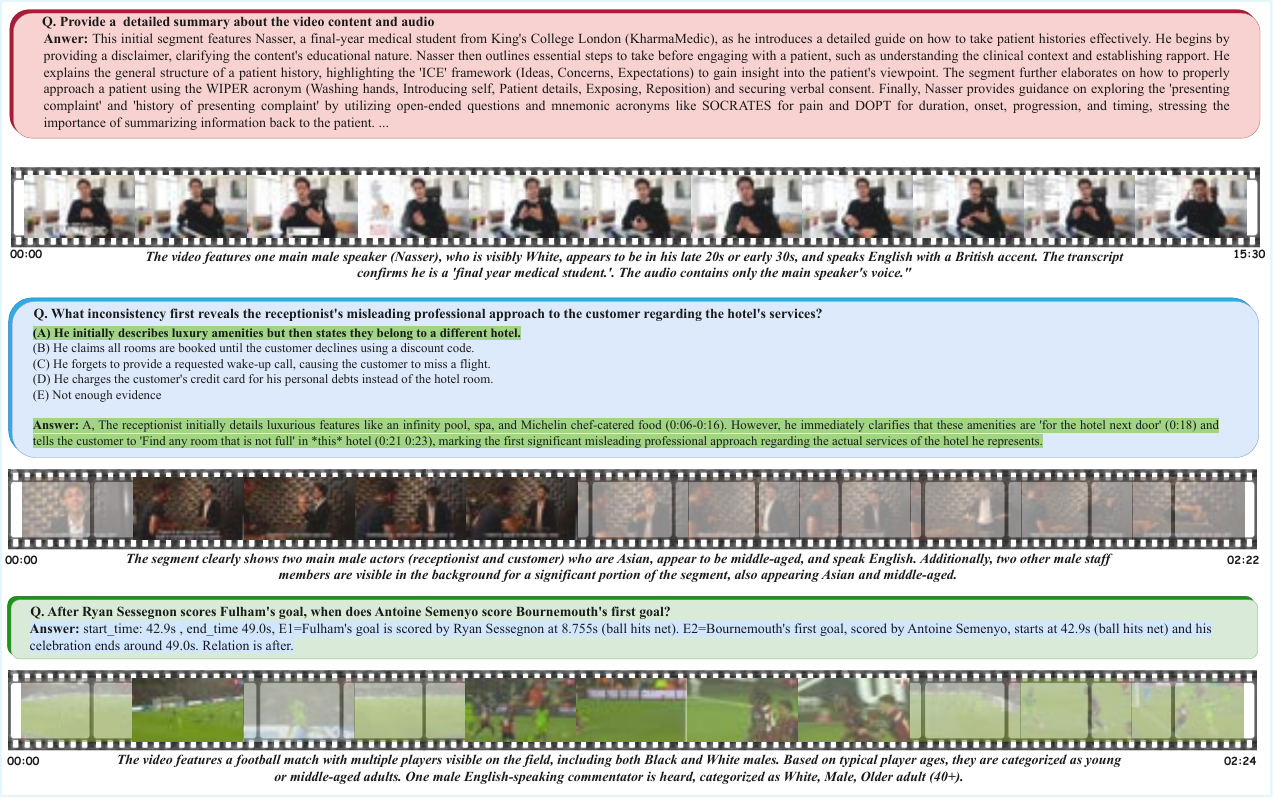}
    \caption{Overview of SONIC-O1 tasks and evaluation format: \textbf{(Top)} video summarization, \textbf{(Middle)} evidence-grounded MCQ, \textbf{(Bottom)} temporal localization (event timing). Each example shows the input clip (frames) and the expected output format; demographic attributes shown beneath each clip represent associated metadata, enabling group-wise evaluation across 13 domains.
    }
    \vspace{-1em}
    \label{fig:teaser}
\end{figure*}

Real-world interactions rely fundamentally on both audio and video modalities. For example, speech conveys affect, emphasis, hesitation, and social intent that are essential for understanding communication. However, existing audio-video benchmarks exhibit two critical gaps. First, audio is frequently treated as optional or replaced with transcripts and subtitles~\citep{ataallah2024infinibench,wang2025lvbench}, under-exploring paralinguistic cues that shape the interpretation of speaker meaning and their emotional state. Second, even when native audio-video inputs are evaluated, group-wise analysis across demographic groups remains largely absent~\citep{chen2024cg,li2025omnivideobench}, precluding assessment of whether model performance varies systematically with user characteristics. Together, these limitations obscure whether current MLLMs can operate accurately and fairly in real-world deployment scenarios. A comparison of seminal benchmarks is given in Table~\ref{tab:videoqa-coverage}.

\definecolor{rowhighlight}{RGB}{214,220,255} % light purple/blue highlight
% Define a centered version of the X column type
\newcolumntype{C}{>{\centering\arraybackslash}X}
% Define a centered fixed-width column if needed
\newcolumntype{P}[1]{>{\centering\arraybackslash}p{#1}}

% ---------- Table ----------
\begin{table*}[t]
\scriptsize
\centering
\caption{Comparison of video-QA benchmarks. SONIC-O1 uniquely combines comprehensive temporal understanding, audio-visual reasoning, social cue analysis, and open-source availability with both automatic and human annotations across diverse video lengths. \textbf{Legend:} \cmark{} = Yes, \xmark{} = No, $\delta$ = Partial temporal, 
AVQA = Audio-Visual Question Answering, MCQs = Multiple Choice Questions.}
\label{tab:videoqa-coverage}

% Adjust column separation to ensure it fits perfectly without resizebox
\setlength{\tabcolsep}{3pt} 
\renewcommand{\arraystretch}{0.8}  
\begin{tabularx}{\textwidth}{@{} l c P{1.8cm} P{1.5cm} *{8}{C} @{}}
\toprule
\textbf{Benchmark} & \textbf{Data Size} & \textbf{Video Lengths} & \textbf{QA Type} & \textbf{Summ.} & \textbf{Temp.}\textsuperscript{†} & \textbf{Reas.}\textsuperscript{‡} & \textbf{AVQA} & \textbf{Social}\textsuperscript{*} & \multicolumn{2}{c}{\textbf{Annotation}} & \textbf{Open} \\
\cmidrule(lr){10-11}
& & \textbf{(min)} & & & & & & \textbf{Cues} & \textbf{Auto} & \textbf{Human} & \textbf{Src} \\
\midrule

\multicolumn{12}{c}{\textit{Video-QA Benchmarks (AVQA: \xmark)}} \\
\midrule
LongVideoBench~\cite{wu2024longvideobench} & 6,678 & $\sim$8 & MCQs & \xmark & $\delta$ & \cmark & \xmark & \xmark & \xmark & \cmark & \cmark \\
LVBench~\cite{wang2025lvbench} & 1,549 & $\sim$68 & MCQs & \xmark & $\delta$ & \cmark & \xmark & \xmark & \xmark & \cmark & \xmark \\
CinePile~\cite{rawal2024cinepile} & 305K & $\sim$3 & MCQs & \xmark & $\delta$ & \xmark & \xmark & \xmark & \cmark & \cmark & \xmark \\
Sports-QA~\cite{li2024sports} & 94,000 & $<$1 & Open-ended & \xmark & $\delta$ & \cmark & \xmark & \xmark & \cmark & \cmark & \cmark \\
InfiniBench~\cite{ataallah2024infinibench} & 87,700 & $\sim$53 & MCQs + Open & \cmark & $\delta$ & \cmark & \xmark & \xmark & \cmark & \cmark & \cmark \\

\midrule
\multicolumn{12}{c}{\textit{Audio-Visual Benchmarks (AVQA: \cmark)}} \\
\midrule
VAST~\cite{chen2023vast} & 27M & $<$1 & Captioning & \xmark & \xmark & \xmark & \cmark & \xmark & \cmark & \xmark & \xmark \\
Daily-Omni~\cite{zhou2025daily} & 1,197 & $<$1 & MCQs & \xmark & $\delta$ & \cmark & \cmark & \xmark & \cmark & \xmark & \cmark \\
WorldSense~\cite{hong2025worldsense} & 1,662 & $\sim$2.4 & MCQs & \xmark & \xmark & \cmark & \cmark & \xmark & \xmark & \cmark & \cmark \\
LongVALE~\cite{geng2025longvale} & 8,411 & $\sim$3.9 & Captioning & \xmark & \cmark & \xmark & \cmark & \xmark & \cmark & \cmark & \cmark \\
Video-MME~\cite{fu2025video} & 2,700 & $<$2; 4--15; 30--60 & MCQs & \xmark & $\delta$ & \xmark & \cmark & \xmark & \xmark & \cmark & \cmark \\
MAVERIX~\cite{xie2025maverix} & 2,556 & $\sim$6 & MCQs + Open & \xmark & $\delta$ & \cmark & \cmark & \xmark & \xmark & \cmark & \xmark \\
CG-Bench~\cite{chen2024cg} & 12,129 & 10--60 & MCQs + Open & \xmark & \cmark & \cmark & \cmark & \xmark & \xmark & \cmark & \cmark \\
OmniVideoBench~\cite{li2025omnivideobench} & 1,000 & 1--30 & MCQs + Open & \cmark & $\delta$ & \cmark & \cmark & \xmark & \xmark & \cmark & \cmark \\
AURA~\cite{galougah2025aura} & 1,600 & $<$1 & MCQs & \xmark & $\delta$ & \cmark & \cmark & \xmark & \cmark & \xmark & \cmark \\

\midrule
\rowcolor{blue!10} \multicolumn{1}{>{\columncolor{blue!10}[-3pt][3pt]}l}{\textbf{SONIC-O1}} & 4,958 & $<$5; 5--20; 20--60 & MCQs + Open & \cmark & \cmark & \cmark & \cmark & \cmark & \cmark & \cmark & \multicolumn{1}{>{\columncolor{blue!10}[3pt][-3pt]}c}{\cmark} \\
\bottomrule
\end{tabularx}

\vspace{0.01em}
\begin{flushleft}
\scriptsize
\textsuperscript{*}Social Cues: Demographic metadata (race, gender, age) annotated from observable characteristics in videos via AI-assisted human review (see Appendix~\ref{app:ai-assisted}), enabling fairness evaluation across demographic groups.
\textsuperscript{†}Temporal: Temporal localization questions requiring localizing start and end of an event. 
\textsuperscript{‡}Reasoning: Open-ended explanation of the given answer from the model.
\end{flushleft}
\end{table*}

To this end, we introduce \textbf{SONIC-O1}, a \textbf{SO}cial \textbf{N}atural \textbf{I}nteraction \textbf{C}orpus (\textbf{O}mnimodal, v\textbf{1}),  fully open-source, human-verified benchmark combining native audio-video input, open-ended summarization, fine-grained MCQ reasoning, and IoU-grounded temporal localization with demographic metadata across \textbf{4,958} audio-video question-answer (QA) instances derived from $\approx$60 hours of real-world audio-video interactions. As shown in Figure~\ref{fig:sonic_sunburst}, the dataset spans 13 high-impact topics across 5 domains (including legal/civic, educational, and public health), covering real-world scenarios. SONIC-O1 covers videos that range from 30 seconds to 60 minutes (covering short, medium, and long durations) and is tagged with metadata covering 6 racial groups, genders, and age groups.

SONIC-O1 evaluates state-of-the-art MLLMs on three tasks (as shown in Figure~\ref{fig:teaser}): \textbf{(i) summarization} testing global comprehension of the full conversation; \textbf{(ii) multiple-choice QA} testing fine-grained reasoning, and \textbf{(iii) temporal localization} testing whether models can identify when events occur. Unlike prior work that treats audio as optional (as shown in Table~\ref{tab:videoqa-coverage}), SONIC-O1 requires native omnimodal reasoning: models must jointly process audio, video, and text to produce responses.. Our goal with this work is to reveal where omnimodal MLLMs succeed, fail, and diverge across demographic groups in conversational settings.

\textbf{Contributions.}
(1) We introduce SONIC-O1, an open-source benchmark with human-verified domain-expert annotations and demographic metadata for evaluating MLLM performance and group-wise analysis on real-world interactions.
(2) We benchmark omnimodal MLLMs, revealing substantial performance gaps and diagnostic disparities across demographic groups.
(3) We release the full evaluation suite (dataset, scripts, leaderboard) under a research license. \\
Our evaluation reveals several key findings. (1) Gemini 3.0 Pro outperforms the evaluated open-source alternatives, (2) temporal localization remains the most challenging task, and (3) diagnostic demographic disparities are observed across models.

\section{Related Work}
\textbf{Growth in MLLMs.}
MLLMs have rapidly evolved from static image understanding~\citep{radford2021learning,li2022blip} 
to video comprehension~\citep{sun2019videobert,tong2022videomae}, with recent vision-language models~\citep{Qwen-VL,li2024llava,Maaz2023VideoChatGPT} extending these capabilities to video but treating audio as auxiliary rather than integral to multimodal reasoning. A new generation of \textit{omnimodal} MLLMs, including VITA 1.5~\citep{fu2024vita}, Qwen3-Omni~\citep{qwen3omni}, 
Gemini 3.0 Pro~\citep{gemini3pro}, MiniCPM-o-2.6~\citep{minicpmo26}, Baichuan-Omni 1.5~\citep{li2025baichuan}, OLA~\citep{liu2025ola}, and UniMoE-2.0~\citep{unimoe2}, where they jointly process audio, video, and text within a unified framework for integrated reasoning across 
modalities~\citep{jiang2025specific}. 
However, evaluation benchmarks have not kept pace:  existing datasets often omit audio entirely, replace it with text transcripts, or lack the demographic annotations necessary to assess group-wise performance across diverse user populations.

\textbf{Video Benchmarks.} Recent benchmarks have begun to evaluate temporal reasoning across diverse video lengths, including short-form tasks such as MVBench~\citep{li2025omnivideobench} and TempCompass~\citep{liu2024tempcompass}, as well as long-form understanding in Video-MME~\citep{fu2025video} (up to one hour) and LongVideoBench~\citep{wu2024longvideobench}. However, two critical gaps remain. First, \textbf{audio is underutilized}. Even when available, evaluations often default to frame- or subtitle-based setups due to limited model support. Second, \textbf{group-wise analysis is largely absent}, preventing systematic measurement of performance disparities across demographic groups. As shown in Table~\ref{tab:videoqa-coverage}, SONIC-O1 addresses these limitations by combining native audio-video evaluation, temporal localization, and demographic metadata to enable group-wise analysis assessment.

\section{SONIC-O1: Dataset and Benchmark Design}
\label{sec:sonic}
In this section, we describe the design of the SONIC-O1 dataset and benchmark. 
\subsection{Data Collection}
\label{subsec:construction}
\begin{figure}[t]
    \centering
    \includegraphics[width=0.9\columnwidth]{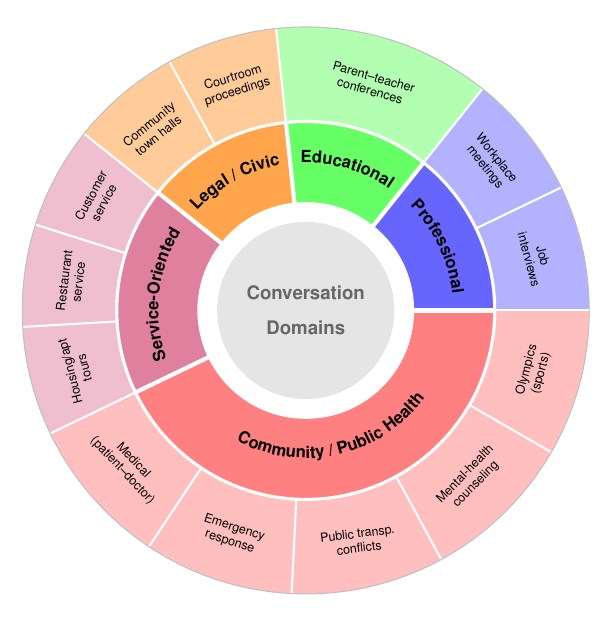}
    \vspace{-0.5em}
    \caption{Video categories. Our benchmark covers 5 key domains and 13 video topics.}
    \label{fig:sonic_sunburst}
    \vspace{-0.8em}
\end{figure}

The construction of SONIC-O1 is grounded in real-world data.
% that follows 3 stages: (1) data collection, (2)  annotation, and (3) quality review. In this section, we discuss the data collection.
To ensure broad coverage, we collected videos from YouTube~\citep{google_youtube_data_api_v3} restricted to CC BY 4.0 licenses. We target five high-stakes domains, such as~\textit{Professional, Educational, Legal/Civic, Service-Oriented, and Public Health}, subdivided into 13 topics (e.g., job interviews, courtroom proceedings) as shown in Figure~\ref{fig:sonic_sunburst}. To benchmark long-context capabilities, we stratified sampling across three duration ranges: short ($<$5~minutes), medium (5-20~minutes), and long (20-60~minutes). Our search strategy combined topic keywords with demographic descriptors (race, gender, age) to maximize representation.
From an initial pool of 2,237 candidates, 1,794 met licensing constraints, and subsequent manual quality filtering yielded 231 final videos totaling approximately 60 hours. SONIC-O1 provides fully human-verified, real-world audio-video content in sensitive domains with perceived demographic annotations, enabling fairness-oriented slice analysis that is typically not reported in movie/TV-based benchmarks~\citep{song2024moviechat,wang2025lvbench} or in many judge-based / synthetically-evaluated evaluation suites~\citep{fang2024mmbench,zhou2025mlvu}. Full details are provided in Appendix~\ref{app:collection}.

\begin{figure}[t]
    \centering
    \begin{subfigure}[c]{\linewidth}
        \centering
        \includegraphics[width=0.9\textwidth]{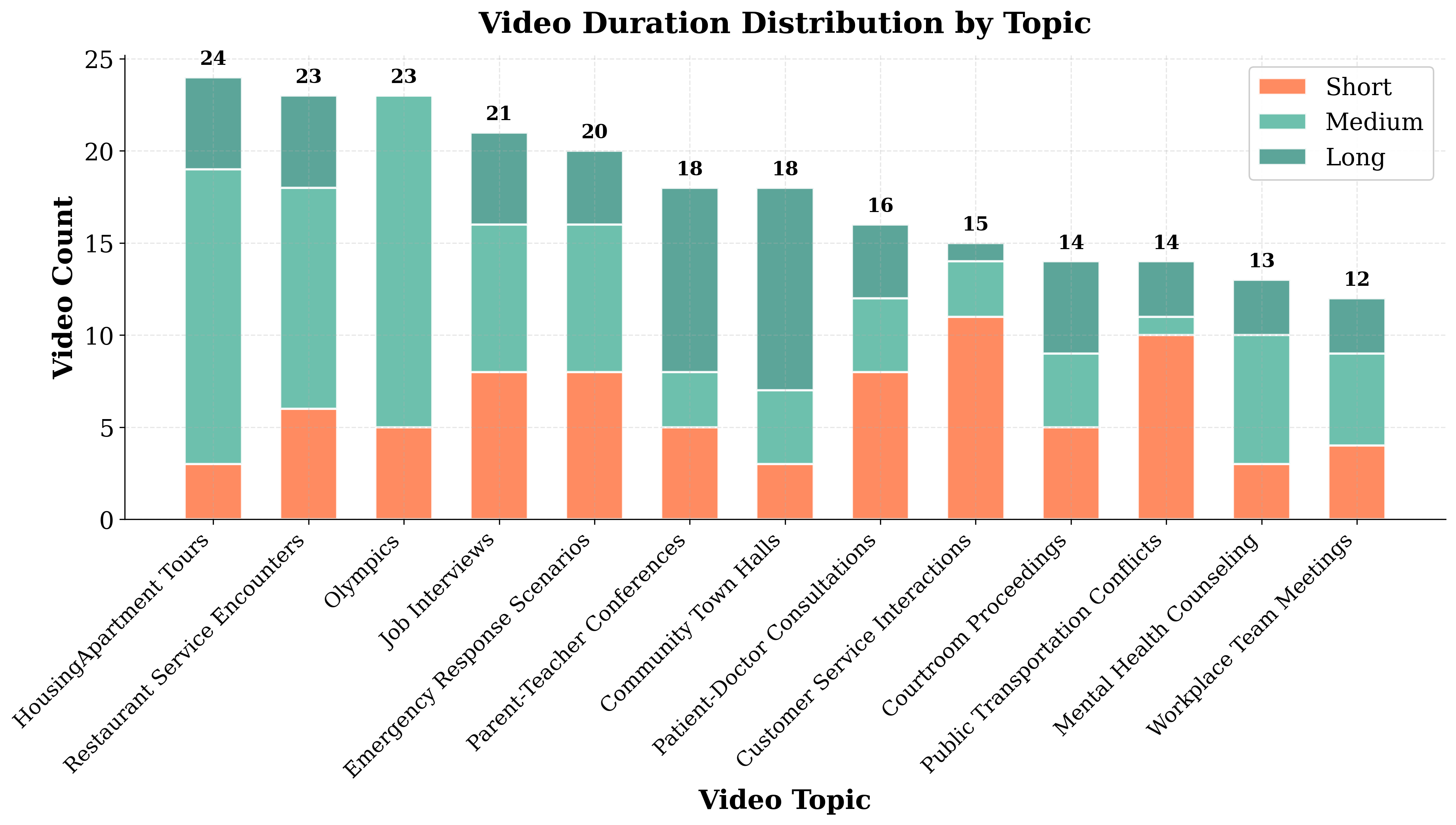}
        \vspace{-0.5em}
        \caption{Video duration distribution over topics}
        \label{fig:sonic_duration}
    \end{subfigure}
    \vspace{0.6em}
    \begin{subfigure}[c]{\linewidth}
        \centering
        \includegraphics[width=0.9\textwidth]{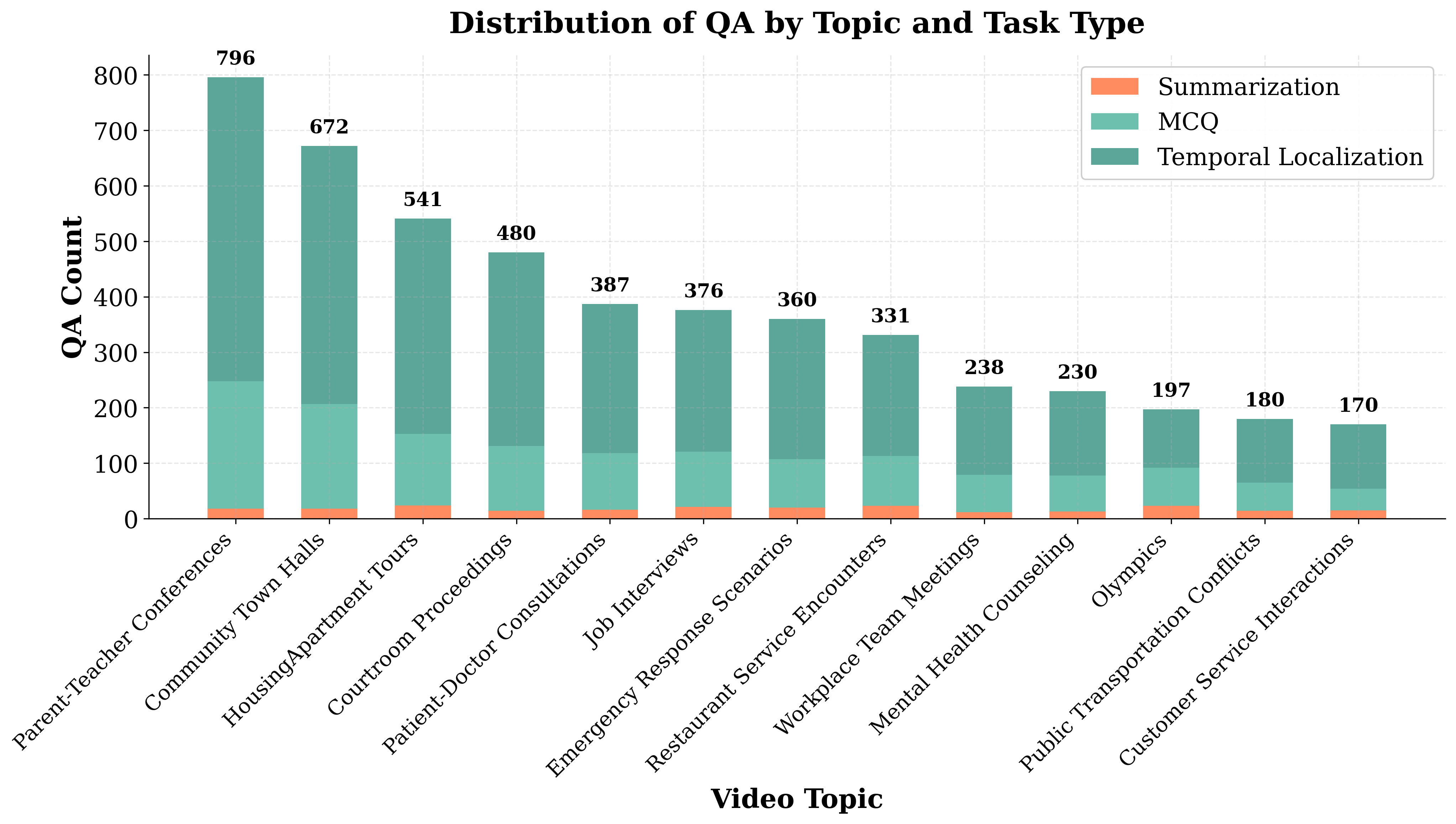}
        \vspace{-0.5em}
        \caption{Question type distribution over topics}
        \label{fig:sonic_qtype}
    \end{subfigure}
    \vspace{-0.6em}
    \caption{Video duration and question type distributions. SONIC-O1 spans a full spectrum of video lengths and evaluates core abilities of MLLMs.}
    \label{fig:sonic_distributions}
    \vspace{-0.8em}
\end{figure}

% Task 1 operate on complete video
% Task 2-3 Segments
% Task 1 Annotation when video is long (gemini oom) we segment 10 minutes, summarize each, then merge (appendix explains) <-- which appendix??  Appendix-\label{app:task-procedures} 

\subsection{Tasks Annotation}
% Following the collection of 231 videos,
We define three evaluation tasks over the same video pool and discuss them below. Figure~\ref{fig:sonic_qtype} shows the annotation counts per task across the 13 topics; Appendix~\ref{app:ai-assisted} describes the AI-assisted annotation workflow and prompts.
% : \textbf{(Task 1) video summarization}, which tests global comprehension; \textbf{(Task 2) multiple-choice question answering (MCQ)}, which evaluates fine-grained reasoning; and \textbf{(Task 3) temporal localization}, which assesses temporal grounding, i.e., whether models can identify not only \textit{what} happens but also \textit{when}. 
% % Task 1 operates on full videos, whereas Tasks 2 and 3 operate on overlapping segments extracted from longer videos to enable fine-grained annotation and evaluation. Task 1 is open-ended, while Tasks 2 and 3 are closed-ended tasks involving reasoning.
% All videos and segments include perceived demographic annotations (race, gender, age), enabling  group-wise evaluation (fairness) of MLLM performance.
% We outline each task and its annotation process below. Figure~\ref{fig:sonic_qtype} shows the annotation counts per task across the 13 topics; Appendix~\ref{app:ai-assisted} describes the AI-assisted annotation workflow and prompts.

\begin{table}[h]
\centering
\footnotesize
\caption{SONIC-O1 task statistics.}
\label{tab:stats_questionsNumbers}
\setlength{\tabcolsep}{2pt}
\renewcommand{\arraystretch}{1.1}  

\begin{tabular}{l c p{4.5cm}}
\toprule
\textbf{Task} & \textbf{\#Inst.} & \textbf{Unit / Ground Truth} \\
\midrule
Summarization & 231 & Full video $\rightarrow$ reference summary \\
MCQ & 1{,}335 & 3-minutes overlap video segment $\rightarrow$ answer + reasoning \\
Temporal loc. & 3{,}392 & 3-minutes overlap video segment $\rightarrow$ relation + $(t_s,t_e)$ + reasoning \\
\midrule
\textbf{Total} & \textbf{4{,}958} & 231 videos ($\approx$60 hours) \\
\bottomrule
\end{tabular}
\vspace{-0.6em}
\end{table}

\textit{Task 1: Video Summarization.}  This task requires models to summarize videos by capturing key events, actions, and outcomes. Following established video summarization benchmarks~\citep{ning2025video}, models generate a detailed narrative summary up to 300 words. During annotation, to preserve coverage for long videos, we apply a hierarchical procedure, where videos longer than 10 minutes are segmented into 10-minute chunks to context limits, each chunk is summarized, and the chunk-level summaries are merged into a final reference summary. The task contains 231 summarization instances, each consisting of an audio-video input paired with a human-verified reference summary.

\textit{Task 2: MCQs with Reasoning.} This task evaluates fine-grained understanding of local video segments using a controlled answer space, a common design for objective comparison in video benchmarks~\citep{ye2024mm,lei2018tvqa,patraucean2023perception}. We operate on segments of up to 3 minutes; longer videos are split into 3-minute windows with 30-second overlap. For each segment, we construct an MCQ with five choices, consisting of four candidate answers plus a \textit{Not enough evidence} option. During evaluation, MLLMs must choose the correct answer and provide a rationale grounded in visual and auditory evidence.

\begin{table*}[t]
\centering
\small
\caption{\textbf{Performance of MLLMs on \textsc{SONIC-O1}} across three tasks: 
summarization (LLM-judge score, ROUGE-L, cosine similarity), MCQ (accuracy and 
rationale quality), and temporal localization (mIoU, R@0.5, rationale quality). 
Percentages: Accuracy, mIoU, R@0.5, ROUGE-L; cosine similarity in [0,1]. All 
metrics are macro-averaged across topics. FPV: frames per video; FPS: frames per 
second. Higher is better. \textbf{Bold}: best; \underline{underline}: second-best 
per metric. $^\dagger$: closed-source. Full results in Appendix~\ref{tab:comprehensive_results}.}
\vspace{-0.5em}
\renewcommand{\arraystretch}{1}
\setlength{\tabcolsep}{5pt}
\resizebox{\textwidth}{!}{%
\begin{tabular}{l|c|ccc|cccc|ccccc}
\toprule
\multirow{2}{*}{Model} & \multirow{2}{*}{\makecell{\textbf{LLM}\\\textbf{Params}}} &
\multicolumn{3}{c|}{\textbf{Summarization}} &
\multicolumn{4}{c|}{\textbf{MCQ}} &
\multicolumn{5}{c}{\textbf{Temporal Localization}} \\
\cmidrule(lr){3-5}\cmidrule(lr){6-9}\cmidrule(lr){10-14}
& & Score & ROUGE-L & Sim
& Acc. & Score & ROUGE-L & Sim
& mIoU & R@0.5 & Score & ROUGE-L & Sim \\
\midrule
Gemini 3.0 Pro$^\dagger$ (1 FPS)    & -   & \textbf{7.07} & \textbf{27.2} & \textbf{0.81} & \textbf{81.4} & \textbf{8.71} & \underline{19.6} & \textbf{0.71} & \textbf{26.6} & \textbf{25.4} & \textbf{5.38} & \textbf{28.3} & \underline{0.67} \\
Qwen3-Omni (256 FPV)                & 30B & \underline{5.72} & \underline{22.8} & \underline{0.71} & \underline{65.7} & \underline{7.47} & \textbf{20.3} & \underline{0.70} & \underline{3.7} & \underline{2.8} & 2.58 & \underline{28.0} & \textbf{0.70} \\
UniMoE-2.0 (256 FPV)                & 33B & 4.71 & 20.8 & 0.70 & 51.6 & 6.17 & 13.1 & 0.63 & 1.8 & 1.0 & 2.11 & 23.7 & 0.55 \\
MiniCPM-o-2.6 (256 FPV)             & 9B  & 3.34 & 14.7 & 0.56 & 51.9 & 6.15 & 10.3 & 0.60 & 1.8 & 0.7 & 3.65 & 19.3 & 0.41 \\
Baichuan-Omni 1.5 (32 FPV)         & 7B   & 3.68 & 18.8 & 0.60 & 56.2 & 6.18 & 15.1 & 0.65 & 2.8 & 1.1 & 2.29 & 18.2 & 0.43 \\
OLA (256 FPV)                       & 7B   & 4.42 & 19.0 & 0.64 & 53.2 & 6.61 & 14.0 & 0.64 & 3.3 & 1.2 & 3.48 & 21.8 & 0.44 \\
VITA 1.5 (64 FPV)                   & 8B  & 2.77 & 16.3 & 0.49 & 51.1 & 6.04 & 13.2 & 0.63 & 1.8 & 1.2 & \underline{3.91} & 22.1 & 0.43 \\
VideoLLaMA 2 (128 FPV)              & 7B  & 1.53 & 12.7 & 0.38 & 26.4 & 4.17 & 9.3  & 0.53 & 3.5 & 0.4 & 2.22 & 22.5 & 0.49 \\
\bottomrule
\end{tabular}}
\label{tab:main_scoreboard_final}
\end{table*}

\textit{Task 3: Temporal Localization.}
This task evaluates temporal reasoning by grounding events with timestamps (e.g., whether a diagnosis occurs before or after symptoms). Temporal localization is standard in audio-video benchmarks for fine-grained event alignment~\citep{zhou2025daily,geng2025longvale}. We extend this by requiring models to predict timestamps and provide rationales grounded in audio-visual evidence, although we treat it as a stress test for fine-grained temproal grounding.
We use an anchor-target formulation that given an anchor event, models must (1) predict the target's start and end times, and (2) provide a reasoning with salient evidence. Domain experts label times relative to each segment's start (0.0s), which are converted to absolute timestamps by adding the segment's offset. The dataset contains 3{,}392 instances (up to three per segment), each with target timestamps and supporting reasoning. 

% it is to exaplin relative and absolute time, task 1 and task 2 don't have this, in task 3 the AI annotation output relative timestamp (0-180s) we convert to absolute by adding the start_duration of segment
% why
% because when first prompt I asked for absolute time, gemini failed, so I did this way to make it easier (But why do we need to convert it in absolute?) Because 1- we chose segment 3 minues design of choice, but an extension could remove this segment and pass complete video-audio (our QA will work because it's in absolute) 2- In real-time scenarios when a video been recording for 2-8 hours ( the questions is 'at what hour this ... happened?' so it is to be realistic should return the absolute actual time not relative to a segment processing, because we chose '3 minutes segments' because current models are barely managing 256 frames, this was to make it easier for them, but in real-life usage it would be [imagine football game, camera recording the 2-3 hours, and you ask when this goal happened or when did this happen, it should return the time in 1:30:26 this action happened (absolute time, but maybe the preprocessing/processing did segment the video to fit context length)] 

% OK keep such points for rebutal here in commented areas
% human-verified by domain experts - added you can check later
%in CONTRIBUTIONS IN INTRO, WE NEED TO MENTION DOMAIN EXPERTS or something

\subsection{Quality Assurance}
Each evaluation task is manually reviewed and verified by domain experts; team details and guidelines are provided in Appendix~\ref{app:team_formation}. Experts watched each full video before labeling, with the ability to revisit any timestamp as needed. To support scalable annotation, we used Gemini 2.5 Flash to draft perceived demographic metadata (race, gender, age) and task annotations across all tasks, these drafts are not used directly, and all final annotations were human-verified, inspired by prior benchmark efforts~\citep{ataallah2024infinibench}. Where there attributes are not self-identified and are used only for aggregate group-wise analysis. Full prompts are given in Appendix~\ref{app:ai-generation-prompts}. All outputs were then verified, corrected, and revised via our internal review interface (Appendix~\ref{app:interface}). We removed any items with ambiguity, insufficient evidence, or annotation errors and resolved disagreements through consensus adjudication. This process results in the construction of SONIC-O1, with dataset statistics reported in Table~\ref{tab:stats_questionsNumbers} and further details provided in Appendix~\ref{app:dataset-viz}.
%We do not report inter-annotator agreement since annotations were finalized through expert review rather than independent parallel labeling. SONIC-O1 data statistics are given in Table~\ref{tab:stats_questionsNumbers}, with details in Appendix~\ref{app:dataset-viz}.

\subsection{Evaluation Suite}
We release an extensible evaluation suite offering a unified framework for models operating on direct audio-video inputs, with consistent segmenting and sampling policies for long videos. 
This benchmark is designed to scale as new MLLMs, interaction formats, and task variants emerge, modularizing the pipeline into data loading, inference, and metric computation across heterogeneous commercial and open-source API formats, lowering the barrier to adoption and supporting reproducible evaluation and future community extensions.

\section{Experiments}
\label{sec:experiments}
% In this section, we define our experimental setup.
\subsection{Settings}
\label{sec:exp_models}
\textbf{Evaluated model suite}
We conduct the evaluation on commercial and open-source MLLMs. For commercial models, we used Gemini-3.0-Pro~\citep{gemini3pro} and GPT-4o~\citep{gpt4o} (ablation only). Representative open-source omni MLLMs including Qwen3-Omni-30B-A3B~\citep{qwen3omni}, Uni-MoE-2.0-Omni~\citep{unimoe2}, MiniCPM-o-2.6~\citep{minicpmo26}, Baichuan-Omni 1.5~\citep{li2025baichuan}, OLA~\citep{liu2025ola} VITA-1.5~\citep{fu2024vita}, and VideoLLaMA2~\citep{cheng2024videollama} are evaluated as well. We prioritize native audio-video models, including GPT-4o, solely for modality ablation. Open-source MLLMs are evaluated using their custom, optimal configurations (up to 256 frames). To handle context limitations, we apply adaptive visual fallback and custom audio preprocessing. Full inference details appear in Table~\ref{tab:model_settings} and Appendix~\ref{app:preprocess}.

\textbf{Metrics}
We employ metrics, following evaluation protocols as in related works\citep{fang2024mmbench,ataallah2024infinibench}.  For summarization, we report ROUGE-L, cosine similarity, and an LLM judge score (GPT-5-mini with 0--10 scale). For MCQs, we report accuracy (\%);  we additionally report the LLM judge score on open-ended reasoning for this task. For temporal localization, we report mean intersection over union (mIoU) and recall at intersection over union (R@0.5). Full definitions and prompts are in Appendix~\ref{app:appendix_metrics}.

\subsection{Results}
\paragraph{Overall Performance.}
Table~\ref{tab:main_scoreboard_final} reports comprehensive evaluation results across all three tasks. Overall, the closed-source Gemini 3.0 Pro achieves the highest performance among all evaluated models, obtaining 81.4\% MCQ accuracy, a 7.07 judge score on summarization, and 25.4\% R@0.5 on temporal localization. Notably, a large performance gap of ~23\% exists between Gemini 3.0 Pro and the best open-source model on temporal localization, highlighting the difficulty of this task for current open-source systems.

Among open-source models, Qwen3-Omni demonstrates the strongest overall performance, achieving 65.7\% MCQ accuracy and a 5.72 judge score on summarization. Interestingly, model scale does not strictly determine closed-ended reasoning ability: Baichuan-Omni 1.5 and OLA (both 7B) attain 56.2\% and 53.2\% MCQ accuracy respectively, outperforming the larger UniMoE-2.0 (33B, 51.6\%). However, scale remains important for generation and grounding tasks. Smaller models show noticeable drops compared to Qwen3-Omni (30B): MiniCPM-o-2.6 and VideoLLaMA2 exhibit up to 41.6\% lower summarization judge scores, and 6.7$\times$ lower R@0.5 on temporal localization.\\
\textit{Key Finding:} Gemini 3.0 Pro remains ahead of the evaluated open-source alternatives. Temporal localization is the most challenging task for current MLLMs, while summarization and MCQ are comparatively easier. Larger open-source models provide measurable gains but do not fully close the performance gap with closed-source systems.

%Based on the results in Table~\ref{tab:main_scoreboard_final}, we report below the most informative metrics for each task, with additional analyses provided in the Appendix~\ref{app:detailed_results}.

\begin{table}[t]
\centering
\caption{\textbf{Performance of MLLMs on summarization task across demographic groups.} Results are on all dataset reported using LLM judge Score (0--10), higher scores indicate better performance. \textbf{Bold}: best, \worst{highlighted}: worst within each group per column. Detailed results in Appendix Table~\ref{tab:demo_summ_a}}
\label{tab:demo_summ}
\vspace{-0.5em}
\setlength{\tabcolsep}{2pt}
\resizebox{\columnwidth}{!}{%
\begin{tabular}{@{}lccccccccc@{}}
\toprule
Group &
\makecell{Gemini 3.0\\Pro$^\dagger$} &
\makecell{Qwen3\\Omni} &
\makecell{UniMoE\\2.0} &
\makecell{MiniCPM\\o-2.6} &
\makecell{Baichuan\\Omni 1.5} &
\makecell{OLA} &
\makecell{VITA\\1.5} &
\makecell{Video\\LLaMA2} \\
\midrule
\multicolumn{9}{l}{\textit{Race}}\\
Arab       & 6.90 & \high{5.95} & \high{5.00} & 3.57 & \high{4.29} & \high{4.76} & \high{2.76} & \worst{1.00} \\
Indigenous & 6.70 & \worst{4.13} & 4.35 & \high{3.61} & 3.70 & 4.39 & \worst{1.65} & 1.04 \\
Asian      & \high{7.05} & 5.71 & 4.62 & 3.26 & 3.61 & 4.29 & 2.65 & \high{1.63} \\
White      & 6.68 & 5.28 & 4.29 & 3.26 & 3.22 & 4.27 & 2.50 & 1.45 \\
Hispanic   & 6.41 & 4.99 & 3.70 & 3.04 & \worst{2.44} & 3.62 & 2.21 & 1.23 \\
Black      & \worst{6.02} & 4.39 & \worst{3.45} & \worst{2.92} & 2.55 & \worst{3.60} & 2.31 & 1.38 \\
\midrule
\multicolumn{9}{l}{\textit{Gender}}\\
Male   & \worst{6.36} & \worst{4.97} & \worst{3.93} & \worst{2.98} & \worst{2.94} & \worst{4.00} & \worst{2.31} & \worst{1.36} \\
Female & \high{7.02} & \high{5.47} & \high{4.55} & \high{3.53} & \high{3.47} & \high{4.29} & \high{2.65} & \high{1.53} \\
\midrule
\multicolumn{9}{l}{\textit{Age}}\\
18--24 & \worst{6.28} & \worst{4.93} & 4.02 & \high{3.30} & \worst{3.04} & 4.12 & \worst{2.41} & 1.38 \\
25--39 & 6.52 & 5.01 & \worst{4.01} & \worst{3.14} & 3.11 & \worst{3.96} & 2.45 & \high{1.50} \\
40+    & \high{6.91} & \high{5.47} & \high{4.45} & 3.21 & \high{3.25} & \high{4.29} & \high{2.47} & \worst{1.36} \\
\bottomrule
\end{tabular}}
\end{table}

\begin{table}[t]
\centering
\caption{\textbf{MCQ performance across demographic groups.} Accuracy (\%), with higher values indicating better performance. Detailed results in Appendix Tables~\ref{tab:demo_mcq_a},~\ref{tab:demo_mcq_b}.}
\label{tab:demo_mcq}
\vspace{-0.5em}
\setlength{\tabcolsep}{2pt}
\resizebox{\columnwidth}{!}{%
\begin{tabular}{@{}lccccccccc@{}}
\toprule
Group &
\makecell{Gemini 3.0\\Pro$^\dagger$} &
\makecell{Qwen3\\Omni} &
\makecell{UniMoE\\2.0} &
\makecell{MiniCPM\\o-2.6} &
\makecell{Baichuan\\Omni 1.5} &
\makecell{OLA} &
\makecell{VITA\\1.5} &
\makecell{Video\\LLaMA2} \\
\midrule
\multicolumn{9}{l}{\textit{Race}}\\
Arab       & \high{82.7} & 63.9 & 48.7 & 50.3 & 58.6 & 51.3 & 47.1 & 20.9 \\
Indigenous & 80.0 & \high{68.6} & \worst{42.9} & 48.6 & \high{62.9} & \worst{31.4} & \worst{42.4} & \worst{8.6} \\
Asian      & 80.3 & \worst{61.8} & \high{53.6} & 51.0 & 55.7 & \high{53.4} & \high{49.5} & 23.2 \\
White      & 81.2 & 66.5 & 51.0 & \high{51.2} & 55.8 & 52.7 & 49.0 & \high{26.7} \\
Hispanic   & \worst{79.1} & {63.5} & 48.5 & \worst{48.5} & \worst{48.2} & 46.5 & 48.5 & 22.6 \\
Black      & 79.2 & {63.5} & 48.9 & 49.5 & 52.2 & 48.3 & 48.1 & 24.6 \\
\midrule
\multicolumn{9}{l}{\textit{Gender}}\\
Male   & \high{81.4} & \high{65.1} & \high{51.4} & \worst{49.7} & \high{55.7} & \high{51.6} & \high{49.1} & \high{25.1} \\
Female & \worst{79.4} & \worst{64.5} & \worst{49.2} & \high{51.5} & \worst{53.2} & \worst{50.3} & \worst{48.2} & \worst{24.6} \\
\midrule
\multicolumn{9}{l}{\textit{Age}}\\
18--24 & 80.9 & \worst{59.5} & 49.3 & \high{50.7} & 55.4 & \worst{46.3} & \high{49.5} & \worst{23.8} \\
25--39 & \worst{79.5} & 63.5 & \worst{48.7} & \worst{50.2} & \worst{51.8} & 49.2 & \worst{48.6} & \high{25.4} \\
40+    & \high{81.4} & \high{67.3} & \high{52.4} & \high{50.7} & \high{57.0} & \high{53.7} & 48.7 & 24.7 \\
\bottomrule
\end{tabular}}
\end{table}

\begin{table}[h]
\centering
\caption{\textbf{Temporal localization performance across demographic groups.} Results are reported as R@0.5 (\%), where higher values indicate better temporal grounding. Detailed results in Appendix Tables~\ref{tab:demo_temporal_a}, \ref{tab:demo_temporal_b}.}
\label{tab:demo_temporal}
\small
\vspace{-0.5em}
\setlength{\tabcolsep}{2pt}
\resizebox{\columnwidth}{!}{%
\begin{tabular}{@{}lccccccccc@{}}
\toprule
Group &
\makecell{Gemini 3.0\\Pro$^\dagger$} &
\makecell{Qwen3\\Omni} &
\makecell{UniMoE\\2.0} &
\makecell{MiniCPM\\o-2.6} &
\makecell{Baichuan\\Omni 1.5} &
\makecell{OLA} &
\makecell{VITA\\1.5} &
\makecell{Video\\LLaMA2} \\
\midrule
\multicolumn{9}{l}{\textit{Race}}\\
Arab       & 21.1 & 1.6 & 0.2 & \high{2.6} & \worst{0.3} & 1.4 & 1.2 & \worst{0.0} \\
Indigenous & \high{40.9} & \worst{0.0} & \high{1.3} & \worst{0.0} & \high{5.8} & \high{9.2} & 1.3 & \high{1.3} \\
Asian      & 30.7 & \high{2.9} & 0.6 & 0.8 & 1.6 & 1.3 & \high{1.4} & 0.4 \\
White      & 23.0 & 2.6 & 1.2 & 0.9 & 1.3 & \worst{1.2} & \high{1.4} & 0.5 \\
Hispanic   & 23.8 & 2.3 & \worst{0.1} & 0.2 & 0.8 & 1.4 & \worst{0.8} & \worst{0.0} \\
Black      & \worst{19.5} & 1.8 & 0.6 & 0.3 & 0.8 & 1.7 & \high{1.4} & 0.3 \\
\midrule
\multicolumn{9}{l}{\textit{Gender}}\\
Male   & \worst{23.5} & \high{2.6} & \high{0.9} & \worst{0.7} & \worst{1.1} & \worst{1.3} & \worst{1.2} & 0.4 \\
Female & \high{24.2} & \worst{2.1} & \worst{0.7} & \high{0.9} & \high{1.4} & \high{1.5} & \high{1.4} & 0.4 \\
\midrule
\multicolumn{9}{l}{\textit{Age}}\\
18--24 & \high{27.3} & 2.4 & \high{1.2} & \high{1.1} & \high{1.8} & \high{2.7} & \worst{1.1} & \worst{0.3} \\
25--39 & 25.2 & \high{2.5} & \high{1.2} & \worst{0.7} & 1.3 & 1.4 & 1.2 & \worst{0.3} \\
40+    & \worst{21.9} & \worst{2.3} & \worst{0.4} & 0.8 & \worst{0.9} & \worst{1.2} & \high{1.5} & \high{0.4} \\
\bottomrule
\end{tabular}}
\end{table}
\begin{table*}[t]
\scriptsize
\caption{\textbf{Duration robustness of MLLMs across video lengths.} Performance 
reported on all dataset, macro-averaged across topics, for short, medium, and long 
videos across three tasks: summarization (LLM-judge score), MCQ (accuracy, \%), 
and temporal localization (R@0.5, \%). Higher is better. \textbf{Bold}: best; 
\underline{underline}: second-best per duration bin and task.}
\centering
\scriptsize
\resizebox{\linewidth}{!}{%
\begin{tabular}{l|c|ccc|ccc|ccc}
\toprule
\multirow{2}{*}{Model} & \multirow{2}{*}{\makecell{\textbf{LLM}\\\textbf{Params}}} &
\multicolumn{3}{c|}{\textbf{Summarization (Score)}} &
\multicolumn{3}{c|}{\textbf{MCQ (Accuracy)}} &
\multicolumn{3}{c}{\textbf{Temporal Localization (R@0.5)}} \\
\cmidrule(lr){3-5}\cmidrule(lr){6-8}\cmidrule(lr){9-11}
& & Short & Medium & Long & Short & Medium & Long & Short & Medium & Long \\
\midrule
Gemini 3.0 Pro       & -   & \textbf{8.16} & \textbf{6.11} & \textbf{6.63} & \textbf{81.9} & \textbf{82.8} & \textbf{79.5} & \textbf{50.6} & \textbf{26.0} & \textbf{18.6} \\
Qwen3-Omni           & 30B & \underline{6.64} & \underline{4.95} & \underline{4.87} & \underline{64.9} & \underline{67.1} & \underline{65.8} & \underline{9.2} & \underline{2.6} & \underline{2.5} \\
UniMoE-2.0           & 33B & 5.49 & 4.37 & 3.87 & 55.3 & 50.2 & 50.9 & 6.9 & 0.3 & 0.6 \\
MiniCPM-o-2.6        & 9B  & 4.08 & 2.87 & 2.87 & 59.6 & 50.9 & 51.9 & 0.9 & 1.1 & 0.6 \\
Baichuan-Omni 1.5    & 7B   & 4.70 & 3.06 & 2.94 & 59.6 & 55.3 & 56.3 & 1.9 & 1.4 & 1.1 \\
OLA                  & 7B   & 4.63 & 4.23 & 4.23 & 52.1 & 54.1 & 53.6 & 2.9 & 1.6 & 0.8 \\
VITA 1.5             & 8B  & 3.39 & 2.40 & 2.46 & 53.1 & 50.3 & 47.8 & 3.0 & 1.2 & 1.0 \\
VideoLLaMA 2         & 7B  & 1.89 & 1.46 & 1.27 & 34.0 & 26.3 & 25.4 & 0.9 & 0.3 & 0.4 \\
\bottomrule
\end{tabular}}
\label{tab:duration_bins}
\end{table*}

\paragraph{Performance across demographic groups.}
We report MLLM performance stratified by race, gender, and age in Tables~\ref{tab:demo_summ}, \ref{tab:demo_mcq}, and \ref{tab:demo_temporal} to quantify demographic performance disparities. 
Overall, we observe that the closed-source Gemini 3.0 Pro achieves the strongest performance across nearly all demographic groups, while Qwen3-Omni is the most competitive open-source model. MCQ accuracy is relatively stable across groups, whereas summarization scores show larger variation. The largest gaps emerge in temporal localization (R@0.5), particularly across racial groups (e.g., Indigenous: 40.9\% vs.\ Black: 19.5\%).

Race shows the most pronounced disparities, with smaller gender effects and generally higher performance for participants aged 40+. Female participants achieve slightly higher scores, except for MCQs; one possible contributor is that safety alignment can shift response tone and refusal tendencies, which may interact with our scoring. Smaller models appear less robust across demographic slices, and Mann-Whitney U tests confirm these disparities are statistically significant (Appendix~\ref{app:mann_whitney}), with the Asian--Black summarization gap persists within topics for 5/8 models, although topic, duration, accent, background noise, and video quality remain uncontrolled.

\textit{Key Finding:} Group-wise demographic analysis reveals that temporal localization shows substantially larger disparities across race and age than aggregate results suggest, while MCQ remains stable across groups.

\begin{table}[h]
\centering
\scriptsize
\caption{\textbf{Effect of input modalities} (video-only vs.\ video+audio) on 
representative MLLMs and topics. Summarization: LLM-judge score; MCQ: accuracy 
(\%) and rationale score; temporal localization: R@0.5 (\%) and rationale score. 
\textbf{Bold}: best within each model.}
\setlength{\tabcolsep}{2pt}
\resizebox{\columnwidth}{!}{%
\begin{tabular}{lccccc}
\toprule
Model &
\makecell[c]{Video} &
\makecell[c]{Audio} &
\makecell[c]{Sum. \\(Score)} &
\makecell[c]{MCQ\\(Acc./Score)} &
\makecell[c]{Temporal\\(R@0.5/Score)} \\
\midrule
\multirow{2}{*}{Qwen3-Omni}
  & \cmark & \xmark & 4.46 & 61.6 / 6.67 & 1.6 / \textbf{2.55} \\
  & \cmark & \cmark & \textbf{6.84} & \textbf{71.9} / \textbf{8.06} & \textbf{4.3} / 2.70 \\
\midrule
\multirow{2}{*}{UniMoE-2.0}
  & \cmark & \xmark & 3.77 & 46.4 / 5.97 & 0.7 / \textbf{2.42} \\
  & \cmark & \cmark & \textbf{5.65} & \textbf{58.3} / \textbf{6.95} & \textbf{1.2} / 2.07 \\
\midrule
\multirow{2}{*}{VideoLLaMA2}
  & \cmark & \xmark & \textbf{1.91} & 28.6 / \textbf{4.96} & \textbf{0.9} / \textbf{2.89} \\
  & \cmark & \cmark & 1.78 & \textbf{28.6} / 4.39 & 0.6 / 2.12 \\
\midrule
\multirow{2}{*}{GPT4o}
  & \cmark & \xmark & 4.97 & 65.7 / 7.49 & \textbf{4.3} / 2.89 \\
  & \cmark & \cmark & \textbf{5.50} & \textbf{74.5} / \textbf{8.47} & 3.1 / \textbf{2.94} \\
\bottomrule
\vspace{0.001mm}
\end{tabular}
}
\textbf{Note:} For GPT4o, the "Audio" column corresponds to \emph{Text} input (i.e., GPT4o uses Video+Text instead of Video+Audio).
\label{tab:modality_ablation}
\end{table}

\paragraph{Performance across topics.}
We also observe consistent topic-level variation in generation. As can be seen in Figure~\ref{fig:topic_performance}, Gemini 3.0 Pro performs best across nearly all domains, Qwen3-Omni and UniMoE-2.0 (30B) form a competitive middle tier, while smaller 7-9B models drop sharply here. 
High-stakes scenarios (e.g., emergency response, mental health) are most challenging to solve by MLLMs, whereas some topics in professional settings (e.g., consultations, interviews) are comparatively easier (Appendix~\ref{app:per-topic-detailed}). \
\textit{Key Finding:} Robustness remains uneven across topics, with the largest gaps in emotionally sensitive contexts.

\paragraph{Performance across short, medium, and long durations.} Table~\ref{tab:duration_bins} reports MLLM performance stratified by video duration (short $<5$~minutes, medium 5--20~minutes, long 20--60~minutes) to assess the impact of temporal extent. Overall, we observe that Gemini 3.0 Pro consistently achieves the highest performance across all video lengths (short, medium, long) on all three tasks. Smaller models like VITA 1.5, VideoLLaMA2 shows limited robustness to longer and more complex videos. We also observe that performance degrades as video length increases, with temporal localization most affected, while MCQ remains relatively stable. \\
\textit{Key Finding:} Performance declines on longer videos, with temporal localization most affected by the challenge of sustained event tracking.

\begin{figure*}[t]
\centering
\begin{subfigure}[b]{0.3\textwidth}
    \centering
    \includegraphics[width=\textwidth]{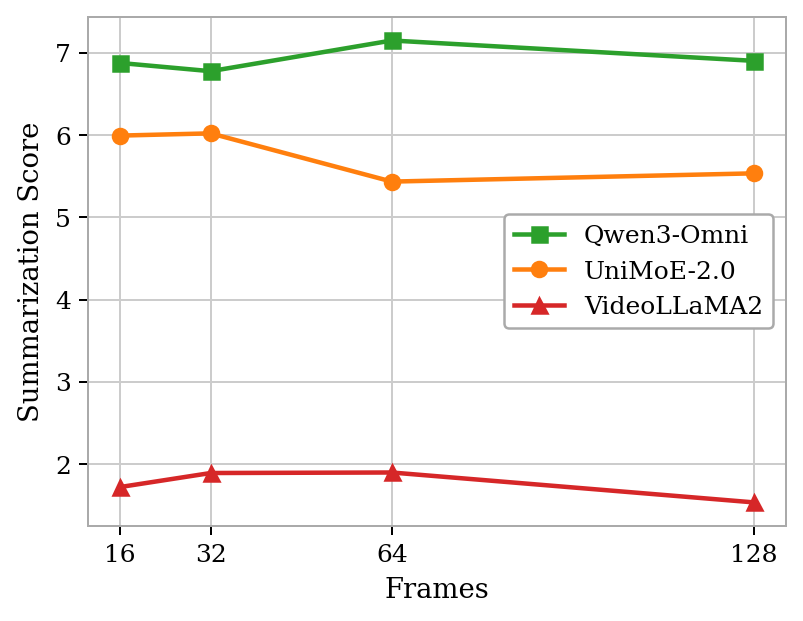}
    \caption{Summarization Score}
    \label{fig:summarization}
\end{subfigure}
\hfill
\begin{subfigure}[b]{0.3\textwidth}
    \centering
    \includegraphics[width=\textwidth]{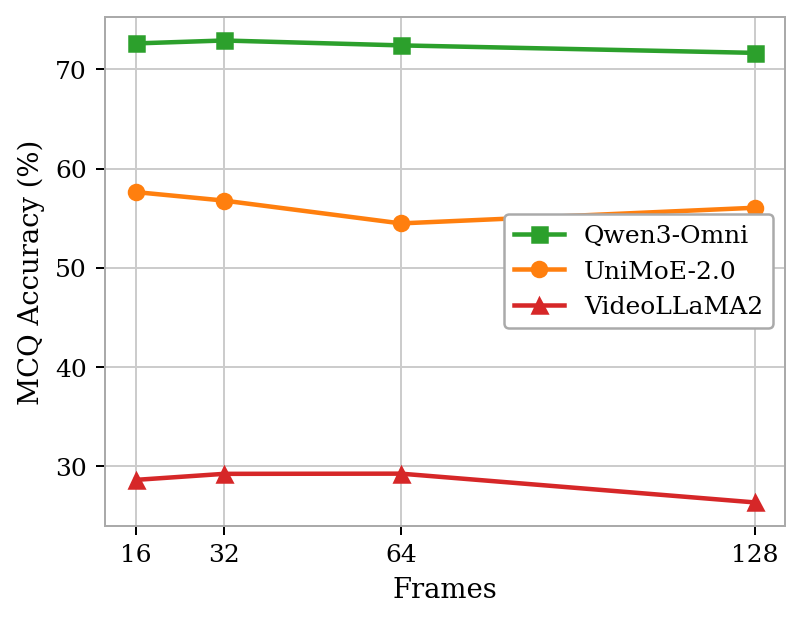}
    \caption{MCQ Accuracy}
    \label{fig:mcq}
\end{subfigure}
\hfill
\begin{subfigure}[b]{0.3\textwidth}
    \centering
    \includegraphics[width=\textwidth]{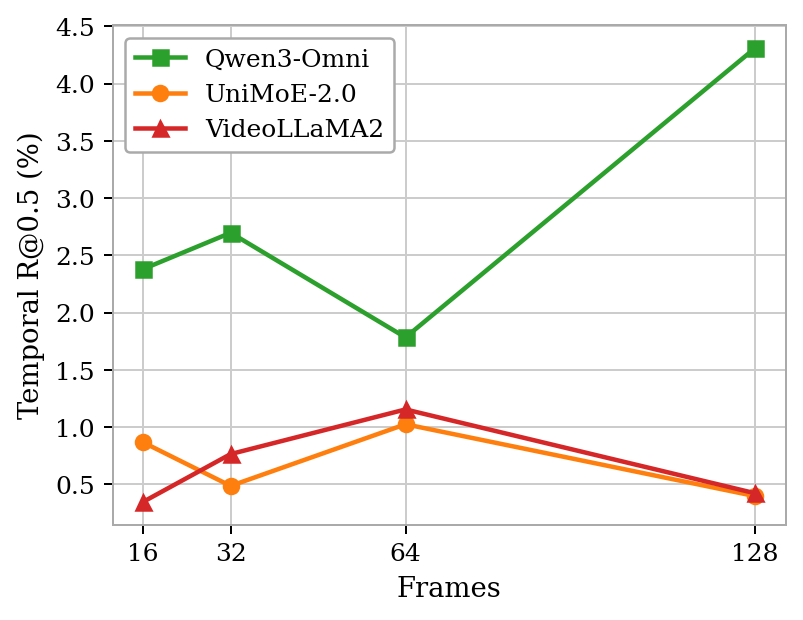}
    \caption{Temporal R@0.5}
    \label{fig:temporal}
\end{subfigure}
\caption{\textbf{Frame-count sensitivity across metrics.} Three MLLMs 
(Qwen3-Omni, UniMoE-2.0, VideoLLaMA2) evaluated at 16, 32, 64, and 128 frames 
across: (a) summarization (LLM-judge score), (b) MCQ accuracy, and (c) temporal 
localization (R@0.5). Models exhibit different saturation patterns across tasks.}
\label{fig:frame_sensitivity}
\end{figure*}

\subsection{Ablation studies}
The impact of input modality and frame sampling rate is examined through ablation studies on three representative topics, including patient–doctor consultations (T1), job interviews (T2), and courtroom proceedings (T5).

\paragraph{Effect of input modalities.}
\label{sec:modality}
Table~\ref{tab:modality_ablation} presents a modality ablation study comparing video-only to combined video+audio input. Adding audio consistently improves performance, with Qwen3-Omni benefiting most (+10.3\% MCQ accuracy, +2.38 summarization score, +2.7\% R@0.5). GPT-4o similarly gains 8.8\% MCQ accuracy when text transcripts are added. However, gains vary by model: smaller models show weaker improvements, and VideoLLaMA2 even exhibits slight summarization degradation ($-$0.13) despite no MCQ change.\\
\textit{Key Finding:} Multimodal inputs improve performance across tasks, with larger models benefiting more consistently than smaller ones, suggesting effective audio-visual fusion requires sufficient model capacity.

\paragraph{Effect of frame sampling rate.}
MLLMs typically operate over a limited set of uniformly sampled frames (e.g., 16 or 32), raising the question of whether denser visual coverage improves performance. Figure~\ref{fig:frame_sensitivity} shows that increasing the number of input frames does not consistently yield gains. For summarization and MCQ (Figures~\ref{fig:summarization} and \ref{fig:mcq}), accuracy remains largely unchanged, suggesting that additional frames provide limited benefit for high-level semantic understanding. In contrast, temporal localization (Figure~\ref{fig:temporal}) improves more noticeably with higher frame counts for stronger models (e.g., Qwen3-Omni), while weaker models (e.g., VideoLLaMA2) exhibit inconsistent trends. \\
\textit{Key Finding:} Higher frames primarily benefit temporal localization, but yield little gain for summarization or MCQ.

\subsection{Qualitative Analysis}
\label{sec:empathy}
% LIWC: see backward recent citations if any strong LLMs doing, also is it not a problem that strong model results are evaluated with less pwoerful tool but as u said its linguistic so shud be ok 
We analyze generated empathic summaries using LIWC-22~\citep{tausczik2010psychological,boyd2022development}, a validated dictionary-based tool (details in Appendix~\ref{app:app_empathy}), examining total emotional expression, negative emotion, and overall tone. Results in Table~\ref{tab:liwc_compact} reveal distinct default profiles: Gemini 3.0 Pro adopts a neutral, clinical style with higher emotional acknowledgment but lower tone scores, while Qwen3-Omni and UniMoE-2.0 exhibit a warmer, more supportive profile. Baichuan-Omni 1.5 achieves the highest tone (60.09), followed by OLA (60.00), both reflecting their shared Qwen2.5-7B backbone. Smaller models (e.g., MiniCPM-o-2.6, VideoLLaMA2) show limited emotional modulation, indicating reduced affective capacity.\\
\textit{Key Finding:} MLLMs demonstrate distinct baseline emotional language styles, with larger models showing richer affective expression reflecting default tendencies rather than controlled modulation.
 
\begin{table}[t]
\centering
\caption{\textbf{Emotional language analysis} of MLLM empathic summaries on all 
dataset, macro-averaged across topics (Task~1). Total Emotion (\%): words 
expressing any emotion; Neg. Emotion (\%): negative affect words; Tone: overall 
warmth and positivity (higher = warmer). \textbf{Bold}: best; \worst{highlighted}: 
worst.}
\label{tab:liwc_compact}
\small
\setlength{\tabcolsep}{3pt}
\resizebox{\columnwidth}{!}{%
\begin{tabular}{lccc}
\toprule
Model & Total Emotion (\%) & Neg. Emotion (\%) & Tone \\
\midrule
Gemini 3.0 Pro & \textbf{4.35} & \textbf{2.03} & {46.16} \\
Qwen3-Omni & 3.88 & 1.39 & 57.99 \\
UniMoE-2.0 & 3.54 & 0.93 & 57.94 \\
MiniCPM-o-2.6 & \worst{1.41} & \worst{0.38} & \worst{46.10} \\
Baichuan-Omni 1.5 & 3.47 & 0.82 & \textbf{60.09} \\
OLA & 4.02 & 1.15 & {60.00} \\
VITA-1.5 & 2.65 & 0.59 & 55.45 \\
VideoLLaMA2 & {2.02} & {0.49} & 49.83 \\
\bottomrule
\end{tabular}
}
\end{table}

%\section{Discussion}
%Our results reveal a clear gap in current MLLMs. They excel at content understanding but struggle with temporal grounding. Video inputs offer limited gains, and temporal localization remains especially challenging for open-source models. Frequent hallucinated time references point to deeper weaknesses in internal temporal reasoning.

\section{Conclusion}
We introduced \textbf{SONIC-O1}, a real-world audio-video benchmark for evaluating MLLMs through the lens of diagnostic perceived demographic fairness evaluation, covering three tasks: summarization, MCQ, and temporal localization. Our results show that Gemini 3.0 Pro mostly outperform open-source alternatives, with temporal localization being the most challenging task. Source videos are public YouTube content and may appear in pretraining corpora, though low temporal scores across all models are not consistent with memorization.
% , where open-source models lag behind Gemini 3.0 Pro by 22.6\% R@0.5.
MCQ task shows competitive accuracy, however, we observe that some open-source models exhibit systematic temporal reference frame hallucination. Demographic analysis further reveals diagnostic performance disparities, with gaps of up to 21.4\% R@0.5 on temporal localization across racial groups, while MCQ remains stable, underscoring that demographic performance evaluation must be task-specific. We hope SONIC-O1 guides future work on stronger temporal reasoning and more equitable multimodal  evaluation.

% --- ACL required: after Conclusion, before References, not counted in page limit ---
\section*{Limitations}
We acknowledge some limitations of this study. SONIC-O1 focuses on English-language interactions, which may limit cultural and linguistic coverage. Although annotations are human-verified, some subjectivity remains, particularly for open-ended tasks. The benchmark is diagnostic rather than population-representative, is restricted in scale and task scope, and may not fully generalize to other domains, modalities, or future model generations.

From an \textbf{evaluation perspective}, some constraints are observed. First, many MLLMs rely on similar Qwen-based backbones but we observe that they differ substantially in audio processing. In practice, long audio streams are often truncated to fixed durations rather than adaptively aligned with sampled video frames. Second, models frequently struggle with absolute temporal localization. When clips do not begin at zero (e.g., 500--800s), predictions are often reported relative to the segment start (e.g., 50--80s instead of 550--580s), suggesting reliance on relative playback position rather than robust temporal representations (see error taxonomy in Appendix~\ref{app:temporal_error_analysis}). Third, demographic groups have unequal sample sizes (Figure~\ref{fig:qa_demographics}), which may increase variance for underrepresented groups.
Fourth, all models are evaluated with a single run, so marginal differences (e.g., 51.6 vs.\ 51.9 MCQ) should not be over-interpreted.
Finally, context-length limitations require video segmentation for most models, which can further amplify temporal errors. In contrast, Gemini's larger context window allows processing full videos at 1 FPS, reducing the need for segmentation. For MCQ and temporal localization tasks, however, videos are segmented into 3-minute clips during evaluation, at which point Gemini receives at most 180 frames per segment, at or below the 256-frame budget allocated to open-source models. The performance gap on these tasks therefore cannot be attributed to Gemini's context advantage.
% \href{https://arxiv.org/pdf/2512.02425}{WorldMM}

% --- ACL optional: broader impacts, after Limitations, not counted in page limit ---
\section*{Ethical Considerations}
SONIC-O1 is designed to support the responsible development of audio–video models. It is intended to help researchers, practitioners, and organizations better understand how these systems behave, especially in sensitive areas such as media analysis, education, and public information. Like other public benchmarks, the results could be taken out of context or selectively used in adversarial, reputational, or legal situations if the limitations are ignored. While SONIC-O1 examines gaps affecting marginalized demographic groups, this analysis is meant to identify and reduce potential harms, not to reinforce stereotypes, offend communities, or be used against any group.

\section*{Acknowledgments}
Resources used in preparing this research were provided, in part, by the Province of Ontario and the Government of Canada through CIFAR, as well as companies sponsoring the Vector Institute (\url{http://www.vectorinstitute.ai/\#partners}).
This research was funded by the European Union's Horizon Europe research and innovation programme under the AIXPERT project (Grant Agreement No.\ 101214389), which aims to develop an agentic, multi-layered, GenAI-powered framework for creating explainable, accountable, and transparent AI systems.

% ACL style sets the bibliography style automatically
\bibliography{3-references}

@manual{google_youtube_data_api_v3,
  title        = {YouTube Data API v3},
  author       = {{Google Developers}},
  organization = {Google},
  year         = {2025},
  note         = {Accessed: 2025-11-10},
  url          = {https://developers.google.com/youtube/v3}
}

@article{fu2024mme,
  title={Mme-survey: A comprehensive survey on evaluation of multimodal llms},
  author={Fu, Chaoyou and Zhang, Yi-Fan and Yin, Shukang and Li, Bo and Fang, Xinyu and Zhao, Sirui and Duan, Haodong and Sun, Xing and Liu, Ziwei and Wang, Liang and others},
  journal={arXiv preprint arXiv:2411.15296},
  year={2024}
}

@article{fang2024mmbench,
  title={Mmbench-video: A long-form multi-shot benchmark for holistic video understanding},
  author={Fang, Xinyu and Mao, Kangrui and Duan, Haodong and Zhao, Xiangyu and Li, Yining and Lin, Dahua and Chen, Kai},
  journal={Advances in Neural Information Processing Systems},
  volume={37},
  pages={89098--89124},
  year={2024}
  }

@inproceedings{geng2025longvale,
  author={Geng, Tiantian and Zhang, Jinrui and Wang, Qingni and Wang, Teng and Duan, Jinming and Zheng, Feng},
  booktitle={2025 IEEE/CVF Conference on Computer Vision and Pattern Recognition (CVPR)}, 
  title={LongVALE: Vision-Audio-Language-Event Benchmark Towards Time-Aware Omni-Modal Perception of Long Videos}, 
  year={2025},
  volume={},
  number={},
  pages={18959-18969},
  doi={10.1109/CVPR52734.2025.01766}
  }

@inproceedings{ye2024mm,
  title={MMEgo: Towards building egocentric multimodal LLMs for video QA},
  author={Ye, Hanrong and Zhang, Haotian and Daxberger, Erik and Chen, Lin and Lin, Zongyu and Li, Yanghao and Zhang, Bowen and You, Haoxuan and Xu, Dan and Gan, Zhe and others},
  booktitle={International Conference on Learning Representations},
  volume={2025},
  pages={71705--71723},
  year={2025}
}

@inproceedings{lei2018tvqa,
  title={Tvqa: Localized, compositional video question answering},
  author={Lei, Jie and Yu, Licheng and Bansal, Mohit and Berg, Tamara},
  booktitle={Proceedings of the 2018 conference on empirical methods in natural language processing},
  pages={1369--1379},
  year={2018}
}

@inproceedings{jiang2025specific,
  title={From specific-MLLMs to OMNI-MLLMs: a survey on MLLMs aligned with multi-modalities},
  author={Jiang, Shixin and Liang, Jiafeng and Wang, Jiyuan and Dong, Xuan and Chang, Heng and Yu, Weijiang and Du, Jinhua and Liu, Ming and Qin, Bing},
  booktitle={Findings of the Association for Computational Linguistics: ACL 2025},
  pages={8617--8652},
  year={2025}
}

@inproceedings{Maaz2023VideoChatGPT,
  title={Video-chatgpt: Towards detailed video understanding via large vision and language models},
  author={Maaz, Muhammad and Rasheed, Hanoona and Khan, Salman and Khan, Fahad},
  booktitle={Proceedings of the 62nd Annual Meeting of the Association for Computational Linguistics (Volume 1: Long Papers)},
  pages={12585--12602},
  year={2024}
}

@article{Qwen-VL,
      title={Qwen-VL: A Versatile Vision-Language Model for Understanding, Localization, Text Reading, and Beyond}, 
      author={Jinze Bai and Shuai Bai and Shusheng Yang and Shijie Wang and Sinan Tan and Peng Wang and Junyang Lin and Chang Zhou and Jingren Zhou},
      year={2023},
      eprint={2308.12966},
      archivePrefix={arXiv},
      primaryClass={cs.CV},
      url={https://arxiv.org/abs/2308.12966}, 
}

@techreport{gemini3pro,
  title        = {Gemini 3 Pro Model Card},
  author       = {{DeepMind}},
  year         = {2025},
  institution  = {Google DeepMind},
  note         = {Technical Report},
  url          = {https://storage.googleapis.com/deepmind-media/Model-Cards/Gemini-3-Pro-Model-Card.pdf}
}

@inproceedings{lin2004rouge,
  title        = {{ROUGE}: A Package for Automatic Evaluation of Summaries},
  author       = {Lin, Chin-Yew},
  booktitle    = {Text Summarization Branches Out},
  year         = {2004},
  organization = {Association for Computational Linguistics},
  pages        = {74--81},
  url          = {https://aclanthology.org/W04-1013}
}

@misc{gpt4o,
  title        = {GPT-4o System Card},
  author       = {{OpenAI}},
  year         = {2024},
  howpublished = {\url{https://arxiv.org/abs/2410.21276}},
  note         = {arXiv preprint arXiv:2410.21276}
}

@misc{gpt5mini,
  title        = {GPT-5 Mini Model Documentation},
  author       = {{OpenAI}},
  year         = {2026},
  howpublished = {\url{https://platform.openai.com/docs/models/gpt-5-mini}},
  note         = {Accessed January 2026}
}

@article{qwen3omni,
  title        = {Qwen3-Omni Technical Report},
  author       = {Xu, J. and Guo, Z. and Hu, H. and Chu, Y. and Wang, X. and He, J. and others},
  journal      = {arXiv preprint arXiv:2509.17765},
  year         = {2025},
  url          = {https://arxiv.org/abs/2509.17765}
}

@article{unimoe2,
  title={Uni-moe-2.0-omni: Scaling language-centric omnimodal large model with advanced moe, training and data},
  author={Li, Yunxin and Chen, Xinyu and Jiang, Shenyuan and Shi, Haoyuan and Liu, Zhenyu and Zhang, Xuanyu and Deng, Nanhao and Xu, Zhenran and Ma, Yicheng and Zhang, Meishan and others},
  journal={arXiv preprint arXiv:2511.12609},
  year={2025}
}

@misc{minicpmo26,
  title={MiniCPM-V: A GPT-4V Level MLLM on Your Phone},
  author={Yao, Yuan and Yu, Tianyu and Zhang, Ao and Wang, Chongyi and Cui, Junbo and Zhu, Hongji and Cai, Tianchi and Li, Haoyu and Zhao, Weilin and He, Zhihui and others},
  journal={arXiv preprint arXiv:2408.01800},
  year={2024}
}

@inproceedings{zhou2025mlvu,
  title={Mlvu: Benchmarking multi-task long video understanding},
  author={Zhou, Junjie and Shu, Yan and Zhao, Bo and Wu, Boya and Liang, Zhengyang and Xiao, Shitao and Qin, Minghao and Yang, Xi and Xiong, Yongping and Zhang, Bo and others},
  booktitle={2025 IEEE/CVF Conference on Computer Vision and Pattern Recognition (CVPR)},
  pages={13691--13701},
  year={2025},
  organization={IEEE}
}

@inproceedings{song2024moviechat,
  title={Moviechat: From dense token to sparse memory for long video understanding},
  author={Song, Enxin and Chai, Wenhao and Wang, Guanhong and Zhang, Yucheng and Zhou, Haoyang and Wu, Feiyang and Chi, Haozhe and Guo, Xun and Ye, Tian and Zhang, Yanting and others},
  booktitle={Proceedings of the IEEE/CVF Conference on Computer Vision and Pattern Recognition},
  pages={18221--18232},
  year={2024}
}

@article{patraucean2023perception,
  title={Perception test: A diagnostic benchmark for multimodal video models},
  author={Patraucean, Viorica and Smaira, Lucas and Gupta, Ankush and Recasens, Adria and Markeeva, Larisa and Banarse, Dylan and Koppula, Skanda and Malinowski, Mateusz and Yang, Yi and Doersch, Carl and others},
  journal={Advances in Neural Information Processing Systems},
  volume={36},
  pages={42748--42761},
  year={2023}
}

@article{ning2025video,
  title={Video-bench: A comprehensive benchmark and toolkit for evaluating video-based large language models},
  author={Ning, Munan and Zhu, Bin and Xie, Yujia and Lin, Bin and Cui, Jiaxi and Yuan, Lu and Chen, Dongdong and Yuan, Li},
  journal={Computational Visual Media},
  year={2025},
  publisher={TUP}
}

@article{liu2024tempcompass,
  title={Tempcompass: Do video llms really understand videos?},
  author={Liu, Yuanxin and Li, Shicheng and Liu, Yi and Wang, Yuxiang and Ren, Shuhuai and Li, Lei and Chen, Sishuo and Sun, Xu and Hou, Lu},
  booktitle={Findings of the Association for Computational Linguistics: ACL 2024},
  pages={8731--8772},
  year={2024}
}

@article{fu2024vita,
  title={Vita: Towards open-source interactive omni multimodal llm},
  author={Fu, Chaoyou and Lin, Haojia and Long, Zuwei and Shen, Yunhang and Dai, Yuhang and Zhao, Meng and Zhang, Yi-Fan and Dong, Shaoqi and Li, Yangze and Wang, Xiong and others},
  journal={arXiv preprint arXiv:2408.05211},
  year={2024}
}

@article{li2024llava,
  title={Llava-next-interleave: Tackling multi-image, video, and 3d in large multimodal models},
  author={Li, Feng and Zhang, Renrui and Zhang, Hao and Zhang, Yuanhan and Li, Bo and Li, Wei and Ma, Zejun and Li, Chunyuan},
  journal={arXiv preprint arXiv:2407.07895},
  year={2024}
}

@article{cheng2024videollama,
  title={Videollama 2: Advancing spatial-temporal modeling and audio understanding in video-llms},
  author={Cheng, Zesen and Leng, Sicong and Zhang, Hang and Xin, Yifei and Li, Xin and Chen, Guanzheng and Zhu, Yongxin and Zhang, Wenqi and Luo, Ziyang and Zhao, Deli and others},
  journal={arXiv preprint arXiv:2406.07476},
  year={2024}
}

@article{tong2022videomae,
  title={Videomae: Masked autoencoders are data-efficient learners for self-supervised video pre-training},
  author={Tong, Zhan and Song, Yibing and Wang, Jue and Wang, Limin},
  journal={Advances in neural information processing systems},
  volume={35},
  pages={10078--10093},
  year={2022}
}

@inproceedings{sun2019videobert,
  title={Videobert: A joint model for video and language representation learning},
  author={Sun, Chen and Myers, Austin and Vondrick, Carl and Murphy, Kevin and Schmid, Cordelia},
  booktitle={Proceedings of the IEEE/CVF international conference on computer vision},
  pages={7464--7473},
  year={2019}
}

@inproceedings{li2022blip,
  title={Blip: Bootstrapping language-image pre-training for unified vision-language understanding and generation},
  author={Li, Junnan and Li, Dongxu and Xiong, Caiming and Hoi, Steven},
  booktitle={International conference on machine learning},
  pages={12888--12900},
  year={2022},
  organization={PMLR}
}

@inproceedings{fu2025video,
  title={Video-mme: The first-ever comprehensive evaluation benchmark of multi-modal llms in video analysis},
  author={Fu, Chaoyou and Dai, Yuhan and Luo, Yongdong and Li, Lei and Ren, Shuhuai and Zhang, Renrui and Wang, Zihan and Zhou, Chenyu and Shen, Yunhang and Zhang, Mengdan and others},
  booktitle={Proceedings of the Computer Vision and Pattern Recognition Conference},
  pages={24108--24118},
  year={2025}
}

@inproceedings{radford2021learning,
  title={Learning transferable visual models from natural language supervision},
  author={Radford, Alec and Kim, Jong Wook and Hallacy, Chris and Ramesh, Aditya and Goh, Gabriel and Agarwal, Sandhini and Sastry, Girish and Askell, Amanda and Mishkin, Pamela and Clark, Jack and others},
  booktitle={International conference on machine learning},
  pages={8748--8763},
  year={2021},
  organization={PmLR}
}

@article{bain2022whisperx,
  title={WhisperX: Time-Accurate Speech Transcription of Long-Form Audio},
  author={Bain, Max and Huh, Jaesung and Han, Tengda and Zisserman, Andrew},
  journal={INTERSPEECH 2023},
  year={2023}
}

@misc{creative_commons_by_4_0,
  title        = {Attribution 4.0 International (CC BY 4.0)},
  author       = {{Creative Commons}},
  year         = {2013},
  howpublished = {\url{https://creativecommons.org/licenses/by/4.0/}},
  note         = {Accessed: 2025-11-10}
}

@article{wu2024longvideobench,
  title={Longvideobench: A benchmark for long-context interleaved video-language understanding},
  author={Wu, Haoning and Li, Dongxu and Chen, Bei and Li, Junnan},
  journal={Advances in Neural Information Processing Systems},
  volume={37},
  pages={28828--28857},
  year={2024}
}

@inproceedings{wang2025lvbench,
  title={Lvbench: An extreme long video understanding benchmark},
  author={Wang, Weihan and He, Zehai and Hong, Wenyi and Cheng, Yean and Zhang, Xiaohan and Qi, Ji and Ding, Ming and Gu, Xiaotao and Huang, Shiyu and Xu, Bin and others},
  booktitle={2025 IEEE/CVF International Conference on Computer Vision (ICCV)},
  pages={22958--22967},
  year={2025},
  organization={IEEE}
}

@article{chen2023vast,
  title={Vast: A vision-audio-subtitle-text omni-modality foundation model and dataset},
  author={Chen, Sihan and Li, Handong and Wang, Qunbo and Zhao, Zijia and Sun, Mingzhen and Zhu, Xinxin and Liu, Jing},
  journal={Advances in Neural Information Processing Systems},
  volume={36},
  pages={72842--72866},
  year={2023}
}

@article{rawal2024cinepile,
  title={Cinepile: A long video question answering dataset and benchmark},
  author={Rawal, Ruchit and Saifullah, Khalid and Farr{\'e}, Miquel and Basri, Ronen and Jacobs, David and Somepalli, Gowthami and Goldstein, Tom},
  journal={arXiv preprint arXiv:2405.08813},
  year={2024}
}

@article{zhou2025daily,
  title={Daily-Omni: Towards Audio-Visual Reasoning with Temporal Alignment across Modalities},
  author={Zhou, Ziwei and Wang, Rui and Wu, Zuxuan},
  journal={arXiv preprint arXiv:2505.17862},
  year={2025}
}

@article{xie2025maverix,
  title={Maverix: Multimodal audio-visual evaluation and recognition index},
  author={Xie, Liuyue and Kuthiala, Avik and Wei, George Z and Zheng, Ce and Bal, Ananya and Dabhi, Mosam and Wen, Liting and Rustagi, Taru and Lai, Ethan and Khyalia, Sushil and others},
  booktitle={Proceedings of the AAAI Conference on Artificial Intelligence},
  volume={40},
  number={32},
  pages={27090--27098},
  year={2026}
}

@article{li2024sports,
  title={Sports-qa: A large-scale video question answering benchmark for complex and professional sports},
  author={Li, Haopeng and Deng, Andong and Liu, Jun and Rahmani, Hossein and Guo, Yulan and Schiele, Bernt and Bennamoun, Mohammed and Ke, Qiuhong},
  journal={International Journal of Computer Vision},
  volume={134},
  number={5},
  pages={196},
  year={2026},
  publisher={Springer}
}

@inproceedings{chen2024cg,
  title={Cg-bench: Clue-grounded question answering benchmark for long video understanding},
  author={Chen, Guo and Liu, Yicheng and Huang, Yifei and Pei, Baoqi and Xu, Jilan and He, Yuping and Lu, Tong and Wang, Yali and Wang, Limin},
  booktitle={International Conference on Learning Representations},
  volume={2025},
  pages={45647--45682},
  year={2025}
}

@article{li2025omnivideobench,
  title={Omnivideobench: Towards audio-visual understanding evaluation for omni mllms},
  author={Li, Caorui and Chen, Yu and Ji, Yiyan and Xu, Jin and Cui, Zhenyu and Li, Shihao and Zhang, Yuanxing and Song, Zhenghao and Zhang, Dingling and He, Ying and others},
  booktitle={International Conference on Learning Representations},
  volume={2026},
  pages={138214--138236},
  year={2026}
}

@article{galougah2025aura,
  title={Aura: A fine-grained benchmark and decomposed metric for audio-visual reasoning},
  author={Galougah, Siminfar Samakoush and Raj, Rishie and Chowdhury, Sanjoy and Nag, Sayan and Duraiswami, Ramani},
  journal={arXiv preprint arXiv:2508.07470},
  year={2025}
}

@inproceedings{ataallah2024infinibench,
  title={InfiniBench: A Benchmark for Large Multi-Modal Models in Long-Form Movies and TV Shows},
  author={Ataallah, Kirolos and Bakr, Eslam Mohamed and Ahmed, Mahmoud and Gou, Chenhui and Pahwa, Khushbu and Ding, Jian and Elhoseiny, Mohamed},
  booktitle={Proceedings of the 2025 Conference on Empirical Methods in Natural Language Processing},
  pages={19496--19523},
  year={2025}
}

@techreport{boyd2022development,
  title       = {The development and psychometric properties of {LIWC-22}},
  author      = {Boyd, Ryan L. and Ashokkumar, Ashwini and Seraj, Sarah and Pennebaker, James W.},
  institution = {University of Texas at Austin},
  address     = {Austin, TX},
  year        = {2022},
  url         = {https://www.liwc.app}
}

@incollection{zimmer2020but,
  title={“But the data is already public”: on the ethics of research in Facebook},
  author={Zimmer, Michael},
  booktitle={The ethics of information technologies},
  pages={229--241},
  year={2020},
  publisher={Routledge}
}

@article{nissenbaum2004privacy,
  title={Privacy as contextual integrity},
  author={Nissenbaum, Helen},
  journal={Wash. L. Rev.},
  volume={79},
  pages={119},
  year={2004},
  publisher={HeinOnline}
}

@article{tausczik2010psychological,
  title={The psychological meaning of words: LIWC and computerized text analysis methods},
  author={Tausczik, Yla R and Pennebaker, James W},
  journal={Journal of language and social psychology},
  volume={29},
  number={1},
  pages={24--54},
  year={2010},
  publisher={Sage Publications Sage CA: Los Angeles, CA}
}

@misc{eeoc_race_color,
  author = {{U.S. Equal Employment Opportunity Commission}},
  title  = {Section 15: Race and Color Discrimination},
  year   = {2006},
  url    = {https://www.eeoc.gov/laws/guidance/section-15-race-and-color-discrimination},
  note   = {EEOC Compliance Manual. Accessed: 2025-01-27}
}

@article{li2025baichuan,
  title={Baichuan-omni-1.5 technical report},
  author={Li, Yadong and Liu, Jun and Zhang, Tao and Chen, Song and Li, Tianpeng and Li, Zehuan and Liu, Lijun and Ming, Lingfeng and Dong, Guosheng and Pan, Da and others},
  journal={arXiv preprint arXiv:2501.15368},
  year={2025}
}

@article{hong2025worldsense,
  title={Worldsense: Evaluating real-world omnimodal understanding for multimodal llms},
  author={Hong, Jack and Yan, Shilin and Cai, Jiayin and Jiang, Xiaolong and Hu, Yao and Xie, Weidi},
  booktitle={International Conference on Learning Representations},
  volume={2026},
  pages={52423--52443},
  year={2026}
}

@article{liu2025ola,
  title={Ola: Pushing the frontiers of omni-modal language model},
  author={Liu, Zuyan and Dong, Yuhao and Wang, Jiahui and Liu, Ziwei and Hu, Winston and Lu, Jiwen and Rao, Yongming},
  journal={arXiv preprint arXiv:2502.04328},
  year={2025}
}

@misc{eeoc_sex,
  author = {{U.S. Equal Employment Opportunity Commission}},
  title  = {Sex-Based Discrimination},
  year   = {2025},
  url    = {https://www.eeoc.gov/sex-based-discrimination},
  note   = {Accessed: 2025-01-27}
}

@misc{eeoc_age,
  author = {{U.S. Equal Employment Opportunity Commission}},
  title  = {Age Discrimination},
  year   = {2025},
  url    = {https://www.eeoc.gov/age-discrimination},
  note   = {Accessed: 2025-01-27}
}

\appendix

\section{Data Collection Details}
\label{app:collection}
\begin{figure*}[!h]
    \centering
    \includegraphics[width=\linewidth]{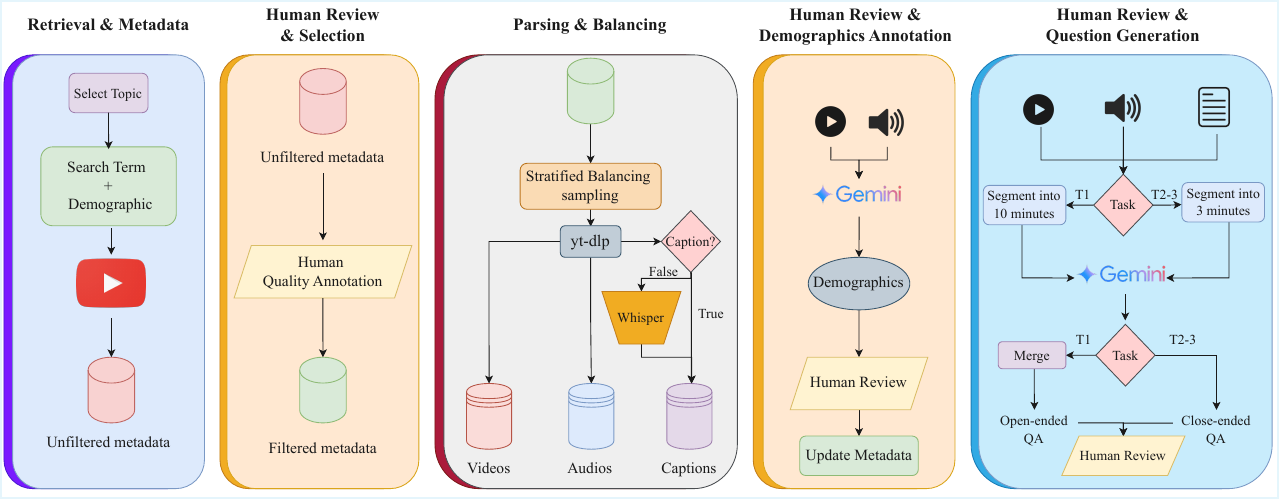}
    \caption{\textbf{SONIC-O1 data curation and annotation pipeline.}
    We first retrieve candidate real-world scenario videos via topic- and demographic-augmented search queries and filter by licensing constraints (CC BY 4.0), then download video/audio and any available captions. Human reviewers screen candidates for quality and coverage, after which we parse metadata and perform balanced sampling (e.g., by topic and duration). For benchmark construction, videos are segmented into 10-minute chunks for summarization (T1) and into overlapping 3-minute segments for MCQ and temporal localization (T2--T3). If captions are unavailable, we obtain speech transcripts via ASR; otherwise we use downloaded captions. Draft demographic metadata and task annotations can be bootstrapped with model-assisted tools (e.g., Gemini/ASR) and are subsequently corrected and verified by domain experts, with ambiguous or low-evidence items removed. Arrows indicate the flow from raw retrieval to finalized, human-verified instances.}
    \label{fig:pipeline}
\end{figure*}

As illustrated in the \textit{Retrieval \& Metadata} stage of Figure~\ref{fig:pipeline}, we retrieve real-world scenario videos using the YouTube Data API v3~\cite{google_youtube_data_api_v3}, following established ethical frameworks for web-based data collection that respect contextual integrity and avoid privacy violations~\cite{zimmer2020but, nissenbaum2004privacy}. We restrict candidates to videos published under the Creative Commons Attribution 4.0 (CC BY 4.0) license~\cite{creative_commons_by_4_0}, which explicitly permits redistribution for research purposes. Videos under YouTube's Standard License are excluded due to redistribution restrictions.

Our search strategy uses topic-specific base queries that capture real-world interactions (e.g., consultations, interviews, hearings), and expands them with demographic descriptors (e.g., ``young adult'', ``woman'', ``Black'', ``Middle Eastern'', ``First Nations''), following federal standards for demographic categorization~\cite{eeoc_race_color}, to improve coverage. These demographic keywords are used only for query expansion and are not treated as labels; final demographic annotations are produced later through a human-verified review process (Appendix~\ref{app:ai-assisted}). This approach ensures that our data collection respects the original context and licensing terms under which content was shared~\cite{zimmer2020but}, avoiding privacy violations that can occur when data is extracted from contexts with restricted access expectations.

\subsection{Curation Strategy and Licensing Compliance}
% %help me repharase this which lisence is it (i) or (ii)
% by search do u mean curation strategy??
% % search strategy for curation + setting the API to commonCreative, it's explicit (oh yes, like how we curated the metadata, like the queries of each topic, demographics how did we add them, it is stage 1 in figure A1- the subsection can be "curation of metadta" or "Search Strategy")  Check now? 

\label{app:search}
To construct SONIC-O1, we followed a structured video curation pipeline combining keyword-based API retrieval with strict license filtering. We restricted candidates to CC BY 4.0 videos and performed additional manual verification to ensure redistribution and research-use compliance.

\begin{figure*}
    
\begin{tcolorbox}[
    colback=blue!5!white,
    colframe=blue!75!black,
    width=\textwidth,
    title=Example Search Query Construction (4 of 13 Topics Shown),
    fonttitle=\bfseries,
    boxrule=0.5mm,
    arc=2mm
]
\textbf{Base Query Examples Across Topics:}

\textit{Topic 2: Job Interviews (Professional Domain)}
\begin{itemize}
    \item \texttt{panel interview full interview}
    \item \texttt{technical interview session full}
    \item \texttt{candidate interview recording}
\end{itemize}

\textit{Topic 1: Patient-Doctor Consultations (Community/Public Health)}
\begin{itemize}
    \item \texttt{doctor patient conversation full session}
    \item \texttt{clinic consultation recording}
    \item \texttt{telehealth visit recording}
\end{itemize}

\textit{Topic 5: Courtroom Proceedings (Legal/Civic)}
\begin{itemize}
    \item \texttt{oral argument}
    \item \texttt{sentencing hearing full recording courtroom}
    \item \texttt{small claims court full hearing official recording}
\end{itemize}

\textit{Topic 7: Public Transportation Conflicts (Community/Public Health)}
\begin{itemize}
    \item \texttt{bus passenger fight driver -news -compilation}
    \item \texttt{train passenger confrontation cctv}
    \item \texttt{airport security passenger meltdown bodycam}
\end{itemize}

\vspace{0.2cm}
\textbf{Demographic Variations Generated (Example: Job Interviews):}
\begin{itemize}
    \item \textbf{Race:} \texttt{Black panel interview full interview}, \texttt{Asian technical interview session full}, \texttt{Hispanic candidate interview recording}, \texttt{Middle Eastern panel interview full interview}, \texttt{White technical interview session full}
    \item \textbf{Gender:} \texttt{woman panel interview full interview}, \texttt{man technical interview session full}
    \item \textbf{Age:} \texttt{young adult panel interview full interview}, \texttt{middle aged technical interview session full}, \texttt{older adult candidate interview recording}
\end{itemize}

\vspace{0.2cm}
\textbf{Special Handling for Underrepresented Groups:}
\begin{itemize}
    \item \textbf{Arab:} Multiple variations used: \texttt{Arab}, \texttt{Middle Eastern}, \texttt{Arabic}, \texttt{MENA}, \texttt{Arab American}
    \item \textbf{Indigenous:} Multiple variations used: \texttt{Indigenous}, \texttt{Native American}, \texttt{First Nations}, \texttt{Aboriginal}, \texttt{tribal}
\end{itemize}

\vspace{0.2cm}
This approach generated approximately 40--50 search queries per topic (4--6 base queries $\times$ 11 demographic variations), yielding diverse representation across racial, gender, and age groups. Total queries across 13 topics: $\sim$600 searches.
\end{tcolorbox}
\end{figure*}

\subsection{Inclusion Criteria}
\label{app:inclusion}

Candidate videos must satisfy the following requirements:

\begin{itemize}
    \item \textbf{Time of curation:} Only videos published from January 2020 to October 2025 to ensure topical coverage, availability, and consistency in retrieval conditions.
    \item \textbf{Keywords:} Topic-specific base queries that capture authentic, real-world interactions, further expanded with demographic descriptors for race, gender, and age.
    \item \textbf{Video lengths:} Recordings sufficient to capture complete interactions, typically ranging from 30 seconds-60 minutes.
    \item \textbf{Language:} Primarily English-language recordings with clear, intelligible speech to support multimodal transcription and reasoning tasks.
    \item \textbf{Topics \& categories:} Videos covering 13 topics across five domains, selected to reflect common, high-impact real-world interactions relevant for multimodal reasoning.
    \item \textbf{Demographic diversity:} Videos featuring speakers across diverse \textbf{race} (White, Black, Asian, Indigenous, Arab, Hispanic), \textbf{gender} (male, female), and \textbf{age} groups (18--24, 25--39, 40+), following federal standards for demographic categorization~\cite{eeoc_race_color, eeoc_sex, eeoc_age}.
\end{itemize}

\subsection{Exclusion Criteria}
\label{app:exclusion}

Our upstream retrieval produces many candidates that are off-topic, low quality, or unsuitable for ethical/legal use. We therefore apply a two-stage filtering procedure: (i) lightweight metadata-only heuristics to remove obvious spam and outliers, followed by (ii) a \textbf{domain-expert human screening pass} (Figure~\ref{fig:quality_review}) in which candidates are manually viewed and labeled as \textit{Good} or \textit{Bad}. We retain only videos labeled \textit{Good} and licensed under \textbf{CC BY 4.0}.

A video is rejected if it meets any of the following criteria:
\begin{itemize}
    \item \textbf{Low media quality:} poor audio or visual quality that hinders transcription, comprehension, or annotation.
    \item \textbf{Non-continuous content:} heavily edited or montage-style videos that do not preserve continuous interactions.
    \item \textbf{Irrelevant topic:} content not aligned with the target domains or study scope.
    \item \textbf{Sensitive or unsafe content:} videos containing sensitive personal data or content that cannot be used in an ethical manner.
    \item \textbf{Language constraint:} non-English, audio-dominant content that prevents reliable English annotation.
    \item \textbf{License constraint:} videos under YouTube's Standard License (i.e., not CC BY 4.0).
\end{itemize}

\subsection{Filtering and Media Processing}
\label{app:processing}

\paragraph{Stage 1: Metadata-only pre-filtering.}
Before human review, we apply lightweight heuristics using only public metadata to remove obviously low-quality uploads while preserving licensing and safety constraints. We drop candidates with missing basic signals (e.g., missing duration), enforce a duration window, and require a minimum view count (500 by default; 100 for scarce topics such as emergency response). We further filter clickbait/spam via simple title-based signals (e.g., excessive capitalization with caps ratio $>0.55$, heavy punctuation/repetition, emoji spam with $>5$ emojis, and common spam phrases), suspicious engagement (e.g., $>10$k views with 0 likes and 0 comments, or $>5$k views with engagement rate $<10^{-4}$), and invalid channels (e.g., extremely short or numeric-only channel identifiers). Remaining candidates receive a simple metadata quality score that rewards suitable durations and healthy engagement and penalizes generic/clickbait titles.

\paragraph{Stage 2: Human Review \& Selection.}
During \textit{Human Review \& Selection} (Figure~\ref{fig:pipeline}), reviewers label each candidate video as \textit{Good} or \textit{Bad} based on topic relevance, audio--visual quality, safety, and licensing.

\paragraph{Balanced sampling and media pre-processing.}

From the human-accepted pool,  we construct the final set using a topic-capped,  a balanced sampling procedure designed to maximize coverage under licensing and quality constraints. Concretely, we cap each topic at up to 25 videos when sufficient candidates exist and stratify selections across short, medium, and long durations to avoid over-representing any single length regime. When a topic contains fewer than the target after filtering, we include all eligible videos for that topic (i.e., no downsampling).

For each selected video, we download the MP4 (capped at 1080p to balance quality and storage), extract stereo audio (48\,kHz, 192\,kbps), and retrieve English captions via the YouTube caption API when available. Videos without native captions are flagged for automatic transcription. The output is a three-modality package per video---video (MP4), audio (M4A), and text (SRT)---plus a JSON metadata record containing IDs, timestamps, and processing artifacts for reproducibility.

\subsection{Video Filtering Summary}
\label{app:funnel}
%this just to show number of videos from metadta to screening - can be removed if no need
Table~\ref{tab:selection_funnel} documents the filtering process from initial search to final dataset.

\begin{table}[h]
\centering
\caption{\textbf{Video filtering summary.} Counts of candidate videos at each curation stage: initial API retrieval during time 2020-2025, CC BY 4.0 license filtering, and final acceptance after metadata pre-filtering and human \textit{Good/Bad} screening. The sharp reduction reflects deliberate prioritization of licensing compliance, annotation quality, and balanced coverage over scale }
\label{tab:selection_funnel}
\begin{tabular}{l r}
\toprule
\textbf{Stage} & \textbf{Count} \\
\midrule
Initial search results & 2,237 \\
After license filtering (CC BY 4.0 only) & 1,794 \\
Final dataset (post quality assurance) & 231 \\
\bottomrule
\end{tabular}
\end{table}

\subsection{Topical Coverage}
\label{app:topics}

Table~\ref{tab:topics_detailed} summarizes SONIC-O1's topical coverage across five conversational domains and 13 topics, including video counts and duration statistics for the curated set.

\begin{table*}[h]
  \centering
  \small
    \caption{\textbf{Topical coverage of SONIC-O1.} Breakdown of the 5 domains and 13 topics, reporting the number of curated videos (Samples) and their aggregate/average durations after the full selection pipeline.}
  \label{tab:topics_detailed}
  \begin{tabular}{p{2cm}p{6cm}p{1.5cm}p{2cm}p{2cm}}
    \toprule
    \textbf{Category} & \textbf{Topics} & \textbf{Samples} & \textbf{Total Duration (min)} & \textbf{Avg Duration (min)} \\
    \midrule
    Professional
      & Job interviews; workplace meetings
      & 49
      & 708.8
      & 14.5 \\
    Educational
      & Parent--teacher conferences
      & 18
      & 651.5
      & 36.2 \\
    Legal / Civic
      & Courtroom proceedings; community town halls
      & 32
      & 843.8
      & 26.4 \\
    Service-Oriented
      & Customer service; restaurant service; housing/apartment tours
      & 63
      & 726.3
      & 11.3 \\
    Community / Public Health
      & Medical (patient--doctor); emergency response; public transportation conflicts; mental-health counseling; Olympics/sports
      & 69
      & 681.2
      & 9.7 \\
    \bottomrule
  \end{tabular}
\end{table*}

\subsection{AI-Assisted Annotation Process}
\label{app:ai-assisted}

\paragraph{Model-assisted drafting.}
In the \textit{Demographics Annotation} and \textit{Question Generation} stages (Figure~\ref{fig:pipeline}), we use Gemini 2.5 Flash to produce \textbf{initial} drafts from joint multimodal signals (video, audio, and captions), where captions are always available (downloaded captions when present, otherwise generated via automatic speech recognition (ASR) using Whisper~\cite{bain2022whisperx}). This approach follows established practices in recent benchmarking efforts~\cite{galougah2025aura,ataallah2024infinibench}. Importantly, all drafts are then reviewed and corrected by domain expert annotators (Appendix~\ref{app:team_formation}) using our review interface (Appendix~\ref{app:interface}). Only human-verified annotations are retained for dataset construction and benchmarking.

\paragraph{Demographic annotation workflow.}
We first draft demographics at the \textbf{video level} as a set of tags (e.g., race, gender, age, language) using Gemini 2.5 Flash, and annotators review and correct these tags. During audio-video question-answering (AVQA) instance construction, we \textbf{propagate} the verified video-level tags to each segment by expanding them into per-segment counts of unique individuals per demographic combination. This expansion step is model-assisted but constrained to the human-verified tag set (i.e., it cannot introduce new demographic values), and is again reviewed and corrected by the domain expert annotators. For each segment, we store human-verified demographic tags, optional confidence scores, and per-segment counts. For multi-person segments, we store counts of individuals grouped by unique demographic combinations (e.g., "2 White males aged 40+, 1 Asian female aged 25-39").

\paragraph{Task-specific annotation workflow.}
For all three tasks (summarization, MCQ, temporal localization), we use Gemini 2.5 Flash to draft questions, answers, and task-specific annotations. All model outputs undergo mandatory human verification, correction, and revision. The full prompts are provided in Appendix~\ref{app:ai-generation-prompts}.

\paragraph{Annotation schema.}
Each item includes the question or summary and a reference answer; for Tasks 2--3, we additionally store a short reasoning rationale. Each segment also includes human-verified demographic metadata (and optional confidence scores) with counts for each unique demographic combination when multiple individuals appear.

\subsubsection{AI Generation Prompts}
\label{app:ai-generation-prompts}
Below are the complete prompts used to generate preliminary annotation drafts for each task and the initial demographics prompt. All AI-generated outputs underwent mandatory human verification as described in Section~\ref{app:task-procedures}.

\paragraph{Demographic Generation Prompt.}

\begin{tcolorbox}[breakable, width=\linewidth, colback=gray!5!white, colframe=gray!50!black,
                  title=Demographics: AI Annotation Generation, fontupper=\small]
\begin{lstlisting}[breaklines=true, basicstyle=\ttfamily\scriptsize, columns=fullflexible]

You are a demographics annotation specialist for academic research. Analyze the 
provided multimodal media (video, audio, and captions/transcripts) and identify the 
demographic characteristics of all individuals who appear 
visually or speak. Be objective and avoi stereotype-based
assumptions. Return ONLY valid JSON that can be directly parsed (no extra text).

Analyze this MULTIMODAL media to identify demographics of ALL people who appear
visually or speak.

MEDIA INFORMATION:
- Title: {title}
- Duration: {duration_seconds} seconds
- Topic: {topic_name}

INPUTS PROVIDED:
You have access to multiple modalities for this analysis:
1. VIDEO: Visual content showing individuals (if available)
2. AUDIO: Sound/speech from individuals (may be embedded in video or separate)
3. TRANSCRIPT/CAPTIONS: Text representation of spoken content
{transcript_preview}

IMPORTANT: Use ALL available modalities together for the most accurate analysis.
Cross-reference visual, audio, and text cues to identify and confirm demographics.

---

ANALYSIS GUIDELINES BY MODALITY:

**VIDEO ANALYSIS (when available):**
- Race/Ethnicity: Primary visual assessment of facial features, skin tone
- Gender: Visual presentation (appearance, clothing, mannerisms)
- Age: Visual appearance (facial features, hair, physical characteristics)
- Language: Can support audio analysis with lip movements

**AUDIO ANALYSIS (always available in video or as separate audio):**
- Gender: Vocal pitch and timbre (deep pitch -> Male, high pitch -> Female)
- Age: Vocal characteristics (youthful energy vs. mature tone 
vs. older/quavering voice)
- Language: Primary method for identifying spoken languages and accents
- Race/Ethnicity: MAY provide supporting evidence via accent, but use LOW confidence

**TRANSCRIPT/CAPTION ANALYSIS (when available):**
- Language: Confirms which languages are spoken
- Speaker identification: Helps count unique individuals
- Context: Provides semantic understanding of the interaction
- Names/References: May help distinguish between speakers

**CROSS-MODAL VERIFICATION:**
- When you have both video and audio, verify gender and age across both modalities
- Use transcript to confirm language identification from audio
- Count unique individuals by combining visual appearances with distinct voices
- Higher confidence when multiple modalities agree

---

DEMOGRAPHIC CATEGORIES TO IDENTIFY:

1. RACE/ETHNICITY (select all that apply):
   - White: European descent appearance
   - Black: African descent appearance
   - Asian: East/Southeast/South Asian appearance
   - Indigenous: Native American/Aboriginal appearance
   - Arab: Middle Eastern/North African appearance
   - Hispanic: Latin American appearance
   - Note: Primarily visual assessment. Audio-only inference should have LOW confidence
     unless very strong accent indicators.

2. GENDER (select all that apply):
   - Male: Masculine presenting individuals OR deep vocal pitch
   - Female: Feminine presenting individuals OR high vocal pitch
   - Use visual cues first, audio cues second

3. AGE GROUPS (select all that apply):
   - Young (18-24): Visual appearance OR youthful voice
   - Middle (25-39): Visual appearance OR mature voice
   - Older adults (40+): Visual appearance OR older voice characteristics
   - Combine visual and audio cues for best accuracy

4. LANGUAGE (select all spoken):
   - Identify all languages and distinct accents heard
   - Use audio AND transcript to confirm
   - Default to ["English"] if only English is spoken

---

ANALYSIS METHODOLOGY (MULTIMODAL):

**Step 1: IDENTIFY INDIVIDUALS**
- Count unique people visible in video
- Count unique voices in audio (use transcript speaker labels if available)
- Total = unique individuals across both modalities

**Step 2: ASSESS EACH INDIVIDUAL**
For each person:
- If visible: Use video for race, gender, age (HIGH confidence)
- If speaking: Use audio for gender, age, language (MEDIUM-HIGH confidence)
- If both: Cross-verify and use HIGHEST confidence
- Use transcript to confirm language and count speakers

**Step 3: ASSIGN CONFIDENCE**
- 0.9-1.0: Clear visual evidence OR audio + visual agreement
- 0.7-0.89: Clear audio evidence OR single modality with good clarity
- 0.5-0.69: Uncertain (e.g., accent-based inference, unclear visuals)
- Below 0.5: Do not include

**Step 4: LIST UNIQUE DEMOGRAPHICS**
- Aggregate all unique demographics across all individuals
- Include only those meeting minimum confidence threshold

---

OUTPUT FORMAT:

Return ONLY this JSON structure with no additional text:

{
  "demographics_detailed": {
    "race": [list unique races observed with sufficient confidence],
    "gender": [list unique genders observed with sufficient confidence],
    "age": [list unique age groups observed with sufficient confidence],
    "language": [list languages/accents actually spoken]
  },
  "demographics_confidence": {
    "race": {"race1": confidence1, "race2": confidence2},
    "gender": {"gender1": confidence1, "gender2": confidence2},
    "age": {"age1": confidence1, "age2": confidence2},
    "language": {"language1": confidence1}
  },
  "demographics_annotation": {
    "model": "{model_name}",
    "annotated_at": "{timestamp}",
    "individuals_count": total_number_of_unique_individuals,
    "modalities_used": [list of "video", "audio", "transcript" that were available],
    "explanation": "Brief factual description combining
    visual, audio, and transcript observations."
  }
}

CRITICAL REMINDERS:
- Use ALL available modalities (video, audio, transcript) together
- Cross-verify demographics across modalities for higher confidence
- Return ONLY valid JSON
- No text before or after the JSON
- Empty arrays are acceptable if no confident matches found
\end{lstlisting}
\end{tcolorbox}

\paragraph{Task 1: Summarization Generation Prompt.}

\begin{tcolorbox}[breakable, width=\linewidth, colback=gray!5!white, colframe=gray!50!black, 
                  title=Task 1: AI Annotation Generation (Map Phase), fontupper=\small]
\begin{lstlisting}[breaklines=true, basicstyle=\ttfamily\scriptsize, columns=fullflexible]
You are a precise video segment summarizer.

SEGMENT INFORMATION:
- Segment time: {start_time}s to {end_time}s ({duration}s duration)
- Video title: {title}
- Topic: {topic_name}

TRANSCRIPT/CAPTIONS:
{transcript_text}

YOUR TASK:
Summarize this video segment concisely and accurately.

RULES:
- Maximum {max_words} words
- Keep strict chronology - describe events in the order they occur
- Prefer facts that are audible in the transcript or visible in the video
- If nothing meaningful happens, state "No salient events in this segment"
- Be specific: include names, actions, objects, and key details
- Do NOT speculate beyond what you see/hear

OUTPUT FORMAT (JSON):
{
  "segment_start": "{start_time}",
  "segment_end": "{end_time}",
  "summary": "120 words maximum describing what happens in this segment...",
  "mini_timeline": [
    {"time": "MM:SS", "title": "Event name", "note": "One line description"},
    {"time": "MM:SS", "title": "Another event", "note": "Brief detail"}
  ],
  "entities": ["names", "objects", "places", "key terms mentioned"],
  "confidence": 0.85
}

CRITICAL:
- Return ONLY valid JSON
- No markdown, no extra text
- Timestamps in mini_timeline should be relative to SEGMENT start time
- Include 2-5 timeline items for this segment

Begin analysis:
\end{lstlisting}

\end{tcolorbox}

\begin{tcolorbox}[breakable, width=\linewidth, colback=gray!5!white, colframe=gray!50!black, 
                  title=Task 1: AI Annotation Generation (Reduce Phase), fontupper=\small]
\begin{lstlisting}[breaklines=true, basicstyle=\ttfamily\scriptsize, columns=fullflexible]
You are an editor merging ordered segment summaries into a comprehensive video summary.

VIDEO INFORMATION:
- Video ID: {video_id}
- Title: {title}
- Topic: {topic_name}
- Total duration: {duration}s
- Number of segments: {num_segments}

SEGMENT SUMMARIES:
{segment_summaries_json}

YOUR TASK:
Merge these segment summaries into a single comprehensive video summary.

PRODUCE:
1. TL;DR: {num_bullets} concise bullet points
2. Detailed summary: {max_words_detailed} words (structure: Purpose -> Key Points -> 
Outcomes)
3. Global timeline: {timeline_min}-{timeline_max} items covering the entire video
4. Glossary: {glossary_min}-{glossary_max} key entities/terms from the full video

RULES:
- Remove duplicates across segments
- Keep strict chronological order
- Be concise and factual
- Prefer events that appear in multiple segments or are clearly important
- Timeline should span from video start to end with 
actual timestamps (HH:MM:SS format)
- Glossary should include: people's names, important objects, 
locations, technical terms

OUTPUT FORMAT (JSON):
{
  "summary_short": [
    "- First key point...",
    "- Second key point...",
    "- Third key point...",
    "- Fourth key point...",
    "- Fifth key point..."
  ],
  "summary_detailed": "200-300 word comprehensive summary. 
  Start with the video's purpose,
  cover key points in chronological order, and conclude with
  outcomes or main takeaways...",
  "timeline": [
    {
      "start": "00:00:12",
      "end": "00:00:45",
      "title": "Introduction",
      "note": "Brief description"
    },
    {
      "start": "00:05:30",
      "end": "00:06:15",
      "title": "Main topic",
      "note": "Key details"
    },
    {
      "start": "00:12:00",
      "end": "00:13:30",
      "title": "Conclusion",
      "note": "Final points"
    }
  ],
  "glossary": ["Entity 1", "Entity 2", "Important Term", "Key Person", "Location"],
  "confidence": 0.88
}

CRITICAL:
- Return ONLY valid JSON
- No markdown, no extra text
- Timeline timestamps must be in HH:MM:SS format relative to VIDEO start
- Confidence should reflect how well segments agreed and information quality

Begin merging
\end{lstlisting}

\end{tcolorbox}

\paragraph{Task 2: Multiple-Choice Question Generation Prompt.}
\begin{tcolorbox}[breakable, width=\linewidth, colback=gray!5!white, colframe=gray!50!black, 
                  title=Task 2: AI Annotation Generation, fontupper=\small]
\begin{lstlisting}[breaklines=true, basicstyle=\ttfamily\scriptsize, columns=fullflexible]
You are a meticulous multimodal annotator creating challenging multiple-choice 
questions that test deep understanding of the content.

SEGMENT INFORMATION:
- Video ID: {video_id}
- Topic: {topic_name}
- Time range: {start_time}s to {end_time}s
- Duration: {duration}s

TRANSCRIPT/CAPTIONS:
{transcript_text}

YOUR TASK:
Create challenging, thought-provoking multiple-choice questions that require 
reasoning and inference to answer correctly.

MULTIMODAL UNDERSTANDING FOCUS:
This task tests a model's ability to comprehend and reason about:
- Visual information: Actions, objects, settings, body language, demonstrations, 
  equipment, text on screen
- Auditory information: Spoken dialogue, explanations, tone, background sounds, 
  verbal instructions
- Integrated understanding: Connecting what is shown with what is said to form 
  complete understanding

CRITICAL: Questions should naturally require multiple sources of information 
when possible
- Prioritize questions where explanations clarify what is demonstrated (or vice versa)
- Create questions about relationships between actions and explanations
- Test understanding that requires integrating multiple cues

QUESTION DIFFICULTY REQUIREMENTS:
AVOID simple recall questions - Don't ask questions that can be answered by 
simply remembering one fact.

PREFER questions that require:
- Integration: "Based on the procedure shown and the safety warnings mentioned, 
  what complication is being prevented?"
- Correlation: "What is the key difference between the described technique and 
  the demonstrated approach?"
- Inference: "What condition is most likely being assessed based on the complaints 
  and examination techniques?"
- Guided Analysis: "According to the explanation, what is the purpose of the 
  specific hand positioning observed?"
- Application: "If a patient presented with the symptoms described and signs shown, 
  which intervention would be most appropriate?"
- Cause-and-effect: "What is the physiological reason mentioned for the clinical 
  finding observed?"

Question complexity guidelines:
- Require connecting 2-3 pieces of information
- Ask "why" or "how" questions that need comprehensive understanding
- Test comprehension of relationships between different elements
- Require understanding of underlying principles from available evidence
- Questions should be 15-30 words to allow for complexity

OPTIONS REQUIREMENTS:
- Provide EXACTLY {num_options} options labeled (A) through ({last_option_letter})
- The LAST option ({last_option_letter}) must ALWAYS be: 
  "({last_option_letter}) Not enough evidence"
- IMPORTANT: "Not enough evidence" IS a valid correct answer when:
  * The content doesn't provide sufficient information to answer confidently
  * Multiple interpretations are equally plausible from the evidence
  * Key information needed for reasoning is missing
  * The question requires inference beyond what can be reasonably concluded
- For content-based answers (positions A through {second_to_last_letter}), 
  create principled distractors:
  * One near-miss (plausible but incorrect reasoning)
  * One salient decoy (addresses part of the question but misses integration)
  * One partial trap (correct if considering incomplete information)
  * The correct answer (requires proper comprehensive understanding)

ANSWER POSITION DISTRIBUTION:
- Distribute correct answers evenly across ALL {num_options} positions 
  (~{distribution_percentage}% each)
- DO NOT favor middle positions - Consciously vary answer placement
- DO NOT avoid position ({last_option_letter}) - Use it when genuinely appropriate

EVIDENCE TAGS:
Choose from this controlled vocabulary ONLY:
{evidence_tags_list}
Use tags that are actually present in the content.

requires_audio:
- Set to true when transcript/audio is NECESSARY to answer correctly
- Set to false ONLY if visual information alone is sufficient
- Default to true for questions requiring comprehensive understanding

OUTPUT FORMAT:
CRITICAL: Return a SINGLE JSON object (not an array):
{
  "question": "string",
  "options": ["(A) ...", "(B) ...", "(C) ...", "(D) ...", 
              "({last_option_letter}) Not enough evidence"],
  "answer_index": integer (0-{max_index}),
  "answer_letter": "string (A-{last_option_letter})",
  "rationale": "string",
  "evidence_tags": ["tag1", "tag2"],
  "requires_audio": boolean,
  "confidence": float (0.0-1.0)
}

CRITICAL RULES:
- Return ONLY ONE JSON object (not an array)
- Return ONLY valid JSON with no markdown, no extra text
- DO NOT wrap in ```json``` markers
- DO NOT include "Question:" prefix
- Each option MUST include its letter: "(A)", "(B)", etc.
- Include BOTH "answer_index" AND "answer_letter"
- The last option must ALWAYS be "({last_option_letter}) Not enough evidence"
- USE "Not enough evidence" as correct answer ~10-15% of the time
- Distribute correct answers evenly across ALL positions 
  (~{distribution_percentage}% each)
- Only use evidence_tags from the provided list
- Questions must require reasoning, inference, or application (15-30 words)

Now generate ONE MCQ question as a single JSON object for this segment:
\end{lstlisting}

\end{tcolorbox}

\paragraph{Task 3: Temporal Localization Generation Prompt.}
\begin{tcolorbox}[breakable, width=\linewidth, colback=gray!5!white, colframe=gray!50!black, 
                  title=Task 3: AI Annotation Generation, fontupper=\small]
\begin{lstlisting}[breaklines=true, basicstyle=\ttfamily\scriptsize, columns=fullflexible]
You are a careful video annotator creating temporal reasoning questions.

VIDEO CONTEXT:
- Video ID: {video_id}
- Duration: {duration} seconds (you are watching from 0.0s to {duration}s)
- Target: Generate {num_questions} questions
- Time unit: SECONDS (all timestamps must be decimal seconds)

TRANSCRIPT (if available):
{transcript_text}

================================
STEP 1: WATCH AND UNDERSTAND THE VIDEO
================================

First, watch the entire video carefully from start to finish.
- Note major events, actions, and scene changes
- Pay attention to both audio (speech, sounds) and visual (actions, objects) cues
- Observe the temporal flow and relationships between events
- Prefer anchors tied to sharp audio/speech events (because they are easier to 
  re-locate)

While watching, internally number events in the order they occur: E1, E2, E3, ...
You will refer to these IDs in your rationale to make the temporal chain 
unambiguous.

================================
STEP 2: IDENTIFY KEY EVENTS WITH PRECISE TIMESTAMPS
================================

As you watch, note down significant events with their exact timing in DECIMAL 
SECONDS:

CRITICAL TIMESTAMP FORMAT:
[OK] ALWAYS use decimal seconds: 5.2, 45.0, 78.5, 125.0
[X] NEVER use MM:SS format: 1:18, 2:30, 5:45
[X] NEVER concatenate minutes+seconds: "1 minute 18 seconds" is 78.0, NOT 118

CONVERSION RULE (important!):
- If you think "1 minute 18 seconds" -> calculate 1 x 60 + 18 = 78.0 seconds
- If you think "2 minutes 30 seconds" -> calculate 2 x 60 + 30 = 150.0 seconds
- If you think "5 seconds" -> write 5.0 seconds

Example event list (create your own):
- Person says "hello" at 5.2 seconds
- Door closes at 23.8 seconds
- Phone rings at 67.5 seconds
- Person picks up phone at 71.0 seconds

All times must be between 0.0 and {duration}.

================================
STEP 3: PLAN YOUR QUESTIONS
================================

Now select {num_questions} event pairs that have clear temporal relationships.

For each question, identify:
1. ANCHOR event (the reference point) - ideally something sharp like speech or 
   a distinct visual change
2. TARGET event (what we're searching for)
3. TEMPORAL RELATION between them

TEMPORAL RELATIONS TO USE (mix these):

after - Target occurs sometime after anchor completes
Example: "After the speaker says 'let's begin', when does the camera show 
the desk?"

once_finished - Target occurs immediately after anchor completes
Example: "Once the woman finishes writing, when does she turn around?"

next - Target is the next occurrence of a similar action/person/event
Example: "When is the next time the teacher speaks after the student asks 
a question?"

during - Target happens while anchor is ongoing
Example: "While the blue slide is displayed, when does the speaker point to 
the chart?"

before - Target happens before anchor (use sparingly)
Example: "Before the host introduces the guest, when does the music start?"

QUALITY CHECKS FOR EACH QUESTION:
- Both anchor and target clearly exist in the video
- There's a genuine temporal relationship (not random)
- The question tests temporal reasoning, not just recognition
- Timestamps are verifiable by watching the video
- Question is specific and unambiguous
- CRITICAL: end_s captures the COMPLETE event, not just when it begins
  - For speech: end when speaker finishes the complete thought/sentence
  - For actions: end when action fully completes
  - For visual elements: end when element disappears or transitions away

================================
STEP 4: VERIFY YOUR TIMESTAMPS
================================

For each question you plan to generate:

VERIFICATION CHECKLIST:
1. [OK] Locate anchor event -> Note time in decimal seconds
2. [OK] Locate target event START -> Note time in decimal seconds
3. [OK] Locate target event END -> Watch until event FULLY COMPLETES
   - Don't stop at first appearance - watch the entire event unfold
   - For speech: wait until the speaker finishes the complete sentence/explanation
   - For actions: wait until action concludes (not just begins)
   - For visual elements: note when they disappear or transition
4. [OK] Verify temporal relationship is correct
5. [OK] Double-check: converted MM:SS to pure seconds correctly?
6. [OK] Confirm both times are between 0.0 and {duration}
7. If two plausible target moments are <0.4 seconds apart AND the video frame 
   rate is unknown -> treat this as ambiguous and abstain

EXAMPLE VERIFICATION:
- I observe an event that appears to be "1 minute 23 seconds" into video
- CALCULATION: 1 x 60 + 23 = 83 seconds
- [OK] CORRECT: Write "start_s": 83.0
- [X] WRONG: Write "start_s": 123 (this is concatenation error!)
- [X] WRONG: Write "start_s": "1:23" (wrong format!)

================================
STEP 5: GENERATE JSON OUTPUT
================================

Now output EXACTLY {num_questions} questions in JSON array format:

[
  {
    "question_index": 0,
    "question": "After [anchor description], when does [target description] happen?",
    "temporal_relation": "after|once_finished|next|during|before",
    "anchor_event": "Brief description of anchor",
    "target_event": "Brief description of target",
    "answer": {
      "start_s": 78.5,
      "end_s": 82.0
    },
    "requires_audio": true,
    "confidence": 0.9,
    "abstained": false,
    "rationale_model": "E1 (anchor) at 65.0s -> E2 (target) at 78.5s, 
                        relation=after, target spans 78.5-82.0s. 
                        All times in seconds."
  },
  ...
]

REQUIRED FIELDS (do not add or remove fields):
- question_index: 0, 1, 2, ... (integers starting from 0)
- question: Natural language question in English
- temporal_relation: Must be one of: after, once_finished, next, during, before
- anchor_event: One sentence describing the anchor
- target_event: One sentence describing the target
- answer.start_s: Decimal seconds when target BEGINS (or null if abstained)
- answer.end_s: Decimal seconds when target COMPLETES/ENDS (or null if abstained)
  CRITICAL: end_s must capture when the event FINISHES, not just when it starts
  For speech events: end when the complete sentence/thought finishes
  For visual events: end when element disappears or transitions away
  For actions: end when the action fully completes
- requires_audio: true if audio is needed to answer, false if purely visual
- confidence: Float 0.0-1.0 indicating your certainty
- abstained: true only if target event does not exist in video OR events are 
  too temporally close to disambiguate
- rationale_model: Detailed explanation with timestamps in decimal seconds
  - Keep rationale concise (<= 80 words)
  - Refer to event IDs E1, E2, ... to make the sequence clear
  - Always restate anchor time and target time (both start AND end)

ABSTENTION RULES:
- If anchor exists but target does NOT exist -> Set abstained=true, answer times=null
- If temporal relationship cannot be determined -> Set abstained=true, answer times=null
- If two plausible target moments are closer than 0.4 seconds -> Set abstained=true, 
  answer times=null
- Provide clear explanation in rationale_model

CRITICAL REQUIREMENTS:
[OK] Generate EXACTLY {num_questions} question objects
[OK] Use ONLY decimal seconds (5.2, 78.0, 125.5) - NEVER MM:SS format
[OK] Convert any MM:SS thinking to seconds: (M x 60 + S)
[OK] All times must be between 0.0 and {duration}
[OK] Return ONLY the JSON array (no markdown, no extra text)
[OK] Do NOT fabricate events that don't exist in the video
[OK] Include detailed rationale with specific timestamps
[OK] Keep rationale concise (<= 80 words) and refer to E1, E2, ...

[X] Do NOT use MM:SS format in answer fields (like 1:18 or 2:30)
[X] Do NOT concatenate minutes and seconds (78 is NOT "one eighteen")
[X] Do NOT output markdown code blocks or explanations
[X] Do NOT create questions where anchor=target
[X] Do NOT skip required fields

Now begin your annotation process following ALL five steps above.
Think carefully about timestamps - convert any MM:SS to decimal seconds!
Output your JSON array:
\end{lstlisting}
\end{tcolorbox}

\subsection{Task-Specific Annotation Procedures}
\label{app:task-procedures}

\paragraph{Task 1: Video summarization.} This task evaluates comprehension of the full recording. For videos under 10 minutes, we summarize the full audio–visual input and captions directly. Longer videos are split into 10-minute segments (with a shorter final segment); each segment is summarized independently and merged during the reducing phase into (i) a narrative summary and (ii) a bullet-point summary. Summaries emphasize causal structure, key actions, and outcomes. Any model-generated drafts are edited or rewritten by humans for correctness and style.

\paragraph{Task 2: Multiple-Choice Questions (MCQs).} For closed-ended evaluation, videos longer than 3 minutes are segmented with 30-second overlaps to preserve context. For each segment, we generate one MCQ with four answer choices plus a fifth option (\textit{``Not enough evidence''}) for cases where the segment does not provide sufficient information. 
Each MCQ is accompanied by a brief rationale that explicitly references visual and auditory cues supporting the correct answer. Questions are required to target higher-order reasoning across six categories: \textit{integration} (connecting visual and auditory cues), \textit{correlation} (relating demonstrated actions to spoken explanations), \textit{inference} (drawing conclusions from multimodal evidence), \textit{guided analysis}, \textit{application}, and \textit{cause-and-effect}. Simple recall is explicitly prohibited, and each question must draw on at least two information sources across modalities. Correct answers are distributed approximately uniformly across all five option positions (A--E: 20.2\%, 20.1\%, 20.1\%, 19.9\%, 19.6\%; minor deviation from 100\% due to rounding), confirming no positional bias in the benchmark.

\textbf{Task 3: Temporal localization.} This task evaluates temporal grounding and event ordering. Using the same segmentation as the MCQ task, we construct multiple questions per segment based on an \textit{anchor event} 
and a \textit{target event} occurring before, after, during, or immediately following the anchor. During annotation, human annotators watch each segment independently and provide timestamps \textit{relative to the segment start} 
(0.0s to segment duration), which then converted to absolute time by adding the segment start time to the relative predictions giving the ground truth annotations. The dataset contains 3,392 instances, with up to three questions per segment, each annotated with the correct relation, temporal interval, and a supporting rationale. This enables evaluation of both localization accuracy and temporal reasoning quality.

\textbf{Ethical and licensing compliance.} All datasets comply with their original licenses (e.g., Creative Commons Attribution 4.0 International (CC BY 4.0), research-only use) and respect data provenance and API terms. Personally identifiable or sensitive content is excluded wherever it cannot be ethically anonymized. Our process aligns with responsible AI and data governance standards (e.g., EU AI Act and NIST AI RMF principles).

\section{Team and Annotation / Review Guidelines}
\label{app:team}

\subsection{Team Formation}
\label{app:team_formation}

The annotation team consisted of 5 domain experts with strong English proficiency and research experience in vision-language learning: 2 PhD holders, 2 Master's students, and 1 research scientist. The team composition was gender-balanced (2 male, 3 female) and geographically diverse, with members from East Asia, Middle East, North America, and South Asia. All annotators completed task-specific training before beginning annotation work.
The annotation workflow followed a hierarchical review structure: Master's students conducted initial annotations, which were reviewed by PhD holders, with the research scientist providing final oversight and quality control. The team held weekly calibration meetings to discuss edge cases, resolve annotation disagreements, and ensure consistency across the dataset.

When disagreements arose, they were escalated to the weekly calibration meetings and resolved synchronously through discussion among domain experts. Annotators compared their rationales against the labeling guidelines and the available evidence from the multimodal inputs (video, audio, and captions). We explicitly solicited perspectives from multiple domain experts to reduce individual bias and clarify ambiguous cases. Decisions were made by consensus when possible; otherwise we adopted a majority vote among the reviewing domain experts, with the research scientist serving as the final tie-breaker and recording the final decision to maintain consistency.

\subsection{Guidelines}
\label{app:guidelines}

To ensure annotation quality and consistency, annotators followed these core principles:

\begin{tcolorbox}[
    colback=gray!5!white,
    colframe=gray!75!black,
    title=Annotation Quality Checklist,
    fonttitle=\bfseries,
    boxrule=0.5mm,
    arc=2mm,
    left=3mm,
    right=3mm,
    top=1mm,
    bottom=1mm,
    toptitle=1mm,
    bottomtitle=1mm
]
\small  
\textbf{Before Starting Annotation:}
\begin{itemize}[nosep,leftmargin=*,topsep=0pt,partopsep=0pt]
    \item[$\square$] Specify the topic being annotated (e.g., Topic 2: Job Interviews)
    \item[$\square$] Specify the task being reviewed (Task 1: Summarization / Task 2: MCQ / Task 3: Temporal Localization)
    \item[$\square$] Watch the entire video fully before annotation
    \item[$\square$] Review any available captions or transcripts for context
\end{itemize}

\textbf{During Annotation:}
\begin{itemize}[nosep,leftmargin=*,topsep=0pt,partopsep=0pt]
    \item[$\square$] Ground all questions, answers, and rationales in observable audio-visual content
    \item[$\square$] Flag cases with insufficient evidence rather than making assumptions
    \item[$\square$] Assign demographic attributes based on visual and auditory presentation in the video
    \item[$\square$] Revisit any timestamp as needed to verify accuracy
    \item[$\square$] Discuss ambiguous cases with team leads before finalizing
\end{itemize}

\textbf{Quality Control:}
\begin{itemize}[nosep,leftmargin=*,topsep=0pt,partopsep=0pt]
    \item[$\square$] Ensure all required fields are completed (question, answer, rationale, timestamps where applicable)
    \item[$\square$] Verify that rationales explicitly reference audio-visual evidence
    \item[$\square$] Check for internal consistency within annotations you reviewed
\end{itemize}

\textbf{After Review:}
\begin{itemize}[nosep,leftmargin=*,topsep=0pt,partopsep=0pt]
    \item[$\square$] Address reviewer feedback and revise annotations as needed
    \item[$\square$] Participate in weekly calibration meetings to discuss edge cases
    \item[$\square$] Update annotations based on refined guidelines from team discussions
\end{itemize}
\end{tcolorbox}
For task-specific protocols (segmentation, timestamps, rationale structure), see Section~\ref{app:task-procedures}.

\subsection{Review Interface}
\label{app:interface}

We developed a custom web-based review interface to facilitate efficient human verification and correction of AI-generated annotations. The interface supports all three annotation tasks and provides real-time video playback, synchronized captions, and structured editing fields.

\begin{figure*}[h]
    \centering
    \includegraphics[width=1.0\textwidth]{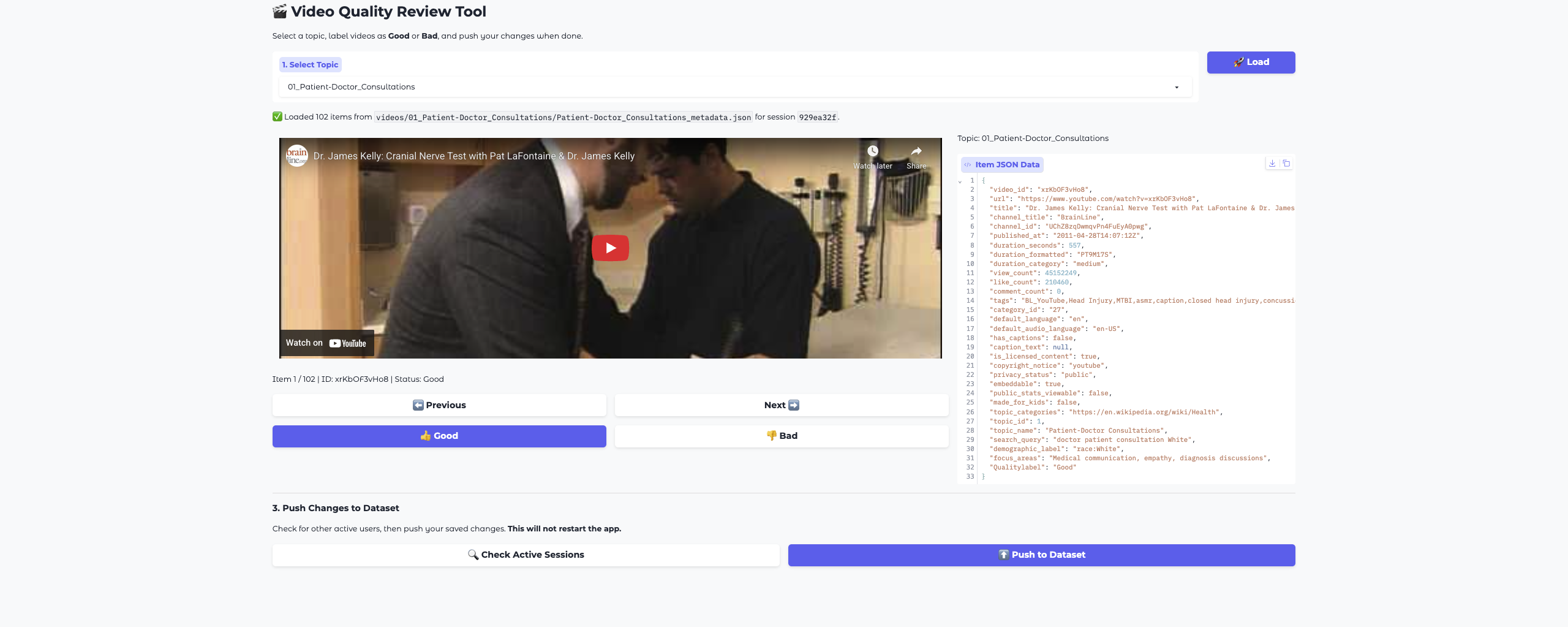}
    \caption{Quality filtering interface for initial video selection. Reviewers watch candidate videos and label them as ``Good'' or ``Bad'' based on inclusion criteria (audio-visual quality, topic relevance, demographic coverage). The interface displays video metadata (duration, topic category, licensing) and allows batch processing with session management.}
    \label{fig:quality_review}
\end{figure*}

\paragraph{Quality filtering interface.}
Figure~\ref{fig:quality_review} shows the initial quality filtering interface used during video selection. Reviewers assess each candidate video for topic relevance, audio-visual quality, and demographic coverage, marking videos as either ``Good'' or ``Bad'' for inclusion in the dataset.

\begin{figure*}[h]
    \centering
    \includegraphics[width=1.0\textwidth]{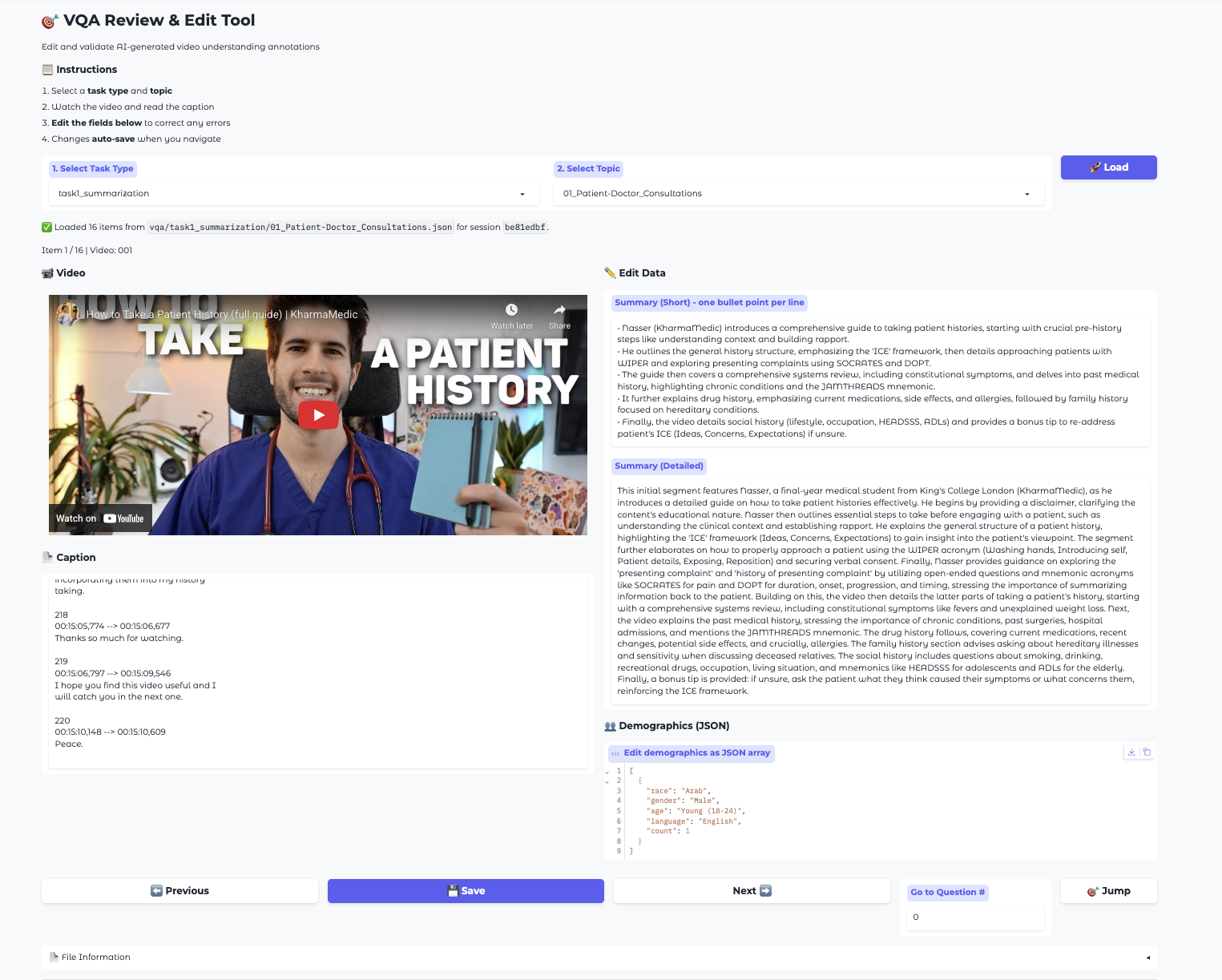}
    \caption{Review interface for Task 1 (Video Summarization). Annotators edit AI-generated summaries (both detailed and bullet-point formats) while watching the full video with synchronized captions. The demographics panel allows verification and correction of demographic labels in JSON format. Changes auto-save when navigating between items.}
    \label{fig:summary_review}
\end{figure*}

\begin{figure*}[h]
    \centering
    \includegraphics[width=1.0\textwidth]{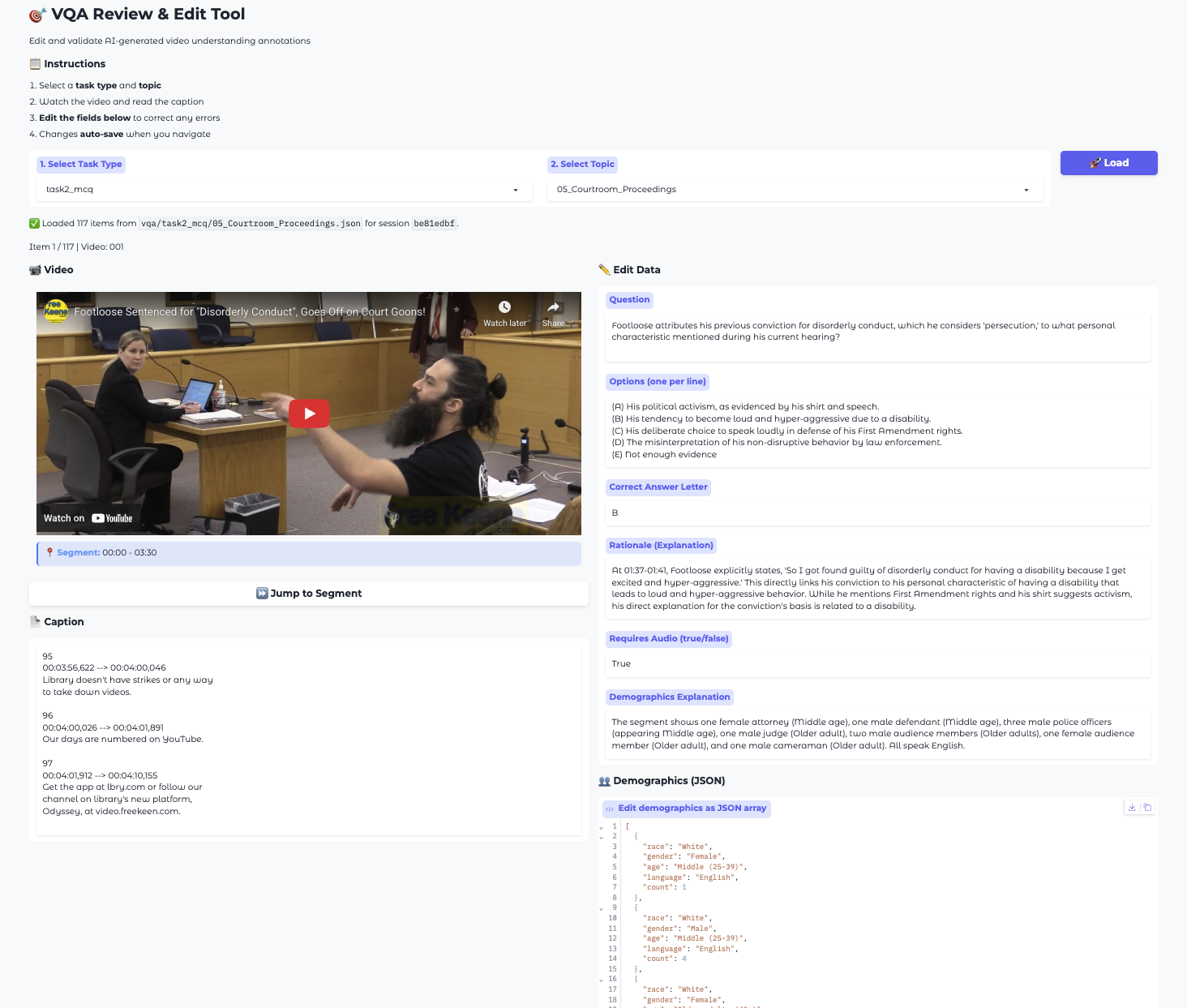}
    \caption{Review interface for Task 2 (Multiple-Choice Questions). Annotators review and edit MCQ questions, answer options (A--E), correct answer labels, and rationales that reference visual and auditory evidence. The segment indicator shows the current 3-minute video segment being annotated.}
    \label{fig:mcq_review}
\end{figure*}

\begin{figure*}[h]
    \centering
    \includegraphics[width=1.0\textwidth]{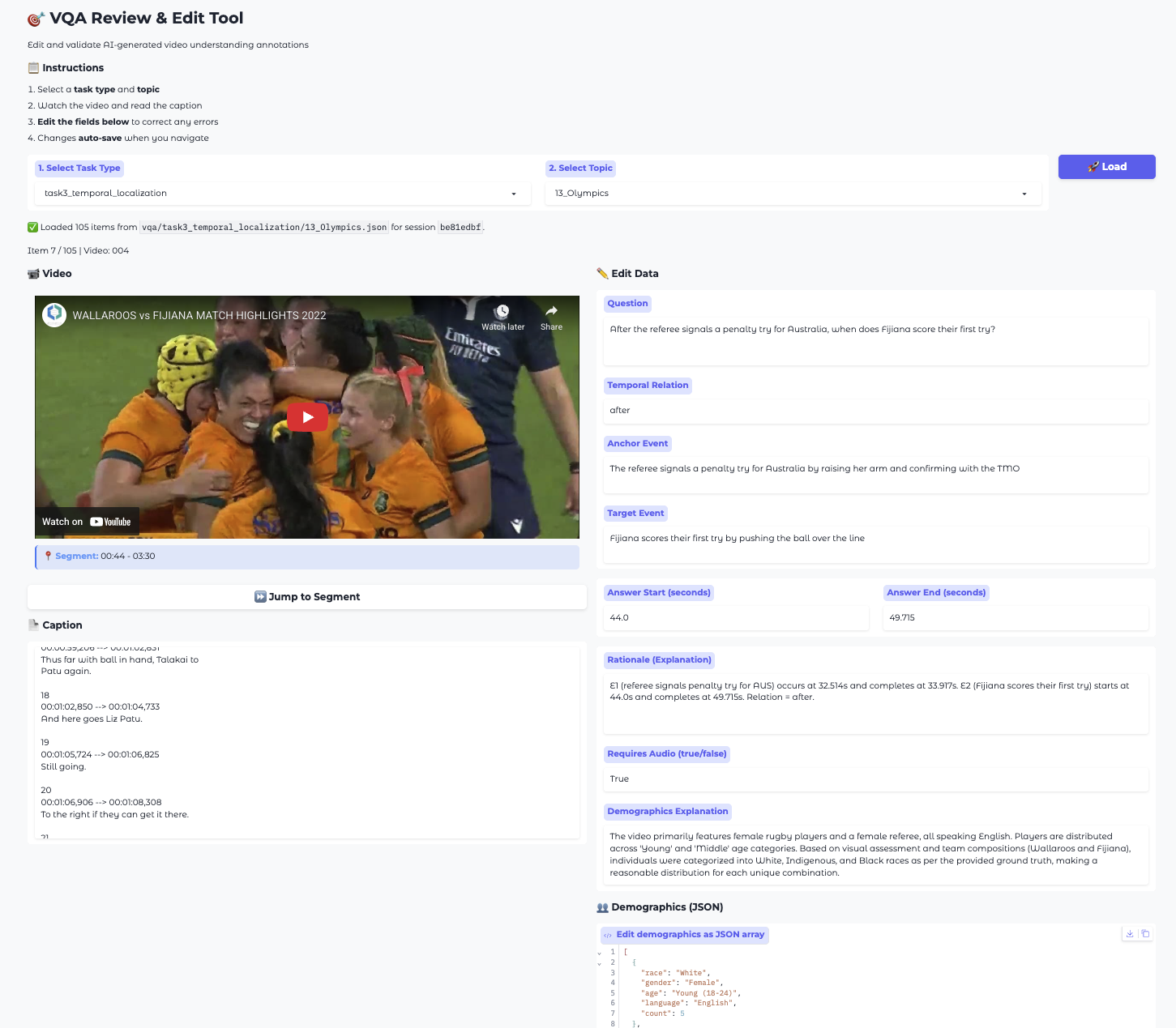}
    \caption{Review interface for Task 3 (Temporal Localization). Annotators edit anchor-target event pairs, temporal relations (before/after/during/immediately), answer start/end timestamps, rationales explaining the temporal reasoning, and audio requirements. The segment timeline helps annotators identify precise event boundaries within 3-minute video segments.}
    \label{fig:temporal_review}
\end{figure*}

\paragraph{Task-specific review interfaces.}
Figures~\ref{fig:summary_review}, \ref{fig:mcq_review}, and \ref{fig:temporal_review} illustrate the task-specific interfaces for editing and validating annotations across the three tasks. Each interface provides:
\begin{itemize}
    \item \textbf{Video player}: Embedded YouTube player with timestamp controls and caption display
    \item \textbf{Edit fields}: Structured input fields for questions, answers, rationales, and timestamps
    \item \textbf{Demographics panel}: JSON editor for verifying and correcting demographic labels with per-person annotations
    \item \textbf{Navigation controls}: Previous/Next buttons, jump functionality, and session management
    \item \textbf{Auto-save}: Changes persist automatically when navigating between items
\end{itemize}

The interface ensures high annotation quality by requiring annotators to watch full videos, providing easy access to AI-generated drafts for comparison, and supporting rapid iteration across large annotation batches.

\begin{figure*}[h]
    \centering
    \begin{subfigure}[b]{0.48\textwidth}
        \centering
        \includegraphics[width=\textwidth]{AVQA_Figures/plot_duration_category_by_topic_videos.png}
        \caption{Video count by duration category per topic}
        \label{fig:duration_by_topic}
    \end{subfigure}
    \hfill
    \begin{subfigure}[b]{0.48\textwidth}
        \centering
        \includegraphics[width=\textwidth]{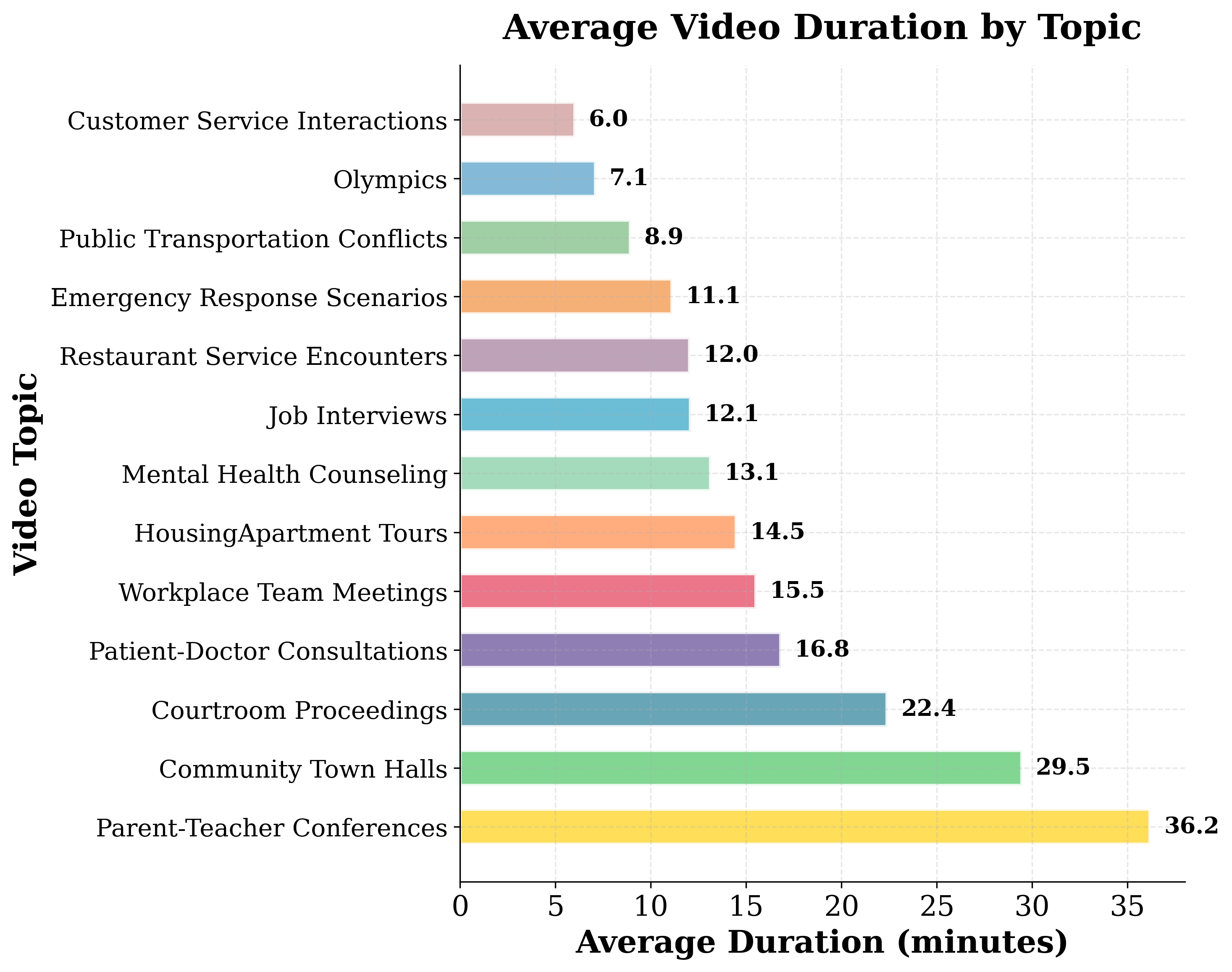}
        \caption{Average video duration per topic (minutes)}
        \label{fig:avg_duration_by_topic}
    \end{subfigure}
    \caption{Video duration analysis by topic. (a) Distribution showing the count of short (orange), medium (teal), and long (dark teal) videos within each of the 13 topics. Housing/Apartment Tours contains the most videos (24), while Workplace Team Meetings has the fewest (12). (b) Average duration per topic, with Parent-Teacher Conferences having the longest average (36.2 min) and Customer Service Interactions the shortest (6.0 min).}
    \label{fig:duration_analysis}
\end{figure*}

\section{Extended Dataset Statistics}
\label{app:dataset-viz}

\subsection{Demographic Distribution}
\label{app:demo-dist}

\begin{figure*}[t]
    \centering
    \begin{subfigure}[b]{0.32\textwidth}
        \centering
        \includegraphics[width=\textwidth]{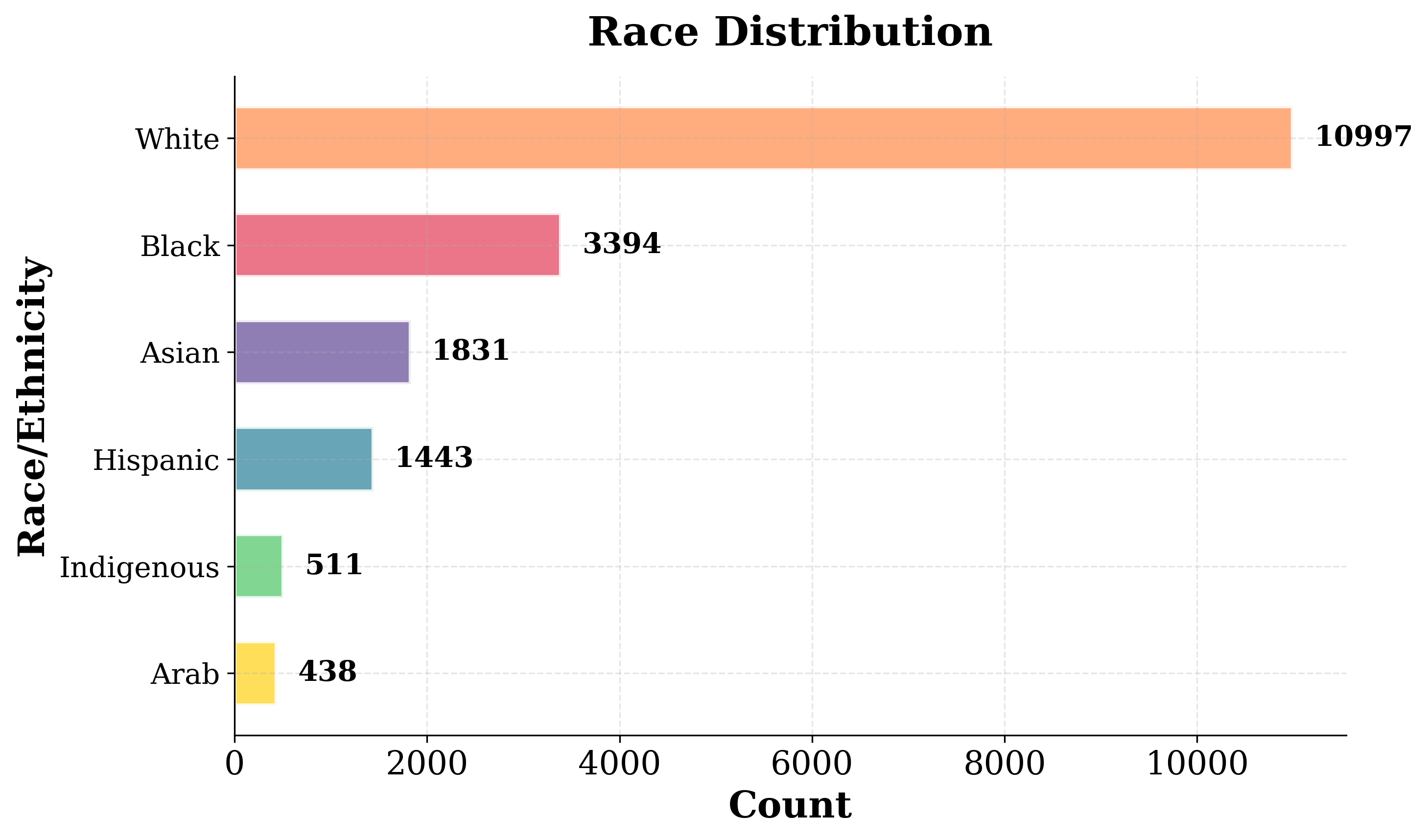}
        \caption{Distribution by race}
        \label{fig:qa_race}
    \end{subfigure}
    \hfill
    \begin{subfigure}[b]{0.32\textwidth}
        \centering
        \includegraphics[width=\textwidth]{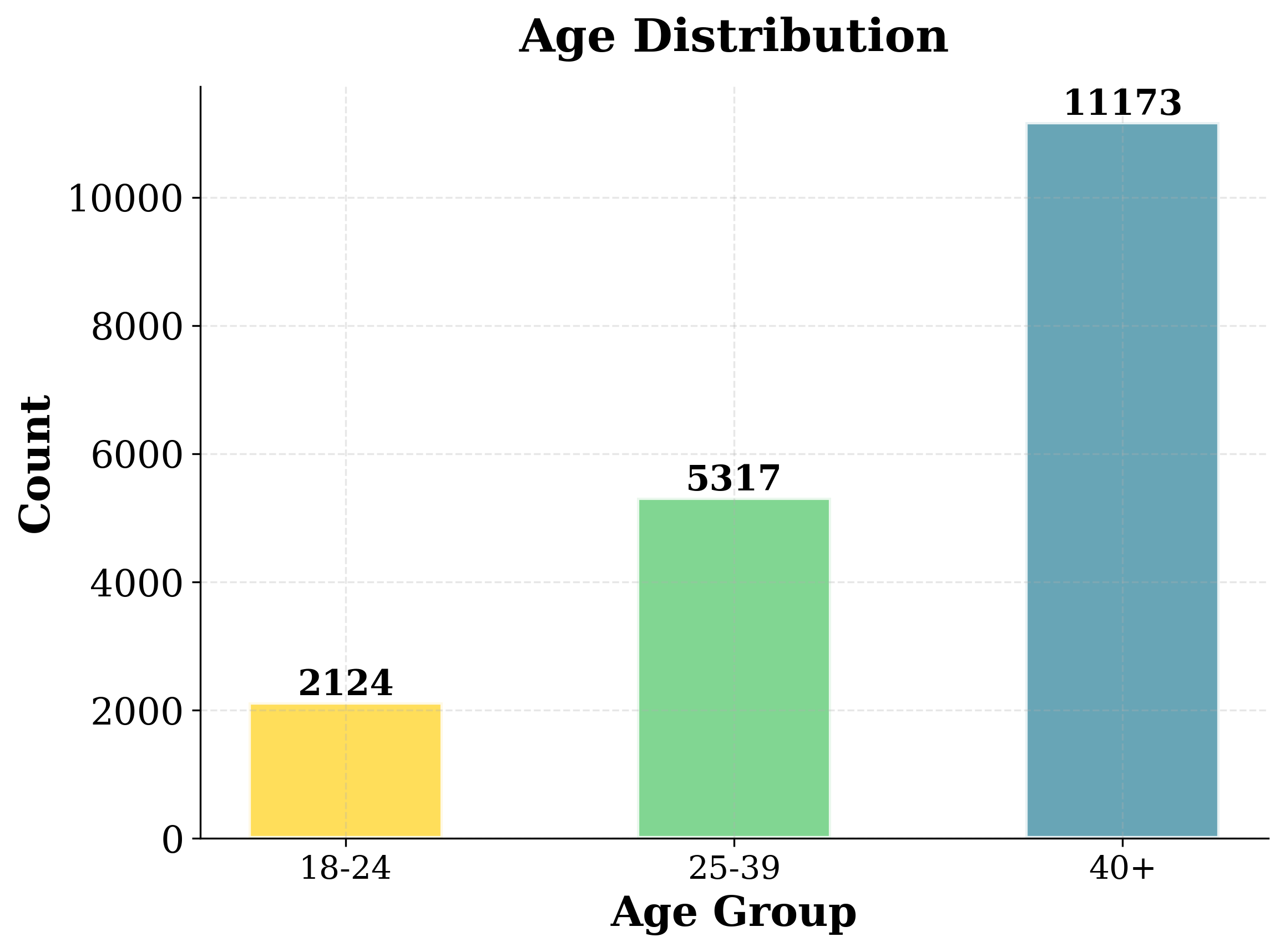}
        \caption{Distribution by age group}
        \label{fig:qa_age}
    \end{subfigure}
    \hfill
    \begin{subfigure}[b]{0.32\textwidth}
        \centering
        \includegraphics[width=\textwidth]{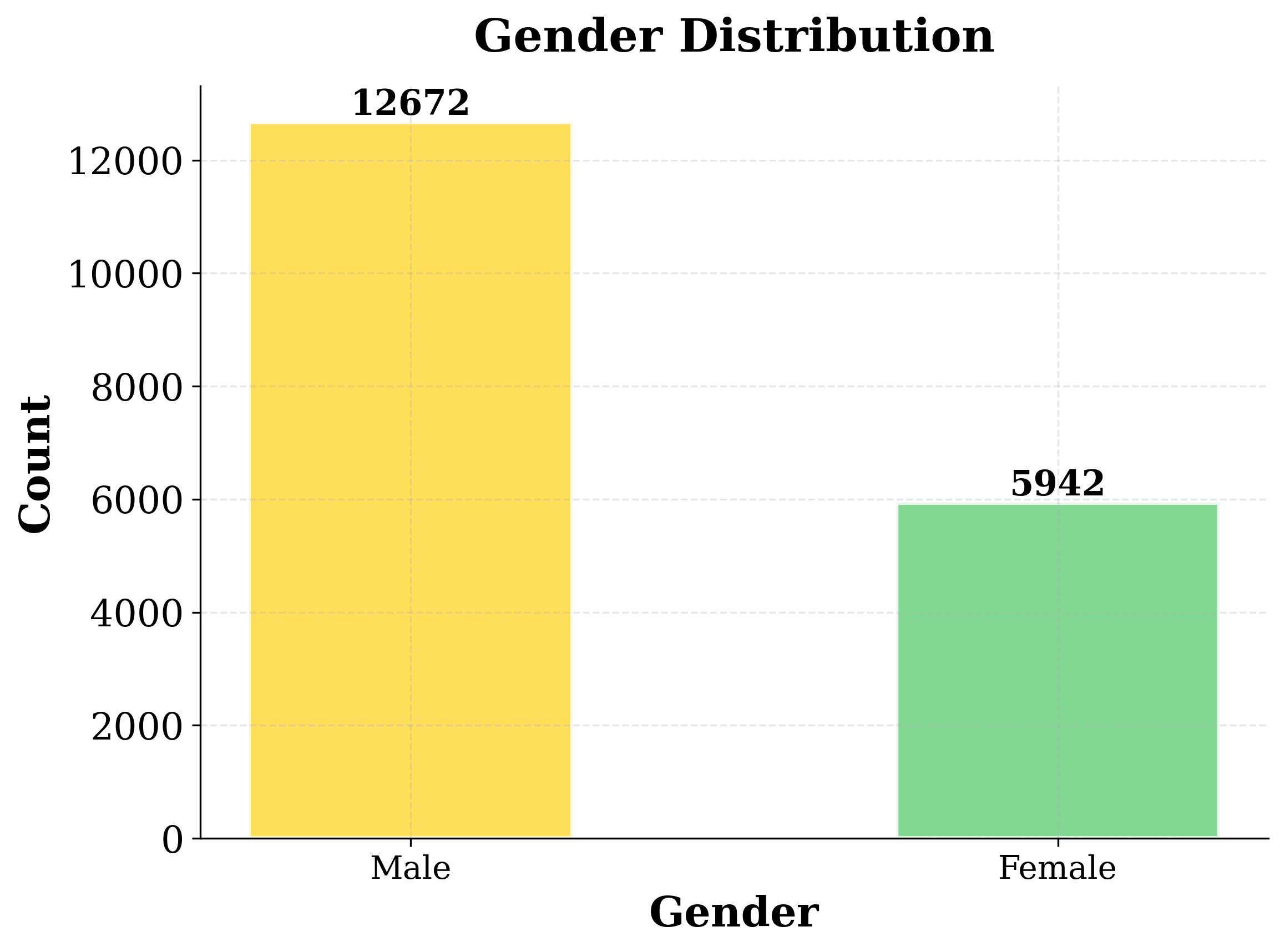}
        \caption{Distribution by gender}
        \label{fig:qa_gender}
    \end{subfigure}
    \caption{Demographic distribution of benchmark instance level (a) race/ethnicity, (b) age group, and (c) gender. Race distribution shows: White (10,997), Black (3,494), Asian (1,831), Hispanic (1,443), Indigenous (511), Arab (438). Age distribution shows: 18-24 (2,124), 25-39 (5,317), 40+ (11,173). Gender distribution shows: Male (12,672), Female (5,942).}
    \label{fig:qa_demographics}
\end{figure*}

Figure~\ref{fig:qa_demographics} reports the demographic distribution \emph{at the benchmark instance level}. Counts reflect how often demographic groups appear across all benchmark instances (Task~1 chunks and Task~2--3 segments), using per-instance demographic counts/mappings, rather than the number of \textbf{unique individuals} or \textbf{unique videos}. As a result, the same individuals may be counted multiple times across different segments or tasks when the underlying content is reused.

This aggregated view should be interpreted as \textbf{demographic exposure frequency} in the benchmark. It is appropriate for evaluating model behavior, since models are evaluated on these instances, but it does not represent a census of unique participants.

We observe moderate skew toward male-presenting speakers and participants aged 40+, which likely reflects the underlying distribution of English-language CC BY 4.0 YouTube content in our target domains. Topics with large groups (e.g., community town halls and courtroom proceedings) further increase exposure counts because many individuals co-occur and persist across multiple segments.

\subsection{Video Duration Statistics}
\label{app:duration-stats}

Figures~\ref{fig:duration_by_topic}, and~\ref{fig:avg_duration_by_topic} illustrate the distribution of video durations across the three length categories (Short: $<$5~minutes, Medium: 5-20~minutes, Long: 20-60~minutes) and by topic.

\section{Preprocessing and Inference Configuration}
\label{app:preprocess}

\subsection{Hardware-Aware Inference Constraints}
To rigorously assess open-source MLLMs, we utilize a standardized compute environment of \textbf{4$\times$ NVIDIA A40 GPUs (40GB VRAM each)}. This configuration allows for substantial parallelization but imposes hard memory limits on long-context processing. A notable exception is \textbf{MiniCPM-o-2.6}, which we evaluate on a single \textbf{NVIDIA A100 (80GB)} GPU, as the official implementation does not currently support distributed inference. These hardware constraints necessitate a strict input capping and fallback policy to prevent Out-Of-Memory (OOM) failures while maximizing the multimodal context available to each model.

\subsection{Visual and Audio Input Strategy}
We adopt a unified prompting strategy, concatenating the textual prompt, audio input, and sampled video frames, strictly adhering to each model's official implementation. 
\begin{itemize}
    \item \textbf{Modality Fusion:} Most models ingest separated modalities (frames + audio). However, \textbf{VideoLLaMA 2} is an exception, requiring the raw video file with embedded audio rather than discrete inputs.
    \item \textbf{Visual Capping:} We cap the initial maximum visual input at $F_{\max}=256$ frames. This limit was empirically determined to fit within the memory constraints of our 4$\times$A40 setup for the majority of architectures.
\end{itemize}

\subsection{Robustness and Adaptive Fallback Protocols}

To ensure evaluation fairness across diverse architectures, we enforce a consistent, dynamic fallback protocol for models that encounter Out-Of-Memory (OOM) errors or context overflow. We initiate inference with each model's designated maximum frame capacity: 256 frames for Qwen3-Omni, UniMoE-2.0, OLA, and MiniCPM-o-2.6; 128 frames for VideoLLaMA2; 64 frames for VITA 1.5; and 32 frames for Baichuan-Omni 1.5. These represent upper bounds; shorter videos naturally use fewer frames. All models initially receive full unchunked audio. Lower frame caps for Baichuan-Omni 1.5, VITA 1.5, and VideoLLaMA2 reflect their smaller context windows and more limited memory footprints under our hardware constraints. Upon failure, we apply a synchronized reduction strategy:

\begin{enumerate}
    \item \textbf{Visual Reduction:} The frame budget is iteratively halved from each model's maximum. For example, Qwen3-Omni reduces through the sequence \{256, 128, 64, 32\}, allowing up to four retry attempts.
    
    \item \textbf{Audio Reduction:} Simultaneously, we adjust audio fidelity through uniform chunk sampling rather than truncation. If the full audio track causes failure, we transition to a chunked representation with a maximum limit of $N=64$ chunks. Each chunk spans 10 seconds of non-overlapping audio. In subsequent retries, this chunk limit halves in lockstep with visual frames through the sequence \{64, 32, 16\}. Given audio of duration $D$ seconds, the natural chunk count is $M = \lceil D / 10 \rceil$. When $M \leq N$, all chunks are retained; when $M > N$, we uniformly sample $N$ chunks using $\text{indices} = \lfloor \text{linspace}(0, M-1, N) \rfloor$ to preserve temporal coverage across the full recording. For Ola and Baichuan-Omni 1.5, we use 30-second chunks instead of 10-second chunks, consistent with their Whisper-based audio encoders which are trained on 30-second windows.
\end{enumerate}

We adopted this audio chunking because not all MLLMs handle long audio tracks, as such they employ naive audio handling strategies such as end-truncation or middle-cropping, introducing uncontrolled confounds in evaluation. Our uniform chunk sampling approach is inspired by MiniCPM-o-2.6~\cite{minicpmo26}. 

This ``start-high, fall-back'' approach ensures that reported performance reflects each model's maximum capability under standard hardware constraints (4$\times$ NVIDIA A40 GPUs, 40GB VRAM each or 1$\times$ NVIDIA A100, 80GB VRAM).

\subsection{Task-Specific Prompts}
\label{sec:task-prompts}
We report the shared task instruction text (T1--T3) used across models. For open-source MLLMs, we rely on each model’s default chat template and required modality tokens/placeholders (e.g., UniMoE-2.0’s \texttt{<video>}/\texttt{<audio>} markers; VITA 1.5’s image/audio special tokens) as implemented in their official codebases.

\textbf{Task 1: Video Summarization.}
\label{sec:prompt-t1}
\begin{tcolorbox}[breakable, colback=blue!5!white, colframe=blue!50!black, 
                  title=Task 1: Standard Prompt, fontupper=\small]
\begin{lstlisting}[breaklines=true, basicstyle=\ttfamily\scriptsize, columns=fullflexible]
You are analyzing a video that is {video_duration} seconds long.

Please provide a comprehensive analysis with the following components:

1. DETAILED SUMMARY: Write a detailed paragraph (150-250 words) that 
   captures the main content, key points, and flow of the video.

2. SHORT SUMMARY: Provide 3-5 concise bullet points highlighting the 
   most important takeaways.

3. TIMELINE: Create a timeline of major sections/topics in the video with:
   - start: timestamp in "HH:MM:SS" format
   - end: timestamp in "HH:MM:SS" format
   - title: brief section title
   - note: one sentence description

4. GLOSSARY: List 5-15 key terms, acronyms, or important concepts 
   mentioned in the video.

5. CONFIDENCE: Rate your confidence in this analysis from 0.0 to 1.0.

Return your response as a JSON object with this exact structure:
{
  "summary_detailed": "detailed paragraph here",
  "summary_short": ["bullet 1", "bullet 2", "bullet 3"],
  "timeline": [
    {
      "start": "00:00:00",
      "end": "00:02:30",
      "title": "Section Title",
      "note": "Brief description"
    }
  ],
  "glossary": ["term1", "term2", "term3"],
  "confidence": 0.95
}

Only return the JSON object, no additional text.
\end{lstlisting}
\end{tcolorbox}

\begin{tcolorbox}[breakable, colback=green!5!white, colframe=green!50!black, 
                  title=Task 1: Empathy Prompt, fontupper=\small]
\begin{lstlisting}[breaklines=true, basicstyle=\ttfamily\scriptsize, columns=fullflexible]
You are analyzing a video that is {video_duration} seconds long.

Please provide an empathetic and emotionally aware analysis. Focus on 
understanding the emotional context, interpersonal dynamics, and human 
elements in the content.

Provide:

1. DETAILED SUMMARY: Write a detailed, empathetic paragraph (150-250 words) 
   that captures not just what happens, but the emotional tone, interpersonal 
   dynamics, and human aspects. Consider how participants might feel and 
   what underlying concerns or needs are being expressed.

2. CONFIDENCE: Rate your confidence in this analysis from 0.0 to 1.0.

Return your response as a JSON object:
{
  "summary_detailed_empathic": "empathetic detailed paragraph here",
  "confidence": 0.95
}

Only return the JSON object, no additional text.
\end{lstlisting}
\end{tcolorbox}

\textbf{Task 2: Multiple-Choice Question Answering.}
\label{sec:prompt-t2}

\begin{tcolorbox}[breakable, colback=orange!5!white, colframe=orange!50!black, 
                  title=Task 2 Prompt, fontupper=\small]
\begin{lstlisting}[breaklines=true, basicstyle=\ttfamily\scriptsize, columns=fullflexible]
You are analyzing a video segment to answer a multiple choice question.

QUESTION:
{question}

OPTIONS:
{options_text}

Please analyze the video content carefully and:
1. Select the best answer from the options
2. Provide a clear rationale explaining why this answer is correct based 
   on what you observed in the video
3. Rate your confidence in this answer from 0.0 to 1.0

Return your response as a JSON object with this exact structure:
{
  "answer_letter": "B",
  "answer_index": 1,
  "rationale": "Explanation of why this answer is correct based on video 
                content",
  "confidence": 0.90
}

Notes:
- answer_letter should be "A", "B", "C", "D", or "E"
- answer_index should be 0, 1, 2, 3, or 4 (corresponding to A, B, C, D, E)
- rationale should reference specific details from the video
- CRITICAL: include all fields in the JSON and ensure proper formatting 
  especially answer_letter

Only return the JSON object, no additional text.
\end{lstlisting}
\end{tcolorbox}

\textbf{Task 3: Temporal Localization.}
\label{sec:prompt-t3}

For temporal localization inference, models receive absolute segment boundaries and must predict timestamps in absolute video time. For example, when evaluating a 3-minute segment spanning 500--680 seconds of a 30-minute video, the model is 
told ``You are analyzing a video segment from 500s to 680s'' and must return  predictions within [500s, 680s], not [0s, 180s].

This design ensures models maintain temporal awareness of their position within the full video, which is critical for long-form understanding. In contrast, during annotation (Section~\ref{app:task-procedures}), human annotators worked with segment-relative timestamps (0.0s to segment duration) for cognitive ease when watching isolated clips. These segment-relative annotations were automatically converted to absolute video time by adding the segment start offset, producing ground truth in the same coordinate system used during inference. This alignment between annotation conversion and inference format ensures fair evaluation: models are tested on their ability to reason about events in absolute video time, matching real-world deployment scenarios where systems must index events within full-length recordings.

The inference prompt below specifies the absolute segment range and constraints:

\begin{tcolorbox}[breakable, colback=purple!5!white, colframe=purple!50!black, 
                  title=Task 3 Prompt, fontupper=\small]
\begin{lstlisting}[breaklines=true, basicstyle=\ttfamily\scriptsize, columns=fullflexible]
You are analyzing a video segment from {segment_start}s to {segment_end}s 
(duration: {segment_duration}s).

QUESTIONS ({num_questions} total):
{questions_text}

REQUIRED OUTPUT FORMAT:
{
  "questions": [
    {
      "question_id": "001",
      "start_s": <timestamp_float>,
      "end_s": <timestamp_float>,
      "confidence": <float_between_0_and_1>,
      "rationale_model": "<your explanation>"
    },
    {
      "question_id": "002",
      "start_s": <timestamp_float>,
      "end_s": <timestamp_float>,
      "confidence": <float_between_0_and_1>,
      "rationale_model": "<your explanation>"
    }
    ... (continue for all {num_questions} questions)
  ]
}

RATIONALE REQUIREMENTS:
Explain: When E1 (anchor) occurs -> When E2 (target) starts/ends -> 
Temporal relationship -> Visual/audio cues

Example: "E1 (anchor) starts at 5.2s with speaker's introduction. E2 
(target) starts at 35.0s when he says 'I am a final year medical student', 
ends at 36.6s. Relationship: 'after'."

CONSTRAINTS:
- All timestamps within [{segment_start}s, {segment_end}s]
- start_s < end_s for each question
- Include ALL {num_questions} questions with their question_id field
- CRITICAL: include all fields in the JSON and ensure proper formatting
- CRITICAL: Your response MUST include these exact question_ids: 
  {question_ids_list} and include the question_id field
- Return ONLY valid JSON (no markdown, no code blocks, no extra text)
\end{lstlisting}
\end{tcolorbox}

\subsection{Per-Model Effective Settings}
\label{app:model_settings}
Table~\ref{tab:model_settings} details the effective frame budget, audio handling, and specific hardware allocation for each evaluated model.

\begin{table*}[t]
\centering
\caption{Effective inference settings and preprocessing strategies. \textbf{Checkpoint} refers to the specific HuggingFace model ID used. \textbf{Native} indicates the model's default preprocessing; \textbf{Adaptive} denotes non-native audio chunking applied as a fallback mechanism.}
\label{tab:model_settings}
\scriptsize
\centering
\setlength{\tabcolsep}{3pt}
\begin{tabular}{p{4cm}p{1.8cm}p{1.2cm}p{1.6cm}p{1.8cm}p{4.0cm}}
\toprule
\textbf{Model / Checkpoint} & \textbf{Backbone} & \textbf{Frames} & \textbf{Audio} & \textbf{Hardware} & \textbf{Notes} \\
\midrule
\texttt{VideoLLaMA2.1-7B-AV}& Qwen2-7B & 128 & Embedded & 4$\times$A40 & Integrated video+audio input. \\
\addlinespace[2pt]
\texttt{Qwen3-Omni-30B-A3B-Instruct}& Qwen3-MoE & 256 & Adaptive$^*$ & 4$\times$A40 & Chunking triggered by length overflow. \\
\addlinespace[2pt]
\texttt{UniMoE-2.0-Omni}& Qwen2.5-7B & 256 & Full & 4$\times$A40 & Native processing. \\
\addlinespace[2pt]
\texttt{MiniCPM-o-2.6} & Qwen2.5-7B & 256 & Full & 1$\times$A100 & Native processing. \\
\addlinespace[2pt]
\texttt{baichuan-inc/Baichuan-Omni-1d5} & Qwen2.5-7B & 32 & Adaptive$^*$ & 1$\times$A100 & Chunking triggered by length overflow. \\
\addlinespace[2pt]
\texttt{THUdyh/Ola-7b} & Qwen2.5-7B & 256 & Adaptive$^*$ & 1$\times$A100 & Chunking triggered by length overflow. \\
\addlinespace[2pt]
\texttt{VITA-1.5} & Qwen2-7B & 64 & Adaptive$^*$ & 4$\times$A40 & Chunking triggered by length overflow. \\
\addlinespace[2pt]
\textbf{Gemini 3.0 Pro} & Proprietary & 1 FPS & Full & API & Native processing. \\
\addlinespace[2pt]
\textbf{GPT-4o} & Proprietary & 256 & Captions$^\dagger$ & API & No native audio support; ASR captions used. \\
\bottomrule
\end{tabular}
\vspace{0.3em}
\\
\raggedright
{\footnotesize $^*$Audio Chunking triggered $^\dagger$Fallback to text modality. All models use temp=0.7, top-p=0.95, max tokens=8192.}
\end{table*}

\section{Evaluation Metrics}
\label{app:appendix_metrics}

This section provides formal definitions of all evaluation metrics used in our benchmark. Metrics are computed internally in the range $[0,1]$ but are reported as percentages (scaled by 100) where indicated. Text similarity metrics remain in the $[0,1]$ range following standard conventions.

\subsection{Task 1: Video Summarization Metrics}
\label{app:t1_metrics}

For Task 1 (Video Summarization), we evaluate both detailed summaries and short bullet-point summaries using the following metrics:

\paragraph{ROUGE-L.}
We compute ROUGE-L (Longest Common Subsequence) F1-score~\citep{lin2004rouge} to measure n-gram overlap between generated and reference summaries. For a reference summary $R$ and predicted summary $P$:

\begin{equation}
\text{ROUGE-L}(R, P) = \frac{2 \cdot \text{Prec}_{\text{LCS}} \cdot \text{Rec}_{\text{LCS}}}{\text{Prec}_{\text{LCS}} + \text{Rec}_{\text{LCS}}} \times 100
\end{equation}

where $\text{Prec}_{\text{LCS}}$ and $\text{Rec}_{\text{LCS}}$ are precision and recall based on the longest common subsequence. The score is reported as a percentage in $[0, 100]$.

\paragraph{Text Similarity.}
We measure semantic similarity using cosine similarity between sentence embeddings. Given embeddings $\mathbf{e}_R$ and $\mathbf{e}_P$ from a pre-trained model (all-MiniLM-L6-v2):

\begin{equation}
\text{TextSim}(R, P) = \frac{\mathbf{e}_R \cdot \mathbf{e}_P}{\|\mathbf{e}_R\| \|\mathbf{e}_P\|}
\end{equation}

This metric is reported in $[0, 1]$, following standard cosine similarity conventions.

\paragraph{Aggregation.}
For a dataset with $N$ videos, we report the mean metric across all samples:

\begin{equation}
\overline{\text{ROUGE-L}} = \frac{1}{N} \sum_{i=1}^{N} \text{ROUGE-L}(R_i, P_i)
\end{equation}

We compute these metrics separately for detailed summaries and short summaries.

\subsection{Task 2: Multiple-Choice Question Answering Metrics}
\label{app:t2_metrics}

For Task 2 (MCQ Question Answering), we evaluate both answer accuracy and rationale quality.

\paragraph{Answer Accuracy.}
For each question $q_i$ with ground truth answer $a_i^{\text{gt}}$ and predicted answer $a_i^{\text{pred}}$:

\begin{equation}
\text{Correct}(q_i) = \begin{cases}
1 & \text{if } a_i^{\text{pred}} = a_i^{\text{gt}} \\
0 & \text{otherwise}
\end{cases}
\end{equation}

The overall accuracy across $N$ questions is:

\begin{equation}
\text{Accuracy} = \frac{1}{N} \sum_{i=1}^{N} \text{Correct}(q_i) \times 100
\end{equation}

reported as a percentage in $[0, 100]$.

\paragraph{Rationale Metrics.}
For each question, we evaluate the quality of the model's rationale (explanation) using:
\begin{itemize}
    \item \textbf{ROUGE-L}: Measures lexical overlap with reference rationale (reported as percentage, $[0, 100]$)
    \item \textbf{Text Similarity}: Semantic similarity via cosine distance (reported in $[0, 1]$)
\end{itemize}

We aggregate rationale metrics by computing the mean across all questions with valid rationales.

\subsection{Task 3: Temporal Localization Metrics}
\label{app:t3_metrics}

For Task 3 (Temporal Localization), we evaluate the model's ability to identify temporal intervals that answer questions about video segments.

\paragraph{Temporal Intersection over Union (tIoU).}
For a ground truth temporal interval $G = [s^{\text{gt}}, e^{\text{gt}}]$ and predicted interval $P = [s^{\text{pred}}, e^{\text{pred}}]$, where $s$ and $e$ denote start and end times in seconds:

\begin{multline}
\text{tIoU}(G, P) = \\
\frac{
\max\!\left(0,\; \min(e^{\text{gt}}, e^{\text{pred}}) - \max(s^{\text{gt}}, s^{\text{pred}})\right)
}{
\max(e^{\text{gt}}, e^{\text{pred}}) - \min(s^{\text{gt}}, s^{\text{pred}})
}
\times 100
\end{multline}

The tIoU measures temporal overlap and is reported as a percentage in $[0, 100]$.

\paragraph{Mean Intersection over Union (mIoU).}
The mean IoU aggregates tIoU scores across all $N$ temporal localization questions:

\begin{equation}
\text{mIoU} = \frac{1}{N} \sum_{i=1}^{N} \text{tIoU}(G_i, P_i)
\end{equation}

This metric is reported as a percentage in $[0, 100]$.

\paragraph{Recall at IoU Threshold (Recall@$\theta$).}
We measure the percentage of predictions whose tIoU meets or exceeds a threshold $\theta$:

\begin{equation}
\text{Recall@}\theta =
\frac{1}{N} \sum_{i=1}^{N}
\mathbb{1}\!\left[\text{tIoU}(G_i, P_i) \ge 100,\theta\right]
\cdot 100
\end{equation}

where $\mathbb{1}[\cdot]$ is the indicator function. We report Recall@0.3, Recall@0.5, and Recall@0.7 as percentages in $[0, 100]$. We abbreviate Recall@$\theta$ as R@$\theta$ in tables.

\paragraph{Mean Absolute Error (MAE).}
We compute the temporal localization error for start and end times:

\begin{align}
\text{MAE}_{\text{start}} &= \frac{1}{N} \sum_{i=1}^{N} |s_i^{\text{gt}} - s_i^{\text{pred}}| \\
\text{MAE}_{\text{end}} &= \frac{1}{N} \sum_{i=1}^{N} |e_i^{\text{gt}} - e_i^{\text{pred}}| \\
\text{MAE}_{\text{avg}} &= \frac{\text{MAE}_{\text{start}} + \text{MAE}_{\text{end}}}{2}
\end{align}

MAE is reported in seconds.

\paragraph{Rationale Metrics.}
Similar to Task 2, we evaluate rationale quality using ROUGE-L (percentage) and Text Similarity ($[0,1]$).

\subsection{LLM-as-Judge Evaluation}
\label{app:llm_judge}

For tasks involving free-form text generation (summaries, rationales), we employ an LLM-as-Judge evaluation protocol adopted from previous work to assess semantic correctness and quality beyond n-gram overlap metrics~\cite{ataallah2024infinibench}. We use GPT-5-mini~\citep{gpt5mini} as our primary judge.

\paragraph{Evaluation Protocol.}
For each generated output, the judge model:
\begin{enumerate}
    \item Receives the original question and both the human-verified reference and predicted answers
    \item Evaluates semantic similarity, factual correctness, and completeness
    \item Assigns a numerical score from 0 (completely incorrect) to 10 (perfect match)
    \item Provides a concise justification (1-3 sentences) for the assigned score
\end{enumerate}

The judge is instructed to accept paraphrases, synonyms, or rephrasings as valid when they preserve the original meaning, and to penalize omissions of key factual elements, hallucinated content, or contradictions. The evaluation focuses on semantic alignment rather than surface-level textual similarity.
\paragraph{Scoring Scale and Output Format.}
Judges assign integer scores from 0 to 10 and return responses in JSON format:
\begin{lstlisting}[breaklines=true, basicstyle=\ttfamily\scriptsize, columns=fullflexible]
{
  "score": <integer 0-10>,
  "justification": "<brief explanation>"
}
\end{lstlisting}
We aggregate judge scores by computing the mean across all evaluated samples. This provides a complementary assessment to automatic metrics (ROUGE-L, cosine similarity), particularly for capturing semantic equivalence that lexical overlap metrics may miss.

\paragraph{Judge Prompt.}
The complete system prompt provided to the judge model is shown in Figure~\ref{fig:judge_prompt}. The prompt emphasizes fair evaluation, human-like judgment, and consistent scoring criteria across all evaluated outputs.

\begin{figure}[h]
\centering
\small
\begin{tcolorbox}[
  colback=gray!5, colframe=gray!40,
  width=\linewidth,
  boxrule=0.5pt,
  left=4pt, right=4pt
]

\textbf{System Prompt:} You are an intelligent and fair evaluator AI that specializes in assessing the correctness and semantic alignment between ground truth answers and predicted responses for question-answering tasks, including those based on video content.

\textbf{Evaluation Instructions:}
\begin{itemize}[leftmargin=*, nosep]
    \item Focus on \textbf{semantic similarity}, \textbf{factual correctness}, and \textbf{completeness}
    \item Accept paraphrases, synonyms, or rephrasings \textbf{as valid}, as long as they preserve the original meaning
    \item \textbf{Do not penalize} for stylistic differences or changes in tone, unless they impact factual accuracy
    \item \textbf{Penalize} if:
    \begin{itemize}[leftmargin=*, nosep]
        \item The predicted answer omits \textbf{key factual elements} present in the correct answer
        \item The prediction includes \textbf{hallucinated content} or unfounded details
        \item The prediction \textbf{contradicts} the correct answer
    \end{itemize}
    \item Use human-like judgment: apply reasoning beyond surface text similarity
    \item When uncertain, provide a \textbf{conservative but fair} score
    \item Use a scoring scale from \textbf{0 (completely incorrect)} to \textbf{10 (perfect match)}
\end{itemize}

\textbf{Output Format:} Return a JSON object with two fields:

\begin{Verbatim}[fontsize=\scriptsize]
{ "score": <integer 0-10>,
  "justification": "<concise explanation (1-3 sentences)>" }
\end{Verbatim}

\textbf{User Prompt Template:}
Please evaluate the following video-based question-answer pair:

\textbf{Question:} \{question\}\\
\textbf{Correct Answer:} \{correct\_answer\}\\
\textbf{Predicted Answer:} \{predicted\_answer\}

Please return your evaluation in the specified JSON format with both a score and a justification.
\end{tcolorbox}
\caption{\textbf{LLM-as-Judge evaluation prompt.} The system prompt instructs the judge to evaluate semantic correctness, factual completeness, and relevance on a 0--10 scale, with specific guidelines for handling paraphrases, hallucinations, and contradictions.}
\label{fig:judge_prompt}
\end{figure}

\subsection{Aggregation Across Topics}
\label{app:aggregation}

Our benchmark spans multiple topics (content domains). For each task, we first compute metrics within each topic, then aggregate across topics by computing the mean metric value. This approach provides a single aggregate score while accounting for performance variation across different content types. Due to space constraints in tables, we report only mean values; per-topic breakdowns are available in our supplementary materials.

\begin{table*}[t]
\centering
\caption{\textbf{Comprehensive benchmark results across all tasks and metrics.} Results are reported for MLLMs on all dataset. Summarization reports judge scores (0--10), ROUGE-L (\%), and similarity (0--1).
MCQ reports accuracy (\%) and rationale quality (score 0--10, ROUGE-L \%, similarity 0--1). Temporal localization reports mIoU (\%), R@0.3/R@0.5 (\%), MAE in seconds, and rationale quality.
\textbf{Bold} indicates best performance, \underline{underline} indicates second-best.}
\label{tab:comprehensive_results}
\small
\setlength{\tabcolsep}{4pt}
\resizebox{\linewidth}{!}{%
\begin{tabular}{@{}l | c c c | c c c c | c c c c c c c@{}}
\toprule
& \multicolumn{3}{c|}{Summarization}
& \multicolumn{4}{c|}{MCQ}
& \multicolumn{7}{c}{Temporal Localization} \\
\cmidrule(lr){2-4} \cmidrule(lr){5-8} \cmidrule(lr){9-15}
Model & Score & RG-L & Sim
& Acc. & Score & RG-L & Sim
& mIoU & R@0.3 & R@0.5 & MAE & Score & RG-L & Sim \\
\midrule
Gemini 3.0 Pro$^\dagger$
& \textbf{7.07} & \textbf{27.2} & \textbf{0.813}
& \textbf{81.4} & \textbf{8.71} & \underline{19.6} & \textbf{0.706}
& \textbf{26.6} & \textbf{39.3} & \textbf{25.4} & \textbf{12.0}
& \textbf{5.38} & \textbf{28.3} & \underline{0.673} \\
Qwen3-Omni
& \underline{5.72} & \underline{22.8} & \underline{0.713}
& \underline{65.7} & \underline{7.47} & \textbf{20.3} & \underline{0.699}
& \underline{3.70} & \underline{5.14} & \underline{2.81} & \underline{60.0}
& 2.58 & \underline{28.0} & \textbf{0.698} \\
UniMoE-2.0
& 4.71 & 20.8 & 0.700
& 51.6 & 6.17 & 13.1 & 0.627
& 1.81 & 2.40 & 1.04 & 685.8
& 2.11 & 23.7 & 0.549 \\
MiniCPM-o-2.6
& 3.34 & 14.7 & 0.563
& 51.9 & 6.15 & 10.3 & 0.600
& 1.81 & 2.38 & 0.73 & 367.1
& 3.65 & 19.3 & 0.406 \\
Baichuan-Omni 1.5
& 3.68 & 18.8 & 0.600
& 56.2 & 6.18 & 15.1 & 0.651
& 2.75 & 2.96 & 1.07 & 158.6
& 2.29 & 18.2 & 0.433 \\
OLA
& 4.42 & 19.0 & 0.638
& 53.2 & 6.61 & 14.0 & 0.642
& 3.26 & 3.38 & 1.24 & 270.8
& 3.48 & 21.8 & 0.439 \\
VITA 1.5
& 2.77 & 16.3 & 0.491
& 51.1 & 6.04 & 13.2 & 0.625
& 1.81 & 2.55 & 1.15 & 116.1
& \underline{3.91} & 22.1 & 0.433 \\
VideoLLaMA2
& 1.53 & 12.7 & 0.383
& 26.4 & 4.17 & 9.3 & 0.530
& 3.50 & 1.68 & 0.42 & 234.5
& 2.22 & 22.5 & 0.493 \\
\bottomrule
\end{tabular}
}
\end{table*}
\subsection{Implementation Details}
\label{app:implementation}

\paragraph{Libraries.}
We use the following libraries for metric computation:
\begin{itemize}
    \item \textbf{ROUGE}: \texttt{rouge-score} package with Porter stemming
    \item \textbf{Text Embeddings}: \texttt{sentence-transformers} with \texttt{all-MiniLM-L6-v2} model
    \item \textbf{LLM Judge}: OpenAI API (GPT-5-mini) 
\end{itemize}

\paragraph{Handling Failed Predictions.}
If a model fails to generate a prediction for a sample (e.g., due to timeout or error), that sample is excluded from metric computation. We report the number of successfully evaluated samples for each task.

\paragraph{Reproducibility.}
All metric computation code and detailed per-topic results are available in our code repository. For statistical comparison between models, we recommend paired bootstrap tests with $n=10,000$ resamples and $\alpha=0.05$ significance level.

\section{Detailed Results}
\label{app:detailed_results}

Table~\ref{tab:comprehensive_results} presents the complete benchmark results across all three tasks, including all automatic metrics and LLM-as-judge rationale scores. We report detailed summarization performance, MCQ accuracy with rationale quality, and temporal localization metrics including mIoU, recall at multiple thresholds, mean absolute error, and rationale quality.

\textbf{Key findings.} Gemini 3.0 Pro demonstrates strong performance across all tasks, achieving the best scores in summarization (7.07), MCQ accuracy (81.4\%), and temporal localization (26.6\% mIoU, 25.4\% R@0.5). Qwen3-Omni achieves second-best performance on most metrics, with particularly strong rationale quality (28.0\% ROUGE-L, 0.698 similarity on temporal task). Among open-source models, Qwen3-Omni shows the best summarization capabilities, while VITA 1.5 produces the highest-quality temporal rationales (3.91 judge score).

The temporal localization task reveals a critical challenge for open-source models: all models except Gemini exhibit large MAE values (60.0--685.8 seconds), indicating severe localization errors, which for UniMoE-2.0 largely reflect temporal reference frame hallucination as discussed in Appendix~\ref{app:temporal_error_analysis}. This failure mode occurs when models process long videos in segments but fail to ground predictions in absolute timestamps, instead treating each segment as starting at time 0.0 seconds. Despite this localization failure, models can still produce reasonable rationales (scores 2.11--3.91), demonstrating a disconnect between semantic understanding and temporal grounding.

Rationale quality metrics show interesting patterns: while Gemini produces the best overall rationales for summarization and MCQ tasks, Qwen3-Omni achieves competitive or superior similarity scores on temporal localization (0.698), and VITA 1.5 receives the highest judge scores for temporal rationales among open-source models (3.91). This suggests that rationale quality does not perfectly correlate with task performance, and that different models may excel at explanation versus execution.

\section{Detailed Results Per-Topic}
\label{app:per-topic-detailed}
This appendix provides comprehensive per-topic performance breakdowns for all three tasks. Each table reports metrics for individual topics (T1--T13), allowing detailed analysis of model capabilities across different conversational domains. All metrics follow the same conventions as the main paper: percentages for ROUGE-L, Accuracy, mIoU, and Recall; a 0--1 scale for cosine similarity; a 0--10 scale for LLM judge scores; and seconds for MAE. Bold denotes best performance, underline denotes second-best within each topic. $^\dagger$ denotes closed-source model.

\begin{table*}[!htbp]
\centering
\caption{\textbf{Task 1: Detailed Summarization Performance of MLLMs on Topics 1--4.} Judge Score (0--10), RG-L: ROUGE-L (\%), and Sim: Similarity (0--1) shown for detailed summaries. \textbf{Bold} = best, \underline{underline} = second-best per topic.}
\label{tab:sum-topics-group1}
\small
\resizebox{\textwidth}{!}{%
\begin{tabular}{lcccccccccccc}
\toprule
\textbf{Model} & \multicolumn{3}{c}{\textbf{T1: Patient-Dr.}} & \multicolumn{3}{c}{\textbf{T2: Job Int.}} & \multicolumn{3}{c}{\textbf{T3: Parent-Tch.}} & \multicolumn{3}{c}{\textbf{T4: Customer}} \\
\cmidrule(lr){2-4} \cmidrule(lr){5-7} \cmidrule(lr){8-10} \cmidrule(lr){11-13}
 & Score & RG-L & Sim & Score & RG-L & Sim & Score & RG-L & Sim & Score & RG-L & Sim \\
\midrule
Gemini 3.0 Pro$^\dagger$ & \textbf{8.12} & \textbf{29.7} & \textbf{0.858} & \textbf{8.33} & \textbf{26.5} & \textbf{0.817} & \textbf{7.94} & \textbf{27.6} & \textbf{0.822} & \textbf{7.67} & \textbf{29.1} & \textbf{0.814} \\
Qwen3-Omni         & \underline{7.38} & \underline{24.6} & \underline{0.785} & \underline{7.14} & \underline{25.5} & \underline{0.755} & \underline{6.72} & \underline{22.2} & \underline{0.744} & \underline{6.13} & \underline{25.6} & \underline{0.743} \\
UniMoE-2.0         & 6.12 & 23.2 & 0.764 & 6.76 & 21.0 & 0.751 & 5.72 & 20.1 & 0.726 & 4.33 & 20.6 & 0.690 \\
MiniCPM-o-2.6      & 4.94 & 15.8 & 0.695 & 4.19 & 15.5 & 0.587 & 4.28 & 13.8 & 0.602 & 3.20 & 14.7 & 0.591 \\
Baichuan-Omni 1.5  & 5.81 & 22.5 & 0.746 & 5.38 & 20.7 & 0.665 & 4.44 & 18.9 & 0.644 & 2.73 & 18.5 & 0.589 \\
OLA                & 5.50 & 20.7 & 0.723 & 5.90 & 19.9 & 0.700 & 5.22 & 17.7 & 0.685 & 2.93 & 18.6 & 0.629 \\
VITA 1.5           & 4.44 & 18.3 & 0.631 & 3.71 & 17.2 & 0.535 & 4.22 & 16.1 & 0.543 & 1.73 & 17.2 & 0.507 \\
VideoLLaMA2        & 2.31 & 13.6 & 0.485 & 2.19 & 11.9 & 0.431 & 2.28 & 11.9 & 0.384 & 1.27 & 12.3 & 0.434 \\
\bottomrule
\end{tabular}}
\end{table*}

\begin{table*}[!htbp]
\centering
\caption{\textbf{Task 1: Detailed Summarization Performance of MLLMs on Topics 5--8.} Judge Score (0--10), RG-L: ROUGE-L (\%), and Sim: Similarity (0--1) shown for detailed summaries. \textbf{Bold} = best, \underline{underline} = second-best per topic.}
\label{tab:sum-topics-group2}
\small
\resizebox{\textwidth}{!}{%
\begin{tabular}{lcccccccccccc}
\toprule
\textbf{Model} & \multicolumn{3}{c}{\textbf{T5: Courtroom}} & \multicolumn{3}{c}{\textbf{T6: Emergency}} & \multicolumn{3}{c}{\textbf{T7: Transport}} & \multicolumn{3}{c}{\textbf{T8: Workplace}} \\
\cmidrule(lr){2-4} \cmidrule(lr){5-7} \cmidrule(lr){8-10} \cmidrule(lr){11-13}
 & Score & RG-L & Sim & Score & RG-L & Sim & Score & RG-L & Sim & Score & RG-L & Sim \\
\midrule
Gemini 3.0 Pro$^\dagger$ & \textbf{6.54} & \textbf{25.1} & \textbf{0.792} & \textbf{6.25} & \textbf{27.6} & \textbf{0.771} & \textbf{7.43} & \textbf{28.1} & \textbf{0.852} & \textbf{7.00} & \textbf{26.7} & \textbf{0.827} \\
Qwen3-Omni         & \underline{6.00} & \underline{22.8} & \underline{0.703} & \underline{5.10} & \underline{22.8} & \underline{0.686} & \underline{4.86} & \underline{21.2} & \underline{0.702} & \underline{6.08} & \underline{20.8} & \underline{0.721} \\
UniMoE-2.0         & 4.08 & 21.1 & 0.667 & 3.90 & 20.4 & 0.640 & 3.29 & 19.5 & 0.687 & 4.75 & 19.2 & 0.689 \\
MiniCPM-o-2.6      & 3.15 & 14.2 & 0.486 & 3.35 & 16.0 & 0.579 & 2.21 & 14.2 & 0.523 & 3.17 & 13.4 & 0.508 \\
Baichuan-Omni 1.5  & 3.31 & 16.8 & 0.571 & 3.35 & 17.5 & 0.563 & 2.14 & 18.2 & 0.566 & 4.75 & 18.2 & 0.643 \\
OLA                & 4.15 & 19.4 & 0.636 & 3.85 & 20.3 & 0.624 & 3.14 & 16.3 & 0.571 & 5.67 & {20.6} & 0.671 \\
VITA 1.5           & 1.77 & 14.5 & 0.407 & 2.70 & 16.8 & 0.520 & 2.00 & 15.6 & 0.527 & 2.50 & 14.2 & 0.431 \\
VideoLLaMA2        & 0.85 & 11.7 & 0.268 & 1.65 & 15.4 & 0.459 & 1.00 & 13.6 & 0.452 & 1.17 & 11.0 & 0.306 \\
\bottomrule
\end{tabular}}
\end{table*}

\begin{table*}[!htbp]
\centering
\caption{\textbf{Task 1: Detailed Summarization Performance of MLLMs on Topics 9--13.} Score: Judge Score (0--10), RG-L: ROUGE-L (\%), and Sim: Similarity (0--1) shown for detailed summaries. \textbf{Bold} = best, \underline{underline} = second-best per topic.}
\label{tab:sum-topics-group3}
\resizebox{\textwidth}{!}{%
\begin{tabular}{lccccccccccccccc}
\toprule
\textbf{Model} & \multicolumn{3}{c}{\textbf{T9: Housing}} & \multicolumn{3}{c}{\textbf{T10: Restaurant}} & \multicolumn{3}{c}{\textbf{T11: Mental Hlth}} & \multicolumn{3}{c}{\textbf{T12: Town Halls}} & \multicolumn{3}{c}{\textbf{T13: Olympics}} \\
\cmidrule(lr){2-4} \cmidrule(lr){5-7} \cmidrule(lr){8-10} \cmidrule(lr){11-13} \cmidrule(lr){14-16}
 & Score & RG-L & Sim & Score & RG-L & Sim & Score & RG-L & Sim & Score & RG-L & Sim & Score & RG-L & Sim \\
\midrule
Gemini 3.0 Pro$^\dagger$ & \textbf{6.58} & \textbf{27.7} & \textbf{0.789} & \textbf{5.91} & \textbf{27.1} & \textbf{0.786} & \textbf{8.08} & \textbf{28.4} & \textbf{0.830} & \textbf{7.00} & \textbf{26.0} & \textbf{0.791} & \textbf{5.04} & \textbf{23.7} & \textbf{0.822} \\
Qwen3-Omni        & \underline{4.58} & \underline{23.0} & \underline{0.671} & \underline{5.00} & \underline{23.2} & \underline{0.721} & \underline{7.23} & \underline{26.3} & 0.773 & \underline{4.72} & 18.8 & 0.599 & \underline{3.39} & 19.2 & 0.671 \\
UniMoE-2.0        & 4.38 & 22.0 & 0.668 & 4.91 & 19.8 & 0.674 & 6.69 & 25.2 & \underline{0.791} & 4.33 & \underline{19.8} & \underline{0.647} & 1.96 & 18.7 & \underline{0.702} \\
MiniCPM-o-2.6     & 2.88 & 14.9 & 0.497 & 3.13 & 15.3 & 0.596 & 4.46 & 15.6 & 0.631 & 2.06 & 11.7 & 0.390 & 2.43 & 16.5 & 0.631 \\
Baichuan-Omni 1.5 & 2.67 & 18.5 & 0.490 & 3.09 & 18.1 & 0.569 & 5.23 & 21.7 & 0.688 & 2.83 & 16.5 & 0.464 & 2.09 & 18.5 & 0.607 \\
OLA               & 4.33 & 20.2 & 0.618 & 3.74 & 18.9 & 0.595 & 6.38 & 21.9 & 0.719 & 4.28 & 17.1 & 0.552 & 2.39 & 16.0 & 0.565 \\
VITA 1.5          & 2.58 & 17.0 & 0.427 & 2.26 & 17.2 & 0.464 & 3.69 & 17.6 & 0.554 & 2.28 & 14.1 & 0.343 & 2.09 & 16.6 & 0.498 \\
VideoLLaMA2       & 1.50 & 14.1 & 0.354 & 1.70 & 13.8 & 0.421 & 2.08 & 13.3 & 0.424 & 0.83 & 9.8  & 0.175 & 1.13 & 12.9 & 0.390 \\
\bottomrule
\end{tabular}}
\end{table*}

\begin{table*}[!htbp]
\centering
\caption{\textbf{Task 2: MCQ Performance of MLLMs on Topics 1--4.} Acc.: Accuracy (\%), Score: Rationale Judge Score (0--10), RG-L: ROUGE-L (\%), and Sim: Similarity (0--1) shown. \textbf{Bold} = best, \underline{underline} = second-best per topic.}
\label{tab:mcq-topics-group1}
\resizebox{\textwidth}{!}{%
\begin{tabular}{lcccccccccccccccc}
\toprule
\textbf{Model} & \multicolumn{4}{c}{\textbf{T1: Patient-Dr.}} & \multicolumn{4}{c}{\textbf{T2: Job Int.}} & \multicolumn{4}{c}{\textbf{T3: Parent-Tch.}} & \multicolumn{4}{c}{\textbf{T4: Customer}} \\
\cmidrule(lr){2-5} \cmidrule(lr){6-9} \cmidrule(lr){10-13} \cmidrule(lr){14-17}
 & Acc. & Score & RG-L & Sim & Acc. & Score & RG-L & Sim  & Acc. & Score & RG-L & Sim  & Acc. & Score & RG-L & Sim  \\
\midrule
Gemini 3.0 Pro$^\dagger$ & \textbf{85.3} & \textbf{9.16} & \underline{19.4} & \textbf{0.721} & \textbf{85.0} & \textbf{9.09} & \underline{20.8} & \textbf{0.713} & \textbf{80.9} & \textbf{8.88} & \underline{20.4} & \textbf{0.707} & \textbf{89.7} & \textbf{8.82} & 19.0 & \textbf{0.725} \\
Qwen3-Omni & \underline{75.5} & \underline{8.25} & \textbf{20.7} & \underline{0.720} & \underline{77.0} & \underline{8.24} & \textbf{21.3} & \underline{0.701} & \underline{68.3} & \underline{8.13} & \textbf{21.7} & \underline{0.695} & 64.1 & \underline{6.95} & \underline{20.0} & \underline{0.703} \\
UniMoE-2.0 & 63.7 & 7.56 & 13.5 & 0.655 & 60.0 & 7.27 & 14.0 & 0.629 & 53.5 & 6.59 & 14.0 & 0.638 & 53.8 & 5.33 & 13.3 & 0.621 \\
MiniCPM-o-2.6 & 60.8 & 6.57 & 10.9 & 0.626 & 57.0 & 6.57 & 11.4 & 0.600 & 50.4 & 6.45 & 11.5 & 0.615 & 48.7 & 5.54 & 9.7 & 0.588 \\
Baichuan-Omni 1.5 & 69.6 & 7.18 & 15.2 & 0.660 & 69.0 & 7.21 & 15.3 & 0.642 & 61.3 & 6.80 & 15.3 & 0.650 & 56.4 & 5.85 & 15.6 & 0.656 \\
OLA & 63.7 & 7.68 & 14.3 & 0.659 & 55.0 & 6.89 & 13.7 & 0.622 & 55.2 & 6.90 & 14.2 & 0.637 & 48.7 & 5.59 & 14.3 & 0.665 \\
VITA 1.5 & 49.0 & 6.28 & 12.0 & 0.634 & 56.0 & 6.39 & 12.2 & 0.603 & 50.4 & 6.37 & 12.3 & 0.610 & \underline{74.4} & 6.92 & \textbf{25.7} & \textbf{0.725} \\
VideoLLaMA2 & 34.3 & 4.75 & 9.6 & 0.561 & 27.0 & 4.67 & 10.1 & 0.525 & 25.7 & 4.18 & 9.9 & 0.542 & 30.8 & 3.79 & 10.6 & 0.556 \\
\bottomrule
\end{tabular}
}
\end{table*}

\begin{table*}[!htbp]
\centering
\caption{\textbf{Task 2: MCQ Performance of MLLMs on Topics 5--8.} Acc.: Accuracy (\%), Score: Rationale Judge Score (0--10), RG-L: ROUGE-L (\%), and Sim: Similarity (0--1) shown. \textbf{Bold} = best, \underline{underline} = second-best per topic.}
\label{tab:mcq-topics-group2}
\resizebox{\textwidth}{!}{%
\begin{tabular}{lcccccccccccccccc}
\toprule
\textbf{Model} & \multicolumn{4}{c}{\textbf{T5: Courtroom}} & \multicolumn{4}{c}{\textbf{T6: Emergency}} & \multicolumn{4}{c}{\textbf{T7: Transport}} & \multicolumn{4}{c}{\textbf{T8: Workplace}} \\
\cmidrule(lr){2-5} \cmidrule(lr){6-9} \cmidrule(lr){10-13} \cmidrule(lr){14-17}
 & Acc. & Score & RG-L & Sim & Acc. & Score & RG-L & Sim  & Acc. & Score & RG-L & Sim  & Acc. & Score & RG-L & Sim \\
\midrule
Gemini 3.0 Pro$^\dagger$ & \textbf{81.7} & \textbf{8.84} & \underline{20.3} & \underline{0.689} & \textbf{74.7} & \textbf{8.30} & \textbf{18.9} & \textbf{0.710} & \textbf{72.5} & \textbf{8.14} & \underline{17.9} & \textbf{0.707} & \textbf{85.1} & \textbf{9.18} & \underline{18.1} & \underline{0.674} \\
Qwen3-Omni & \underline{63.5} & \underline{7.69} & \textbf{21.4} & \textbf{0.691} & \underline{57.5} & \underline{6.86} & \underline{18.8} & \underline{0.692} & \underline{54.9} & {6.35} & \textbf{18.2} & \underline{0.696} & \underline{70.1} & \underline{7.96} & \textbf{20.3} & \textbf{0.683} \\
UniMoE-2.0 & 51.3 & 6.02 & 13.2 & 0.618 & 44.2 & 5.08 & 12.7 & 0.620 & 45.1 & 5.20 & 12.1 & 0.640 & 55.2 & 6.75 & 12.6 & 0.599 \\
MiniCPM-o-2.6 & 47.8 & 6.13 & 11.0 & 0.604 & 54.0 & 6.01 & 10.2 & 0.617 & 51.0 & 6.08 & 9.6 & 0.608 & 53.7 & 6.42 & 10.1 & 0.574 \\
Baichuan-Omni 1.5 & 53.9 & 6.26 & 15.2 & 0.631 & 50.6 & 5.91 & 14.8 & 0.663 & 45.1 & 5.49 & 15.2 & {0.684} & 59.7 & 6.64 & 14.2 & 0.618 \\
OLA & 53.0 & 6.54 & 14.0 & 0.630 & 52.9 & 6.43 & 14.0 & 0.643 & 43.1 & \underline{6.47} & 13.8 & 0.664 & 61.2 & 7.28 & 13.4 & 0.604 \\
VITA 1.5 & 47.8 & 6.17 & 12.0 & 0.620 & 42.5 & 5.47 & 12.0 & 0.623 & 52.9 & 5.78 & 11.8 & 0.643 & 56.7 & 6.48 & 11.7 & 0.584 \\
VideoLLaMA2 & 24.3 & 3.77 & 9.4 & 0.517 & 28.7 & 4.45 & 8.9 & 0.543 & 23.5 & 4.14 & 7.7 & 0.538 & 23.9 & 4.04 & 9.5 & 0.499 \\
\bottomrule
\end{tabular}
}
\end{table*}

\begin{table*}[!htbp]
\centering
\caption{\textbf{Task 2: MCQ Performance of MLLMs on Topics 9--13.} Acc.: Accuracy (\%), Score: Rationale Judge Score (0--10), RG-L: ROUGE-L (\%), and Sim: Similarity (0--1) shown. \textbf{Bold} = best, \underline{underline} = second-best per topic.}
\label{tab:mcq-topics-group3}
\resizebox{\textwidth}{!}{%
\begin{tabular}{lcccccccccccccccccccc}
\toprule
\textbf{Model} & \multicolumn{4}{c}{\textbf{T9: Housing}} & \multicolumn{4}{c}{\textbf{T10: Restaurant}} & \multicolumn{4}{c}{\textbf{T11: Mental Hlth}} & \multicolumn{4}{c}{\textbf{T12: Town Halls}} & \multicolumn{4}{c}{\textbf{T13: Olympics}} \\
\cmidrule(lr){2-5} \cmidrule(lr){6-9} \cmidrule(lr){10-13} \cmidrule(lr){14-17} \cmidrule(lr){18-21}
 & Acc. & Score & RG-L & Sim & Acc. & Score & RG-L & Sim & Acc. & Score & RG-L & Sim &  Acc. & Score & RG-L & Sim &  Acc. & Score & RG-L & Sim \\
\midrule
Gemini 3.0 Pro$^\dagger$ & \textbf{74.4} & \textbf{8.25} & \textbf{20.0} & \textbf{0.701} & \textbf{74.4} & \textbf{7.70} & \underline{19.0} & \textbf{0.683} & \textbf{87.7} & \textbf{9.34} & \underline{21.3} & \textbf{0.723} & \textbf{82.0} & \textbf{8.89} & \underline{20.2} & \underline{0.704} & \textbf{84.1} & \textbf{8.68} & \textbf{18.9} & \textbf{0.725} \\
Qwen3-Omni & \underline{67.4} & \underline{7.40} & \underline{19.3} & \underline{0.685} & \underline{57.8} & \underline{6.64} & \textbf{19.2} & \underline{0.668} & \underline{70.8} & \underline{8.22} & \textbf{22.3} & \underline{0.720} & \underline{66.1} & \underline{7.90} & \textbf{21.1} & \textbf{0.708} & \underline{60.9} & \underline{6.49} & \textbf{18.9} & \textbf{0.725} \\
UniMoE-2.0 & 48.1 & 5.96 & 13.2 & 0.627 & 44.4 & 5.46 & 12.6 & 0.597 & 55.4 & 7.15 & 13.3 & 0.642 & 48.1 & 6.30 & 13.5 & 0.624 & 47.8 & 5.61 & 11.8 & 0.642 \\
MiniCPM-o-2.6 & 55.8 & 6.19 & 9.2 & 0.560 & 50.0 & 5.73 & 9.4 & 0.568 & 56.9 & 7.15 & 10.7 & 0.621 & 49.2 & 6.24 & 10.7 & 0.598 & 39.1 & 4.88 & 9.8 & 0.627 \\
Baichuan-Omni 1.5 & 51.9 & 5.98 & 15.2 & 0.643 & 45.6 & 5.02 & 14.8 & 0.637 & 63.1 & 6.25 & 14.6 & 0.669 & 50.8 & 6.25 & 15.6 & 0.642 & 53.6 & 5.57 & \underline{15.3} & \underline{0.674} \\
OLA & 52.7 & 6.46 & 13.7 & 0.625 & 51.1 & 6.06 & 14.5 & 0.630 & 66.2 & 7.43 & 14.5 & 0.657 & 50.8 & 6.72 & 14.4 & 0.640 & 37.7 & 5.51 & 13.4 & 0.667 \\
VITA 1.5 & 49.6 & 6.03 & 12.4 & 0.610 & 45.6 & 5.50 & 13.0 & 0.606 & 49.2 & 6.12 & 12.3 & 0.622 & 46.0 & 5.73 & 12.7 & 0.614 & 43.5 & 5.30 & 11.9 & 0.637 \\
VideoLLaMA2 & 26.4 & 4.12 & 9.8 & 0.518 & 26.7 & 4.13 & 8.0 & 0.488 & 23.1 & 4.12 & 9.7 & 0.558 & 23.8 & 4.46 & 9.8 & 0.513 & 24.6 & 3.62 & 7.7 & 0.532 \\
\bottomrule
\end{tabular}
}
\end{table*}

\begin{table*}[t]
\centering
\caption{\textbf{Task 3: Temporal Localization Performance of MLLMs on Topics 1--4.} mIoU (\%), R@0.5 (\%), MAE (seconds), and Score: Rationale Judge Score (0--10) shown. \textbf{Bold} = best, \underline{underline} = second-best per topic. $\downarrow$ indicates lower is better.}
\label{tab:temp-topics-group1}
\resizebox{\textwidth}{!}{%
\begin{tabular}{lcccccccccccccccc}
\toprule
\textbf{Model} & \multicolumn{4}{c}{\textbf{T1: Patient-Dr.}} & \multicolumn{4}{c}{\textbf{T2: Job Int.}} & \multicolumn{4}{c}{\textbf{T3: Parent-Tch.}} & \multicolumn{4}{c}{\textbf{T4: Customer}} \\
\cmidrule(lr){2-5} \cmidrule(lr){6-9} \cmidrule(lr){10-13} \cmidrule(lr){14-17}
 & mIoU & R@0.5 & MAE$\downarrow$ & Score & mIoU & R@0.5 & MAE$\downarrow$ & Score & mIoU & R@0.5 & MAE$\downarrow$ & Score & mIoU & R@0.5 & MAE$\downarrow$ & Score \\
\midrule
Gemini 3.0 Pro$^\dagger$ & \textbf{20.2} & \textbf{18.6} & \textbf{9.2} & \textbf{5.01} & \textbf{32.1} & \textbf{31.0} & \textbf{8.9} & \textbf{5.47} & \textbf{22.7} & \textbf{21.0} & \textbf{12.5} & \textbf{5.12} & \textbf{30.1} & \textbf{31.9} & \textbf{13.4} & \textbf{5.82} \\
Qwen3-Omni & \underline{5.0} & \underline{4.5} & \underline{57.4} & 2.73 & \underline{6.3} & \underline{5.5} & \underline{58.3} & 2.78 & 3.1 & \underline{1.5} & \underline{49.9} & 2.72 & 3.7 & \underline{3.4} & 62.7 & 2.40 \\
UniMoE-2.0 & 2.0 & 1.9 & 1049.4 & 2.05 & 2.5 & 1.2 & 618.8 & 2.17 & 0.9 & 1.0 & 1423.0 & 2.22 & 3.1 & 0.9 & 315.8 & 1.98 \\
MiniCPM-o-2.6 & 1.4 & 0.4 & 657.9 & 3.37 & 3.3 & 1.6 & 372.0 & 3.50 & 2.1 & \underline{1.5} & 1041.5 & 3.41 & 2.1 & 0.9 & 123.3 & \underline{4.58} \\
Baichuan-Omni 1.5 & 2.0 & 0.5 & 125.0 & 2.47 & 3.4 & 1.0 & 60.7 & 2.76 & 3.6 & 1.2 & 549.2 & 2.36 & 3.3 & 0.9 & 140.5 & 2.20 \\
OLA & 2.5 & 0.4 & 415.6 & 3.48 & 3.0 & 0.8 & 270.5 & \underline{3.88} & 3.0 & 0.7 & 436.0 & 3.42 & 3.7 & 0.9 & 132.6 & 3.73 \\
VITA 1.5 & 3.5 & 3.8 & 170.1 & \underline{3.77} & 2.6 & 1.2 & 122.2 & 3.79 & 2.3 & 1.3 & 176.2 & \underline{3.58} & 1.2 & 0.9 & 86.2 & 4.04 \\
VideoLLaMA2 & 4.0 & 1.1 & 357.0 & 2.05 & 4.2 & 0.4 & 219.2 & 2.05 & 3.1 & 0.2 & 510.7 & 2.19 & \underline{4.7} & 0.0 & 126.4 & 2.00 \\
\bottomrule
\end{tabular}
}
\end{table*}

\begin{table*}[t]
\centering
\caption{\textbf{Task 3: Temporal Localization Performance of MLLMs on Topics 5--8.} mIoU (\%), R@0.5 (\%), MAE (seconds), and Score: Rationale Judge Score (0--10) shown. \textbf{Bold} = best, \underline{underline} = second-best per topic. $\downarrow$ indicates lower is better.}
\label{tab:temp-topics-group2}
\resizebox{\textwidth}{!}{%
\begin{tabular}{lcccccccccccccccc}
\toprule
\textbf{Model} & \multicolumn{4}{c}{\textbf{T5: Courtroom}} & \multicolumn{4}{c}{\textbf{T6: Emergency}} & \multicolumn{4}{c}{\textbf{T7: Transport}} & \multicolumn{4}{c}{\textbf{T8: Workplace}} \\
\cmidrule(lr){2-5} \cmidrule(lr){6-9} \cmidrule(lr){10-13} \cmidrule(lr){14-17}
 & mIoU & R@0.5 & MAE$\downarrow$ & Score & mIoU & R@0.5 & MAE$\downarrow$ & Score & mIoU & R@0.5 & MAE$\downarrow$ & Score & mIoU & R@0.5 & MAE$\downarrow$ & Score \\
\midrule
Gemini 3.0 Pro$^\dagger$ & \textbf{32.7} & \textbf{33.5} & \textbf{15.1} & \textbf{5.38} & \textbf{20.0} & \textbf{17.8} & \textbf{14.1} & \textbf{5.44} & \textbf{27.0} & \textbf{22.6} & \textbf{7.8} & \textbf{6.09} & \textbf{28.8} & \textbf{29.6} & \textbf{10.2} & \textbf{5.31} \\
Qwen3-Omni & \underline{4.0} & \underline{2.9} & \underline{38.5} & 2.59 & \underline{3.0} & \underline{2.4} & \underline{42.3} & 2.46 & 1.4 & 0.0 & \underline{44.8} & 2.57 & 4.1 & 3.2 & \underline{59.0} & 2.50 \\
UniMoE-2.0 & 1.1 & 0.6 & 1532.0 & 2.00 & 1.4 & 0.4 & 477.4 & 2.03 & 1.6 & \underline{1.0} & 550.3 & 2.64 & 4.8 & \underline{3.3} & 391.5 & 1.94 \\
MiniCPM-o-2.6 & 2.3 & 1.5 & 652.0 & 3.28 & 1.1 & 0.0 & 238.1 & 4.10 & 0.3 & 0.0 & 263.0 & 3.91 & 1.6 & 0.7 & 302.7 & 3.29 \\
Baichuan-Omni 1.5 & 3.3 & 1.3 & 335.4 & 2.52 & 2.3 & 1.3 & 64.0 & 1.63 & 1.4 & 0.0 & 56.6 & 2.29 & 3.1 & 0.9 & 131.5 & 2.32 \\
OLA & 2.9 & 1.2 & 482.1 & 3.79 & 2.4 & 1.2 & 109.3 & 3.53 & 2.2 & 0.9 & 228.9 & 3.84 & 3.4 & 1.9 & 360.7 & 3.05 \\
VITA 1.5 & 1.8 & 1.5 & 128.4 & \underline{4.08} & 0.7 & 0.0 & 63.9 & \underline{4.45} & 0.6 & 0.0 & 137.7 & \underline{4.31} & 1.5 & 0.0 & 166.2 & \underline{3.59} \\
VideoLLaMA2 & 3.2 & 0.3 & 480.8 & 2.25 & 1.8 & 0.4 & 215.6 & 2.52 & \underline{3.4} & 0.0 & 179.7 & 3.40 & \underline{4.9} & 1.3 & 176.0 & 2.33 \\
\bottomrule
\end{tabular}
}
\end{table*}

\begin{table*}[!t]
\centering
\caption{\textbf{Task 3: Temporal Localization Performance of MLLMs on Topics 9--13.} mIoU (\%), R@0.5 (\%), MAE (seconds), and Score: Rationale Judge Score (0--10) shown. \textbf{Bold} = best, \underline{underline} = second-best per topic. $\downarrow$ indicates lower is better.}
\label{tab:temp-topics-group3}
\resizebox{\textwidth}{!}{%
\begin{tabular}{lcccccccccccccccccccc}
\toprule
\textbf{Model} & \multicolumn{4}{c}{\textbf{T9: Housing}} & \multicolumn{4}{c}{\textbf{T10: Restaurant}} & \multicolumn{4}{c}{\textbf{T11: Mental Hlth}} & \multicolumn{4}{c}{\textbf{T12: Town Halls}} & \multicolumn{4}{c}{\textbf{T13: Olympics}} \\
\cmidrule(lr){2-5} \cmidrule(lr){6-9} \cmidrule(lr){10-13} \cmidrule(lr){14-17} \cmidrule(lr){18-21}
 & mIoU & R@0.5 & MAE$\downarrow$ & Score & mIoU & R@0.5 & MAE$\downarrow$ & Score & mIoU & R@0.5 & MAE$\downarrow$ & Score & mIoU & R@0.5 & MAE$\downarrow$ & Score & mIoU & R@0.5 & MAE$\downarrow$ & Score \\
\midrule
Gemini 3.0 Pro$^\dagger$ & \textbf{29.5} & \textbf{27.6} & \textbf{8.5} & \textbf{5.16} & \textbf{27.7} & \textbf{26.6} & \textbf{9.4} & \textbf{5.52} & \textbf{29.5} & \textbf{29.6} & \textbf{9.4} & \textbf{5.45} & \textbf{20.5} & \textbf{15.9} & \textbf{23.5} & \textbf{5.07} & \textbf{25.2} & \textbf{24.8} & \textbf{13.6} & \textbf{5.11} \\
Qwen3-Omni & 3.3 & \underline{3.1} & 73.9 & 2.20 & \underline{5.8} & \underline{4.3} & 82.7 & 2.65 & 3.2 & \underline{2.6} & 67.1 & 2.66 & \underline{4.2} & \underline{3.0} & \underline{95.7} & 2.97 & 1.3 & 0.0 & 47.2 & 2.29 \\
UniMoE-2.0 & 1.3 & 0.8 & 401.3 & 2.08 & 2.2 & 0.5 & 429.0 & 2.45 & 0.9 & 0.7 & 429.7 & 1.74 & 0.4 & 0.2 & 1111.6 & 2.18 & 1.3 & 1.2 & 185.3 & 1.95 \\
MiniCPM-o-2.6 & 2.2 & 1.3 & 170.9 & 3.42 & 1.8 & 0.5 & 201.9 & 3.97 & 2.8 & 0.7 & 208.2 & \underline{3.68} & 2.2 & 0.5 & 469.7 & 3.62 & 0.3 & 0.0 & 71.4 & 3.38 \\
Baichuan-Omni 1.5 & 2.1 & 1.1 & 85.7 & 1.87 & 3.4 & 1.1 & \underline{72.0} & 2.36 & 2.6 & 0.9 & 95.3 & 3.16 & 2.8 & 1.6 & 251.2 & 1.97 & 2.5 & 2.1 & 94.5 & 1.82 \\
OLA & \underline{4.5} & 2.3 & 129.7 & 3.35 & 4.1 & 2.4 & 137.5 & 3.29 & \underline{3.9} & 0.7 & 183.4 & 3.40 & 3.0 & 0.9 & 584.4 & \underline{3.63} & \underline{3.8} & 1.9 & 50.4 & 2.90 \\
VITA 1.5 & 2.4 & 1.6 & \underline{52.5} & \underline{4.36} & 1.7 & 0.5 & 96.9 & \underline{4.39} & 1.1 & 0.7 & \underline{55.6} & 3.37 & 2.3 & 1.3 & 209.0 & 3.39 & 1.9 & \underline{2.3} & \underline{44.6} & \underline{3.66} \\
VideoLLaMA2 & 3.6 & 0.5 & 134.2 & 2.37 & 3.1 & 0.0 & 129.4 & 2.10 & 3.2 & 0.0 & 168.3 & 1.83 & 3.2 & 0.2 & 253.1 & 1.77 & 3.0 & 1.1 & 98.3 & 2.01 \\
\bottomrule
\end{tabular}
}
\end{table*}

\subsection{Key Observations from Per-Topic Analysis}

\textbf{Topic-specific performance variation.} Models exhibit substantial performance variation across topics. In summarization (Tables~\ref{tab:sum-topics-group1}--\ref{tab:sum-topics-group3}), Gemini achieves 8.33 on T2: Job Interviews but drops to 5.04 on T13: Olympics. MCQ accuracy (Tables~\ref{tab:mcq-topics-group1}--\ref{tab:mcq-topics-group3}) shows similar variation: Qwen3-Omni achieves 77.0\% on T2 but only 54.9\% on T7: Transportation.

\textbf{Temporal hallucination in open-source models.} Temporal localization (Tables~\ref{tab:temp-topics-group1}--\ref{tab:temp-topics-group3}) reveals stark differences between models. Gemini maintains consistent grounding (MAE: 7.8--23.5s, R@0.5: 15.9--33.5\%), while open-source models exhibit systematic temporal reference frame hallucination. Despite explicit prompting with absolute timestamps, some open-source models treat each 3-minute segment as starting at 0s. For example, UniMoE-2.0 achieves MAE of 1049.4--1532.0s on T1, T3, and T5 despite reasonable rationale scores (2.00--2.22), indicating models understand \textit{what} happens but fail to ground \textit{when} in absolute video time.

\textbf{Model hierarchy.} Results show clear performance tiers: Gemini (MAE: 7.8--23.5s), Qwen3-Omni (38.5--95.7s),and UniMoE-2.0/MiniCPM-o-2.6 (often 300--1500s). This exposes a fundamental limitation: most open-source models lack mechanisms to maintain temporal coherence across segments, treating each chunk independently rather than as part of a continuous timeline.

\textbf{Rationale-accuracy disconnect.} High rationale scores do not guarantee temporal accuracy. VITA 1.5 achieves rationale scores of 3.37--4.45 but R@0.5 of 0.0--2.3\% (MAE: 44.6--209.0s). UniMoE-2.0 and MiniCPM-o-2.6 similarly produce coherent explanations (scores 1.74--4.10) while exhibiting severe localization errors (MAE $>$1000s on multiple topics).

\textbf{Professional vs. dynamic scenes.} Models consistently perform better on structured interactions (T1: Patient-Doctor, T2: Job Interviews, T3: Parent-Teacher) than dynamic scenarios (T6: Emergency, T7: Transportation, T13: Olympics). Even Gemini's MAE increases from 8.9--12.5s on structured topics to 13.6--22.6s on dynamic ones, though it remains substantially better than open-source alternatives.

\subsection{Task 1: Video Summarization Per-Topic}
Tables~\ref{tab:sum-topics-group1}, \ref{tab:sum-topics-group2}, and~\ref{tab:sum-topics-group3} report summarization performance broken down by topic.

\begin{table*}[t]
\centering
\small
\caption{\textbf{Error taxonomy distribution (\%) for Task~3 temporal localization.} Error breakdown of MLLMs predictions on all dataset. Hallucination aggregates invalid outputs, out-of-bounds timestamps, and extreme ``teleport'' errors. Each prediction is assigned to exactly one category. $\Delta$R@0.5: gain after correcting relative-to-absolute mismatches.}
\setlength{\tabcolsep}{4pt}
\resizebox{\textwidth}{!}{%
\begin{tabular}{l | ccccccc}
\toprule
\textbf{Model} &
\textbf{Relative--Absolute} &
\textbf{Hallucination} &
\textbf{Too Early} &
\textbf{Too Late} &
\textbf{Wrong Event / Right Time} &
\textbf{Other} &
\textbf{$\Delta$R@ 0.5} \\
\midrule
Gemini 3.0 Pro    & 0.2  & 0.4  & 5.6  & 9.2  & 3.1 & 81.5 & +0.09 \\
Qwen3-Omni        & 0.5  & 7.0  & 30.5 & 36.9 & 1.3 & 23.8 & +0.09 \\
UniMoE-2.0       & 39.4 & 27.1 & 13.0 & 9.6  & 0.7 & 10.1 & +1.14 \\
MiniCPM-o-2.6     & 7.3  & 26.8 & 14.8 & 37.5 & 0.6 & 12.9 & +0.04 \\
Baichuan-Omni 1.5 & 0.9  & 13.0 & 25.9 & 38.5 & 0.8 & 21.0 & +0.00 \\
OLA               & 10.7 & 11.9 & 17.8 & 33.2 & 0.8 & 25.5 & +0.13 \\
VITA-1.5          & 0.8  & 7.7  & 38.0 & 36.6 & 0.5 & 16.4 & +0.00 \\
VideoLLaMA2       & 10.4 & 10.0 & 17.6 & 11.6 & 0.3 & 50.2 & +0.18 \\
\bottomrule
\end{tabular}
}
\label{tab:temporal_error_taxonomy}
\end{table*}

\begin{figure*}[t]
    \centering
    \includegraphics[width=0.85\textwidth]{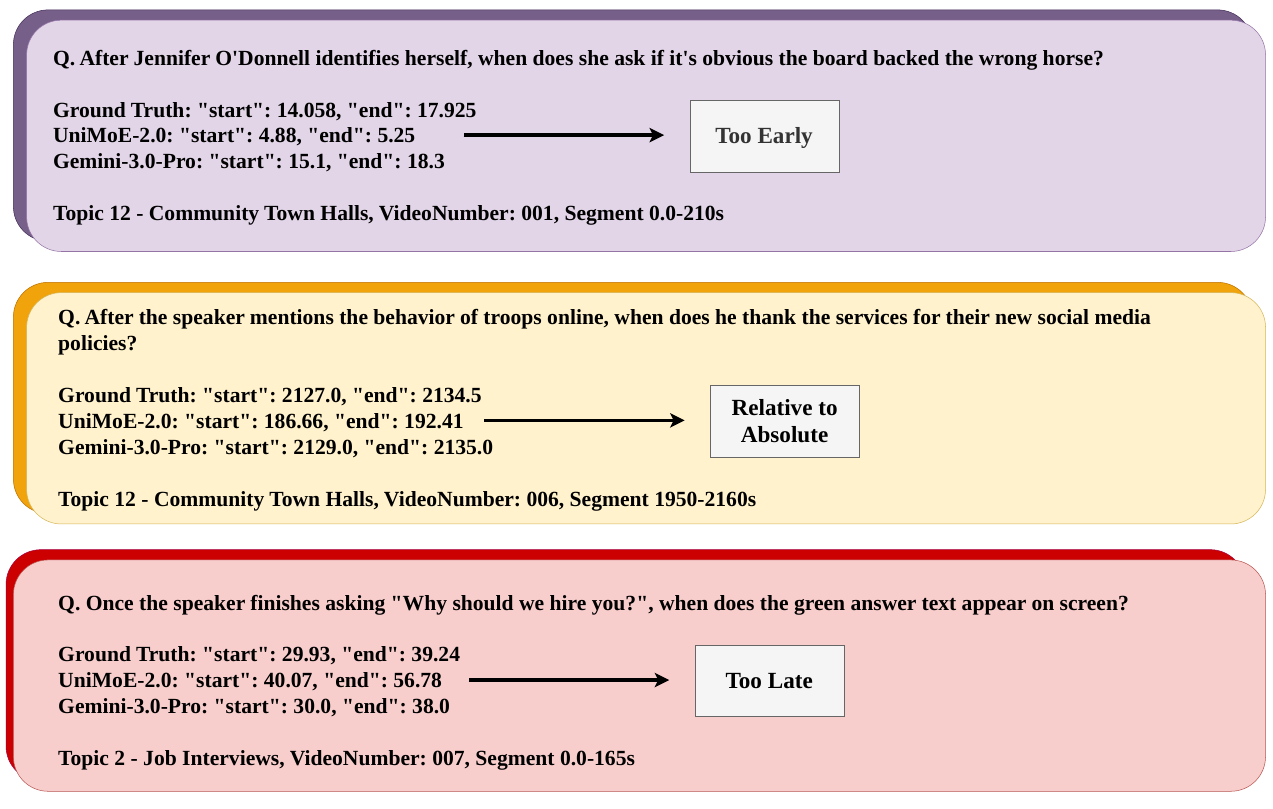}
    \caption{Qualitative examples of temporal localization errors. Three common error types are illustrated: \textbf{Too Early} (top)-predictions preceding the ground truth event; \textbf{Relative-to-Absolute mismatch} (middle)-model outputs segment-relative rather than absolute timestamps; \textbf{Too Late} (bottom)-predictions following the ground truth event. Each example shows the question, ground truth timestamps, and predictions from UniMoE-2.0 and Gemini-3.0-Pro.}
    \vspace{-1em}
    \label{fig:quali_temporal}
\end{figure*}

\section{Error Taxonomy Analysis}
\label{app:temporal_error_analysis}
To better understand failure modes in Task~3 (Temporal Localization), we categorize each prediction into a small set of error types. In this task, models are given absolute segment boundaries and are required to output absolute timestamps within the provided segment.
Ground-truth timestamps are obtained by converting segment-relative annotations to absolute video time by adding the segment start offset.

\paragraph{Notation.}
Let the ground-truth interval be $G=[s_{gt}, e_{gt}]$ and the predicted interval be $P=[s_{pred}, e_{pred}]$, both in seconds.
We use the same IoU and MAE definitions as in the main evaluation.

\subsection{Category Assignment Rules}
Each prediction is assigned to exactly one category using the following rules (evaluated in order).
We use two fixed thresholds: a reference-frame tolerance $\tau_{\text{relabs}}=10$s and a timing-shift tolerance $\tau_{\text{shift}}=5$s.

\textbf{Relative-to-absolute mismatch} captures cases where the model outputs timestamps as if time ``starts at 0'' at the beginning of the segment. We detect this by shifting the prediction by the segment start time: $P'=[s_{pred}+S,\ e_{pred}+S]$, where $S$ is the segment start. If $P'$ is close to the ground truth (within $\tau_{\text{relabs}}$) while the original $P$ is not, we label it a relative-to-absolute mismatch.

\textbf{Hallucination} covers violations of output validity or prompt constraints: (i) malformed intervals ($e_{pred}\le s_{pred}$), (ii) timestamps outside the provided segment bounds, or (iii) extreme ``teleport'' errors far from the segment window.

\textbf{Timing shift (Too Early / Too Late)} is assigned when IoU$(P,G)<0.1$ and the prediction is clearly before or after the ground truth by more than $\tau_{\text{shift}}$: \emph{Too Early} if $e_{pred} < s_{gt} - \tau_{\text{shift}}$; \emph{Too Late} if $s_{pred} > e_{gt} + \tau_{\text{shift}}$.

\textbf{Wrong event, right time} is assigned when IoU$(P,G)>0.5$ but LLM-judge score $<5$, suggesting the model localized a different event at a similar time.

\textbf{Other} includes all remaining errors not covered above.

\subsection{Coordinate Alignment Parser}
To quantify how much performance is lost purely due to reference-frame formatting errors, we apply a coordinate alignment parser to all predictions on segments starting after 0s, checking whether $s_{pred}+S \approx s_{gt}$ within $\tau_{\text{relabs}}=10$s. Results (Table~\ref{tab:temporal_error_taxonomy}, $\Delta$R@0.5 column) show UniMoE-2.0 benefits most (+1.14\% R@0.5, 7.6\% of predictions corrected), consistent with its dominant relative-to-absolute error rate (39.4\%). Modest gains for other models confirm that most failures reflect genuine temporal reasoning limitations rather than formatting errors. We release the parser as part of our evaluation suite.

\subsection{Key Observations}
Figure~\ref{fig:quali_temporal} provides qualitative examples of UniMoE-2.0 and Gemini 3.0 Pro predictions. Table~\ref{tab:temporal_error_taxonomy} shows clear model-dependent failure patterns. UniMoE-2.0 exhibits dominant relative-to-absolute mismatches (39.4\%), while MiniCPM-o-2.6 shows substantial hallucination (26.8\%) and late timing shifts (37.5\%). Qwen3-Omni, VITA-1.5, and Baichuan-Omni~1.5 are dominated by timing shifts (67.4\%, 74.6\%, and 64.4\%), consistent with correct neighborhood selection but imprecise localization. Gemini 3.0 Pro has minimal structured errors with most falling into \textit{Other} (81.5\%), and VideoLLaMA2 and OLA show balanced profiles with moderate reference-frame errors (10.4\% and 10.7\%).

\begin{table*}[!t]
\centering
\caption{\textbf{Performance of MLLMs on the summarization task across demographics.} Results are on all dataset. Score: Judge score (0--10),  RG-L: ROUGE-L (\%), and Sim: similarity (0--1). \textbf{Bold} = best, \worst{highlighted} = worst within each demographic group (Race/Gender/Age) per column.}
\label{tab:demo_summ_a}
\scriptsize
\setlength{\tabcolsep}{1pt}
\resizebox{\textwidth}{!}{%
\begin{tabular}{@{}l | c c c c c c c c c c c c c c c c c c c c c c c c@{}}
\toprule
Group & \multicolumn{3}{c}{Gemini 3.0 Pro$^\dagger$} & \multicolumn{3}{c}{Qwen3-Omni} & \multicolumn{3}{c}{UniMoE-2.0} & \multicolumn{3}{c}{MiniCPM-o-2.6} & \multicolumn{3}{c}{Baichuan-Omni 1.5} & \multicolumn{3}{c}{OLA} & \multicolumn{3}{c}{VITA 1.5} & \multicolumn{3}{c}{VideoLLaMA2} \\
\cmidrule(lr){2-4} \cmidrule(lr){5-7} \cmidrule(lr){8-10} \cmidrule(lr){11-13} \cmidrule(lr){14-16} \cmidrule(lr){17-19} \cmidrule(lr){20-22} \cmidrule(lr){23-25}
& Score & RG-L & Sim & Score & ROUGE-L & Sim & Score & RG-L & Sim & Score & ROUGE-L & Sim & Score & RG-L & Sim & Score & ROUGE-L & Sim & Score & RG-L & Sim & Score & ROUGE-L & Sim \\
\midrule
\multicolumn{25}{l}{\textit{Race}}\\
Arab & 6.90 & 27.8 & 0.811 & \high{5.95} & 22.2 & 0.695 & \high{5.00} & \high{21.9} & 0.706 & 3.57 & \worst{14.1} & 0.576 & \high{4.29} & \high{20.3} & 0.654 & \high{4.76} & \high{19.6} & 0.644 & \high{2.76} & \high{16.4} & 0.465 & \worst{1.00} & 11.5 & 0.289 \\
Indigenous & 6.70 & \worst{24.3} & \high{0.853} & \worst{4.13} & \worst{19.0} & \high{0.745} & 4.35 & \worst{18.9} & \high{0.749} & \high{3.61} & 14.6 & \high{0.689} & 3.70 & 17.9 & \high{0.730} & 4.39 & \worst{16.8} & \high{0.689} & \worst{1.65} & 15.2 & 0.440 & 1.04 & \worst{8.9} & \worst{0.227} \\
Asian & \high{7.05} & \high{27.9} & 0.818 & 5.71 & \high{23.2} & 0.740 & 4.62 & 20.2 & 0.695 & 3.26 & 14.5 & 0.579 & 3.61 & 17.7 & 0.577 & 4.29 & 18.5 & 0.619 & 2.65 & 15.8 & \high{0.488} & \high{1.63} & 13.0 & \high{0.412} \\
White & 6.68 & 25.9 & 0.809 & 5.28 & 21.6 & 0.701 & 4.29 & 20.2 & 0.699 & 3.26 & 14.7 & 0.572 & 3.22 & 18.1 & 0.591 & 4.27 & 18.6 & 0.636 & 2.50 & 16.0 & 0.476 & 1.45 & 12.7 & 0.367 \\
Hispanic & 6.41 & 26.3 & 0.809 & 4.99 & 21.8 & 0.704 & 3.70 & 20.2 & 0.691 & 3.04 & \high{15.0} & \worst{0.558} & \worst{2.44} & 17.4 & 0.564 & 3.62 & 17.8 & 0.604 & 2.21 & \worst{15.1} & 0.466 & 1.23 & 12.6 & 0.352 \\
Black & \worst{6.02} & 25.2 & \worst{0.806} & 4.39 & 20.8 & \worst{0.663} & \worst{3.45} & 19.0 & \worst{0.683} & \worst{2.92} & 14.9 & 0.559 & 2.55 & \worst{16.8} & \worst{0.543} & \worst{3.60} & 16.9 & \worst{0.580} & 2.31 & 15.7 & \worst{0.438} & 1.38 & \high{13.1} & 0.363 \\
\midrule
\multicolumn{25}{l}{\textit{Gender}}\\
Male & \worst{6.36} & \worst{25.8} & \worst{0.809} & \worst{4.97} & \worst{21.4} & \worst{0.698} & \worst{3.93} & \worst{19.8} & \worst{0.690} & \worst{2.98} & \worst{14.6} & \worst{0.567} & \worst{2.94} & \worst{17.7} & \worst{0.578} & \worst{4.00} & \worst{18.1} & \worst{0.613} & \worst{2.31} & \worst{15.8} & \high{0.470} & \worst{1.36} & \high{12.8} & \high{0.370} \\
Female & \high{7.02} & \high{26.9} & \high{0.816} & \high{5.47} & \high{22.2} & \high{0.713} & \high{4.55} & \high{20.3} & \high{0.707} & \high{3.53} & \high{14.8} & \high{0.583} & \high{3.47} & \high{17.9} & \high{0.592} & \high{4.29} & \high{18.3} & \high{0.635} & \high{2.65} & \high{15.9} & \worst{0.467} & \high{1.53} & \worst{12.4} & \worst{0.361} \\
\midrule
\multicolumn{25}{l}{\textit{Age}}\\
18--24 & \worst{6.28} & \worst{25.4} & \high{0.818} & \worst{4.93} & \worst{21.5} & \high{0.713} & 4.02 & 20.0 & \high{0.710} & \high{3.30} & \high{15.3} & \high{0.623} & \worst{3.04} & \high{18.3} & \high{0.616} & 4.12 & \worst{17.8} & \high{0.634} & \worst{2.41} & \high{16.7} & \high{0.493} & 1.38 & \worst{12.5} & \high{0.387} \\
25--39 & 6.52 & 26.2 & 0.811 & 5.01 & \worst{21.5} & 0.703 & \worst{4.01} & \worst{19.7} & \worst{0.689} & \worst{3.14} & 14.7 & 0.570 & 3.11 & \worst{17.7} & 0.583 & \worst{3.96} & 18.0 & \worst{0.616} & 2.45 & 15.7 & 0.466 & \high{1.50} & \high{12.8} & 0.377 \\
40+ & \high{6.91} & \high{26.7} & \worst{0.809} & \high{5.47} & \high{22.1} & \worst{0.700} & \high{4.45} & \high{20.5} & 0.699 & 3.21 & \worst{14.4} & \worst{0.552} & \high{3.25} & \worst{17.7} & \worst{0.568} & \high{4.29} & \high{18.6} & 0.623 & \high{2.47} & \worst{15.5} & \worst{0.461} & \worst{1.36} & 12.6 & \worst{0.345} \\
\bottomrule
\end{tabular}}
\end{table*}

\begin{table*}[t]
\centering
\caption{\textbf{Performance of MLLMs on the MCQ task across demographics (Models 1--4).} Results on all dataset. Acc.: Accuracy (\%), and rationale quality (Score: Judge score (0--10), RG-L: ROUGE-L (\%), Sim: similarity (0--1)). \textbf{Bold} = best, \worst{highlighted} = worst within each demographic group per column.}
\label{tab:demo_mcq_a}
\scriptsize
\setlength{\tabcolsep}{2pt}
\resizebox{\textwidth}{!}{%
\begin{tabular}{@{}l | c c c c c c c c c c c c c c c c@{}}
\toprule
Group & \multicolumn{4}{c}{Gemini 3.0 Pro$^\dagger$} & \multicolumn{4}{c}{Qwen3-Omni} & \multicolumn{4}{c}{UniMoE-2.0} & \multicolumn{4}{c}{MiniCPM-o-2.6} \\
\cmidrule(lr){2-5} \cmidrule(lr){6-9} \cmidrule(lr){10-13} \cmidrule(lr){14-17}
& Acc. & Score & RG-L & Sim & Acc. & Score & RG-L & Sim & Acc. & Score & RG-L & Sim & Acc. & Score & RG-L & Sim \\
\midrule
\multicolumn{17}{l}{\textit{Race}}\\
Arab       & \high{82.7} & 8.65 & \high{20.4} & 0.724 & 63.9 & 7.56 & \high{21.2} & 0.711 & 48.7 & \high{6.37} & \high{13.9} & 0.638 & 50.3 & 6.17 & 10.6 & 0.617 \\
Indigenous & 80.0 & 8.77 & \worst{19.5} & \high{0.746} & \high{68.6} & \worst{7.03} & 20.2 & \high{0.731} & \worst{42.9} & \worst{5.11} & \worst{11.7} & \high{0.665} & 48.6 & 5.80 & \worst{8.4} & \high{0.622} \\
Black      & 79.2 & 8.68 & 19.8 & 0.712 & 63.5 & 7.35 & 20.1 & 0.705 & 48.9 & 6.25 & 13.3 & 0.633 & 49.5 & 6.00 & 10.0 & 0.596 \\
Asian      & 80.3 & 8.63 & 19.8 & \worst{0.706} & \worst{61.8} & 7.33 & 20.5 & 0.705 & \high{53.6} & 6.30 & 13.3 & 0.639 & 51.0 & \high{6.38} & 10.6 & 0.605 \\
White      & 81.2 & \high{8.78} & 19.9 & 0.709 & 66.5 & \high{7.57} & 20.7 & 0.705 & 51.0 & 6.15 & 13.3 & \worst{0.627} & \high{51.2} & 6.08 & \high{10.7} & 0.609 \\
Hispanic   & \worst{79.1} & \worst{8.55} & \worst{19.5} & \worst{0.706} & 63.5 & 7.43 & \worst{19.8} & \worst{0.697} & 48.5 & 5.91 & 13.2 & 0.631 & \worst{48.5} & \worst{5.74} & 10.2 & \worst{0.590} \\
\midrule
\multicolumn{17}{l}{\textit{Gender}}\\
Male   & \high{81.4} & \high{8.73} & \high{19.9} & \high{0.711} & \high{65.1} & \worst{7.41} & \worst{20.4} & \high{0.706} & \high{51.4} & \high{6.19} & \worst{13.2} & \high{0.634} & \worst{49.7} & \worst{6.00} & \worst{10.4} & \high{0.607} \\
Female & \worst{79.4} & \worst{8.68} & \worst{19.8} & \worst{0.708} & \worst{64.5} & \high{7.56} & \high{20.6} & \worst{0.703} & \worst{49.2} & \worst{6.12} & \high{13.4} & \worst{0.628} & \high{51.5} & \high{6.17} & \high{10.5} & \worst{0.602} \\
\midrule
\multicolumn{17}{l}{\textit{Age}}\\
18--24 & 80.9 & \worst{8.64} & \worst{19.0} & \high{0.711} & \worst{59.5} & \worst{6.64} & \worst{19.5} & 0.704 & 49.3 & \worst{5.84} & \worst{12.3} & \worst{0.624} & \high{50.7} & \worst{5.63} & \worst{9.8} & 0.601 \\
25--39 & \worst{79.5} & \worst{8.64} & 19.7 & \worst{0.710} & 63.5 & 7.36 & 20.2 & \worst{0.703} & \worst{48.7} & 5.96 & 13.2 & 0.628 & \worst{50.2} & 5.97 & 10.2 & \worst{0.599} \\
40+    & \high{81.4} & \high{8.79} & \high{20.2} & \worst{0.710} & \high{67.3} & \high{7.76} & \high{21.0} & \high{0.707} & \high{52.4} & \high{6.42} & \high{13.6} & \high{0.636} & \high{50.7} & \high{6.25} & \high{10.8} & \high{0.611} \\
\bottomrule
\end{tabular}}
\end{table*}

\begin{table*}[h]
\centering
\caption{\textbf{Performance of MLLMs on the MCQ task across demographics (Models 5--8).} Results on all dataset. Acc.: Accuracy (\%), and rationale quality (Score: Judge score (0--10), RG-L: ROUGE-L (\%), Sim: similarity (0--1)). \textbf{Bold} = best, \worst{highlighted} = worst within each demographic group per column.}
\label{tab:demo_mcq_b}
\scriptsize
\setlength{\tabcolsep}{2pt}
\resizebox{\textwidth}{!}{%
\begin{tabular}{@{}l | c c c c c c c c c c c c c c c c@{}}
\toprule
Group & \multicolumn{4}{c}{Baichuan-Omni 1.5} & \multicolumn{4}{c}{OLA} & \multicolumn{4}{c}{VITA 1.5} & \multicolumn{4}{c}{VideoLLaMA2} \\
\cmidrule(lr){2-5} \cmidrule(lr){6-9} \cmidrule(lr){10-13} \cmidrule(lr){14-17}
& Acc. & Score & RG-L & Sim & Acc. & Score & RG-L & Sim & Acc. & Score & RG-L & Sim & Acc. & Score & RG-L & Sim \\
\midrule
\multicolumn{17}{l}{\textit{Race}}\\
Arab       & 58.6 & 6.30 & 15.3 & 0.655 & 51.3 & 6.53 & 14.1 & \worst{0.642} & 47.1 & 6.10 & \high{12.6} & \worst{0.614} & 20.9 & 3.92 & \high{9.9} & 0.519 \\
Indigenous & \high{62.9} & \high{6.89} & 15.3 & \high{0.720} & \worst{31.4} & 5.97 & \worst{13.9} & \high{0.688} & \worst{42.4} & \high{6.39} & \worst{12.1} & \high{0.653} & \worst{8.6} & \worst{3.51} & \worst{7.7} & \high{0.546} \\
Black      & 52.2 & 5.97 & 15.2 & 0.657 & 48.3 & 6.41 & 14.1 & 0.646 & 48.1 & 5.69 & 12.3 & 0.624 & 24.6 & 4.19 & 9.1 & \worst{0.518} \\
Asian      & 55.7 & 6.16 & \worst{15.1} & \worst{0.654} & \high{53.4} & \high{6.69} & 14.0 & 0.644 & \high{49.5} & 6.21 & 12.5 & 0.626 & 23.2 & 4.20 & \high{9.9} & 0.523 \\
White      & 55.8 & 6.31 & 15.3 & \worst{0.654} & 52.7 & 6.64 & \high{14.2} & 0.646 & 49.0 & 6.01 & 12.3 & 0.620 & \high{26.7} & \high{4.21} & 9.4 & 0.531 \\
Hispanic   & \worst{48.2} & \worst{5.70} & \high{15.4} & 0.656 & 46.5 & \worst{5.96} & 14.0 & 0.652 & 48.5 & \worst{5.51} & 12.5 & 0.627 & 22.6 & 3.88 & 9.3 & 0.530 \\
\midrule
\multicolumn{17}{l}{\textit{Gender}}\\
Male   & \high{55.7} & \high{6.19} & \high{15.4} & \high{0.657} & \high{51.6} & 6.52 & 14.1 & \high{0.648} & \high{49.1} & \worst{5.84} & 12.4 & \high{0.624} & \high{25.1} & \worst{4.11} & \worst{9.3} & \high{0.528} \\
Female & \worst{53.2} & \worst{6.14} & \worst{15.1} & \worst{0.653} & \worst{50.3} & 6.52 & 14.1 & \worst{0.646} & \worst{48.2} & \high{6.07} & 12.4 & \worst{0.621} & \worst{24.6} & \high{4.20} & \high{9.5} & \worst{0.526} \\
\midrule
\multicolumn{17}{l}{\textit{Age}}\\
18--24 & 55.4 & \worst{5.81} & \worst{14.8} & \high{0.661} & \worst{46.3} & \worst{5.99} & \worst{13.8} & \high{0.650} & \high{49.5} & \worst{5.63} & \worst{12.1} & \high{0.631} & \worst{23.8} & \worst{3.91} & \worst{8.4} & \worst{0.518} \\
25--39 & \worst{51.8} & 5.98 & 15.2 & 0.656 & 49.2 & 6.40 & 14.0 & \worst{0.643} & \worst{48.6} & 5.94 & \high{12.4} & 0.623 & \high{25.4} & 4.10 & 9.3 & 0.524 \\
40+    & \high{57.0} & \high{6.42} & \high{15.4} & \worst{0.654} & \high{53.7} & \high{6.74} & \high{14.4} & 0.649 & 48.7 & \high{6.00} & \high{12.4} & \worst{0.620} & 24.7 & \high{4.23} & \high{9.7} & \high{0.532} \\
\bottomrule
\end{tabular}}
\end{table*}

\section{Detailed Results Across Demographic Groups}
\label{app:demographic}

\subsection{Qualitative Analysis}
\label{app:demographic_qualitative}
Tables~\ref{tab:demo_summ_a}--\ref{tab:demo_temporal_b} report performance stratified by race, gender, and age. Observed demographic performance differences recur across models and tasks, indicating diagnostic robustness gaps.

\textbf{Race disparities.} Models most consistently underperform on Black participants. In summarization, Gemini drops from 7.05 (Asian) to 6.02 (Black; $-1.03$), and Qwen from 5.95 (Arab) to 4.39 (Black; $-1.56$). Indigenous participants show inconsistent treatment: some models remain competitive, while others degrade sharply (Qwen: 4.13, VITA: 1.65), which may reflect limited coverage of such speech patterns in training corpora.

\textbf{Gender disparities.} Female-participant videos consistently yield higher performance across all models and tasks. Female advantages in summarization range from $+0.17$ to $+0.66$ across models. The uniform directionality suggests systematic differences in processing male vs. female voices and interaction styles.

\textbf{Age disparities.} Models favor older speakers. For summarization, the 40+ group outperforms 18--24 (Gemini: 6.91 vs. 6.28), consistent with training data skewed toward formal, professional contexts aligned with older-participant scenarios.

\textbf{Temporal localization failures.} Temporal localization exhibits the most severe disparities. For Indigenous participants, some open-source models reach as low as 0.0\% R@0.5 while Gemini achieves 40.9\%, indicating near-floor performance on this slice. This contrasts with Black participants, where performance is higher (Gemini: 19.5\% R@0.5), suggesting different performance patterns across subgroups.

\textbf{Implications.} These disparities likely reflect upstream biases in pre-training corpora rather than annotation artifacts. Demographic slice sizes are uneven and can inflate variance for smaller groups. Generation-oriented summarization shows relative stability, while reasoning-heavy MCQ and especially temporal localization expose larger diagnostic performance gaps across demographic groups.

\begin{table*}[t]
\centering
\caption{\textbf{Temporal localization performance across demographic groups (Models 1--4).} Results on all dataset. mIoU (\%), R@0.3/R@0.5 (\%), and rationale quality (Score: Judge score (0--10), RG-L: ROUGE-L (\%), Sim: similarity (0--1)). \textbf{Bold} = best, \worst{highlighted} = worst within each demographic group per column.}
\label{tab:demo_temporal_a}
\scriptsize
\setlength{\tabcolsep}{2pt}
\resizebox{\textwidth}{!}{%
\begin{tabular}{@{}l | c c c c c c c c c c c c c c c c c c c c c c c c@{}}
\toprule
Group
& \multicolumn{6}{c}{Gemini 3.0 Pro$^\dagger$}
& \multicolumn{6}{c}{Qwen3-Omni}
& \multicolumn{6}{c}{UniMoE-2.0}
& \multicolumn{6}{c}{MiniCPM-o-2.6} \\
\cmidrule(lr){2-7}\cmidrule(lr){8-13}\cmidrule(lr){14-19}\cmidrule(lr){20-25}
&
mIoU & R@0.3 & R@0.5 & Score & RG-L & Sim &
mIoU & R@0.3 & R@0.5 & Score & RG-L & Sim &
mIoU & R@0.3 & R@0.5 & Score & RG-L & Sim &
mIoU & R@0.3 & R@0.5 & Score & RG-L & Sim \\
\midrule
\multicolumn{25}{l}{\textit{Race}}\\
Arab       & 23.4 & \worst{32.6} & 21.1 & 5.23 & \worst{26.9} & \worst{0.638} & \worst{2.3} & \worst{2.9} & 1.6 & 2.59 & \worst{27.0} & 0.686 & \worst{0.7} & \worst{0.2} & 0.2 & 2.33 & \worst{22.6} & 0.530 & \high{3.0} & \high{3.6} & \high{2.6} & \worst{3.46} & 18.6 & 0.412 \\
Indigenous & \high{36.7} & \high{55.7} & \high{40.9} & \high{6.41} & 28.0 & \high{0.694} & 4.0 & \high{8.1} & \worst{0.0} & \worst{2.48} & 27.5 & \worst{0.677} & 1.2 & 1.3 & \high{1.3} & \high{2.51} & \high{28.3} & \high{0.594} & \worst{0.0} & \worst{0.0} & \worst{0.0} & \high{4.23} & \high{21.1} & \high{0.484} \\
Asian      & 31.3 & 46.6 & 30.7 & 5.69 & 27.6 & 0.658 & \high{4.5} & 6.1 & \high{2.9} & \high{2.67} & 27.6 & 0.690 & \high{1.6} & 1.8 & 0.6 & 2.26 & 22.8 & \worst{0.521} & 2.1 & 3.0 & 0.8 & 3.60 & \worst{18.1} & \worst{0.377} \\
White      & 24.4 & 35.3 & 23.0 & 5.22 & 28.0 & 0.666 & 3.6 & 5.2 & 2.6 & 2.62 & 27.7 & 0.695 & \high{1.6} & \high{2.0} & 1.2 & 2.10 & 23.9 & 0.557 & 2.0 & 2.5 & 0.9 & 3.56 & 19.3 & 0.405 \\
Hispanic   & 25.1 & 36.6 & 23.8 & \worst{5.19} & 27.6 & 0.669 & 3.4 & 5.4 & 2.3 & 2.58 & 27.1 & 0.684 & \worst{0.7} & 0.8 & \worst{0.1} & \worst{2.08} & 23.7 & 0.566 & 1.9 & 2.5 & 0.2 & 3.65 & 18.6 & 0.399 \\
Black      & \worst{22.4} & 33.5 & \worst{19.5} & 5.20 & \high{28.4} & 0.675 & 3.3 & 4.8 & 1.8 & 2.60 & \high{28.0} & \high{0.696} & 0.9 & 0.8 & 0.6 & 2.15 & 24.4 & 0.566 & 2.1 & 2.7 & 0.3 & 3.77 & 19.4 & 0.416 \\
\midrule
\multicolumn{25}{l}{\textit{Gender}}\\
Male   & \worst{25.1} & \high{37.1} & \worst{23.5} & \worst{5.21} & \worst{27.8} & \worst{0.665} & \high{3.7} & \high{5.3} & \high{2.6} & \worst{2.59} & \high{27.8} & \high{0.695} & \high{1.5} & \high{1.7} & \high{0.9} & \worst{2.14} & \high{23.9} & \high{0.556} & \worst{1.9} & \worst{2.5} & \worst{0.7} & \high{3.62} & \high{19.3} & \high{0.405} \\
Female & \high{25.7} & \worst{37.0} & \high{24.2} & \high{5.42} & \high{28.0} & \high{0.666} & \worst{3.5} & \worst{5.1} & \worst{2.1} & \high{2.66} & \worst{27.5} & \worst{0.689} & \worst{1.2} & \worst{1.2} & \worst{0.7} & \high{2.16} & \worst{23.6} & \worst{0.548} & \high{2.3} & \high{2.9} & \high{0.9} & \worst{3.60} & \worst{18.7} & \worst{0.400} \\
\midrule
\multicolumn{25}{l}{\textit{Age}}\\
18--24 & \high{27.9} & \high{42.8} & \high{27.3} & \high{5.64} & \high{29.4} & \high{0.696} & \high{3.9} & \high{6.5} & 2.4 & 2.57 & \high{29.7} & \high{0.712} & \high{1.8} & 1.8 & \high{1.2} & \high{2.34} & \high{26.5} & \high{0.589} & \high{2.2} & \high{3.6} & \high{1.1} & \high{3.97} & \high{21.0} & \high{0.432} \\
25--39 & 26.1 & 38.2 & 25.2 & 5.35 & \worst{27.7} & 0.668 & \worst{3.4} & \worst{4.7} & \high{2.5} & \worst{2.51} & 27.5 & 0.694 & 1.7 & \high{2.1} & \high{1.2} & 2.15 & 23.8 & 0.555 & \worst{1.9} & \worst{2.4} & \worst{0.7} & 3.65 & \worst{18.8} & 0.404 \\
40+    & \worst{24.2} & \worst{34.9} & \worst{21.9} & \worst{5.20} & 27.8 & \worst{0.658} & 3.7 & 5.4 & \worst{2.3} & \high{2.72} & \worst{27.4} & \worst{0.687} & \worst{0.9} & \worst{0.9} & \worst{0.4} & \worst{2.11} & \worst{23.3} & \worst{0.544} & \high{2.2} & 2.7 & 0.8 & \worst{3.52} & 18.9 & \worst{0.396} \\
\bottomrule
\end{tabular}}
\end{table*}

\begin{table*}[t]
\centering
\caption{\textbf{Temporal localization performance across demographic groups (Models 5--8).} Results on all dataset. mIoU (\%), R@0.3/R@0.5 (\%), and rationale quality (Score: Judge score (0--10), RG-L: ROUGE-L (\%), Sim: similarity (0--1)). \textbf{Bold} = best, \worst{highlighted} = worst within each demographic group per column.}
\label{tab:demo_temporal_b}
\scriptsize
\setlength{\tabcolsep}{2pt}
\resizebox{\textwidth}{!}{%
\begin{tabular}{@{}l | c c c c c c c c c c c c c c c c c c c c c c c c@{}}
\toprule
Group
& \multicolumn{6}{c}{Baichuan Omni 1.5}
& \multicolumn{6}{c}{OLA}
& \multicolumn{6}{c}{VITA 1.5}
& \multicolumn{6}{c}{VideoLLaMA2} \\
\cmidrule(lr){2-7}\cmidrule(lr){8-13}\cmidrule(lr){14-19}\cmidrule(lr){20-25}
&
mIoU & R@0.3 & R@0.5 & Score & RG-L & Sim &
mIoU & R@0.3 & R@0.5 & Score & RG-L & Sim &
mIoU & R@0.3 & R@0.5 & Score & RG-L & Sim &
mIoU & R@0.3 & R@0.5 & Score & RG-L & Sim \\
\midrule
\multicolumn{25}{l}{\textit{Race}}\\
Arab       & \worst{2.0} & 3.0 & \worst{0.3} & \high{2.40} & \high{19.7} & 0.452 & \worst{3.0} & \worst{3.2} & 1.4 & 3.38 & 20.9 & 0.419 & 2.0 & 2.6 & 1.2 & \worst{3.62} & 22.0 & 0.433 & \worst{2.5} & \worst{0.5} & \worst{0.0} & 2.19 & 22.7 & 0.496 \\
Indigenous & \high{6.1} & \high{8.1} & \high{5.8} & \worst{1.69} & 19.2 & \high{0.477} & \high{6.9} & \high{10.3} & \high{9.2} & \worst{2.82} & \high{24.2} & \high{0.503} & \worst{1.2} & \worst{1.3} & 1.3 & \high{4.56} & \high{23.9} & \high{0.501} & \high{4.8} & 1.3 & \high{1.3} & \high{2.46} & 22.0 & 0.516 \\
Asian      & 3.5 & 3.3 & 1.6 & 2.35 & 17.1 & \worst{0.412} & 3.5 & 3.9 & 1.3 & \high{3.81} & \worst{20.6} & \worst{0.405} & \high{2.3} & 3.5 & \high{1.4} & 4.03 & \worst{20.8} & \worst{0.403} & 3.8 & 1.4 & 0.4 & \worst{1.99} & 22.9 & 0.512 \\
White      & 2.7 & 3.1 & 1.3 & 2.22 & 18.1 & 0.431 & 3.2 & 3.5 & \worst{1.2} & 3.42 & 21.7 & 0.441 & 1.9 & 2.9 & \high{1.4} & 3.88 & 22.0 & 0.434 & 3.4 & \high{1.8} & 0.5 & 2.14 & 22.2 & \worst{0.493} \\
Hispanic   & 2.8 & \worst{1.8} & 0.8 & 2.20 & \worst{16.7} & 0.414 & 3.6 & 4.4 & 1.4 & 3.61 & 20.7 & 0.442 & 1.7 & 3.0 & \worst{0.8} & 3.70 & 21.8 & 0.431 & 3.0 & 1.5 & \worst{0.0} & 2.27 & \worst{21.3} & 0.494 \\
Black      & 2.6 & 2.6 & 0.8 & 2.04 & 17.7 & 0.429 & 3.7 & 4.5 & 1.7 & 3.46 & 21.8 & 0.457 & \high{2.3} & \high{4.0} & \high{1.4} & 4.08 & 22.1 & 0.445 & 2.8 & 1.1 & 0.3 & 2.02 & \high{23.2} & \high{0.525} \\
\midrule
\multicolumn{25}{l}{\textit{Gender}}\\
Male   & \worst{2.7} & \worst{2.7} & \worst{1.1} & \worst{2.19} & \high{17.9} & \high{0.430} & \worst{3.2} & \worst{3.5} & \worst{1.3} & \worst{3.49} & \high{21.7} & \high{0.440} & \worst{1.9} & \worst{3.0} & \worst{1.2} & \worst{3.90} & \high{21.9} & \high{0.432} & \worst{3.3} & 1.5 & 0.4 & \worst{2.06} & \high{22.6} & \high{0.507} \\
Female & \high{3.1} & \high{3.4} & \high{1.4} & \high{2.23} & \worst{17.7} & \worst{0.423} & \high{3.7} & \high{4.4} & \high{1.5} & \high{3.52} & \worst{21.1} & \worst{0.434} & \high{2.2} & \high{3.4} & \high{1.4} & \high{3.93} & \worst{21.6} & \worst{0.429} & \high{3.4} & 1.5 & 0.4 & \high{2.19} & \worst{22.3} & \worst{0.497} \\
\midrule
\multicolumn{25}{l}{\textit{Age}}\\
18--24 & \high{3.1} & 3.0 & \high{1.8} & \worst{2.08} & \high{18.6} & \high{0.451} & \high{4.3} & \high{5.1} & \high{2.7} & 3.49 & \high{23.5} & \high{0.467} & \worst{1.6} & \worst{2.0} & \worst{1.1} & \high{4.40} & \high{23.3} & \high{0.455} & \worst{3.1} & \worst{0.8} & \worst{0.3} & \high{2.29} & \high{24.7} & \high{0.545} \\
25--39 & 2.9 & \high{3.3} & 1.3 & 2.15 & \worst{17.6} & 0.429 & \worst{3.3} & \worst{3.7} & 1.4 & \worst{3.44} & 21.4 & 0.438 & 2.0 & 3.2 & 1.2 & 3.96 & \worst{21.5} & 0.432 & 3.2 & 1.2 & \worst{0.3} & 2.15 & 22.4 & 0.503 \\
40+    & \worst{2.7} & \worst{2.7} & \worst{0.9} & \high{2.29} & 17.8 & \worst{0.420} & \worst{3.3} & 3.9 & \worst{1.2} & \high{3.56} & \worst{21.0} & \worst{0.431} & \high{2.2} & \high{3.3} & \high{1.5} & \worst{3.79} & 21.8 & \worst{0.426} & \high{3.4} & \high{1.9} & \high{0.4} & \worst{2.05} & \worst{22.1} & \worst{0.494} \\
\bottomrule
\end{tabular}}
\end{table*}

\subsection{Statistical Significance of Demographic Disparities}
\label{app:mann_whitney}

To validate the disparities in Tables~\ref{tab:demo_summ_a}--\ref{tab:demo_temporal_b}, we conduct Mann-Whitney U tests across all model--task--demographic combinations, using rank-biserial correlation $|r|$ as effect size ($n \geq 100$ per group; meaningful threshold $|r| \geq 0.1$).

\textbf{Race.} The Black--Asian gap in T1 Summarization is the most consistent finding, significant across all models (p\,$<$\,0.001, $|r|$\,=\,0.19--0.36) and confirmed by within-topic confounder control in 5/8 models. White--Black gaps are similarly significant ($|r|$\,=\,0.18--0.26). Most T3 race effects are statistically significant but negligible ($|r|$\,$<$\,0.1), attributable to large sample sizes.

\textbf{Gender.} Female participants outperform Male in T1 Summarization across five models (p\,$<$\,0.001--0.03, $|r|$\,=\,0.09--0.19), with no meaningful gender effects in T2 or T3.

\textbf{Age.} Models consistently favor participants aged 40+ in T1 and T2 (p\,$<$\,0.05, $|r|$\,=\,0.10--0.16); T3 age effects are statistically significant but negligible in magnitude.

\textbf{Within-topic control.} The Asian--Black summarization gap persists within topics for 5/8 models, indicating that the gap persists after within-topic control, although residual confounding cannot be ruled out.

\begin{table}[t]
\centering
\scriptsize
\caption{\textbf{Emotional language analysis.} Analysis of MLLM empathic summary outputs on the all dataset (Task~1: summarization) using LIWC-22 categories. \textbf{Bold} denotes best performance, \underline{underline} denotes second-best. Emo.: emotion; all \% columns reflect the percentage of total words matching each LIWC-22 category; Tone: standardized percentile score (1--100).}
\setlength{\tabcolsep}{3pt}
\resizebox{\linewidth}{!}{%
\begin{tabular}{ccccccc}
\toprule
\textbf{Model} & 
\makecell[c]{\textbf{Pro-}\\\textbf{social}\\(\%)} & 
\makecell[c]{\textbf{Affili-}\\\textbf{ation}\\(\%)} & 
\makecell[c]{\textbf{In-}\\\textbf{sight}\\(\%)} & 
\makecell[c]{\textbf{Total}\\\textbf{Emo.}\\(\%)} & 
\makecell[c]{\textbf{Neg.}\\\textbf{Emo.}\\(\%)} & 
\makecell[c]{\textbf{Tone}\\(pct.)} \\
\midrule
Gemini 3.0 Pro  & 1.82 & 2.68 & 3.07 & \textbf{4.35} & \textbf{2.03} & 46.16 \\
Qwen3-Omni      & 1.97 & \underline{2.78} & 3.00 & \underline{3.88} & \underline{1.39} & 56.99 \\
UniMoE-2.0      & \underline{2.07} & 2.69 & \underline{3.39} & 3.54 & 0.93 & 57.94 \\
MiniCPM-o-2.6   & 1.06 & 1.42 & 2.51 & 1.41 & 0.38 & 46.10 \\
Baichuan-Omni 1.5 & 1.79 & 2.67 & 3.26 & 3.47 & 0.82 & \textbf{60.09} \\
OLA             & \textbf{2.72} & \textbf{3.30} & \textbf{3.84} & 4.02 & 1.15 & \underline{60.00} \\
VITA-1.5        & 1.63 & 2.13 & 3.51 & 2.65 & 0.60 & 55.45 \\
VideoLLaMA2     & 0.62 & 0.93 & 2.16 & 2.02 & 0.49 & 49.83 \\
\bottomrule
\end{tabular}
}
\label{tab:liwc_empathy}
\vspace{-1em}
\end{table}

\section{Emotional Language Analysis}
\label{app:app_empathy}
In Section~\ref{sec:empathy}, we discussed empathic capabilities of models. Table~\ref{tab:liwc_empathy} reports the psycholinguistic profile of model empathic summaries across six LIWC-22 dimensions: \textbf{Prosocial} (helping or caring language such as "support", "help"), \textbf{Affiliation} (references to social connections like "friend", "together"), \textbf{Insight} (cognitive understanding words like "realize", "understand"), \textbf{Total Emotion} (any emotional vocabulary), \textbf{Negative Emotion} (distress-specific terms like "anxious", "worried"), and \textbf{Tone} (overall warmth and positivity on 0-100 percentile scale, where higher values indicate warmer, more informal language).

\subsection{Key Observations}

\textbf{Divergent empathy strategies.} Models adopt fundamentally different empathic approaches. Gemini prioritizes emotional validation (4.35\% Total Emotion, 2.03\% Negative Emotion) but maintains formal tone (46.16, below neutral 50th percentile). OLA leads in Prosocial (2.72\%), Affiliation (3.30\%), and Insight (3.84\%), while Baichuan-Omni 1.5 achieves the warmest Tone (60.09), both reflecting their shared Qwen2.5-7B backbone. Qwen3-Omni falls between these extremes (1.97\% Prosocial, 3.88\% Emotion, 56.99 Tone), demonstrating balanced empathic expression.

\textbf{Cognitive vs. affective empathy.} OLA achieves the highest Insight score (3.84\%), reflecting strong cognitive perspective-taking, yet moderate emotional validation (1.15\% Negative Emotion). VITA-1.5 similarly shows high Insight (3.51\%) but minimal emotional validation (0.60\% Negative Emotion), revealing that some models can articulate understanding of mental states without fully expressing emotional responses.

\textbf{Small model limitations.} Smaller models (MiniCPM-o-2.6, VideoLLaMA2) show consistently weak empathic performance. VideoLLaMA2 produces only 0.62\% Prosocial language and 2.02\% Total Emotion with near-neutral Tone (49.83), reading as detached clinical descriptions. These results show that empathic modulation requires model scale since smaller models default to factual reporting without adapting linguistic style to emotional contexts.

\textbf{Tone-emotion independence.} Tone and emotional vocabulary do not correlate: Gemini's high emotion (4.35\%, 2.03\%) pairs with low Tone (46.16), while Baichuan-Omni 1.5's highest Tone (60.09) pairs with moderate emotion (3.47\%, 0.82\%). This pattern reveals that warmth and emotional validation are orthogonal dimensions. The optimal balance depends on context: medical consultations may benefit from Gemini's approach (serious tone, high emotional validation), while customer service may favor Baichuan or OLA's profile (warm tone, supportive language).

\textbf{Implications.} Current models lack nuanced empathic control. While commercial models demonstrate some capacity for empathic modulation when prompted, they adopt fixed strategies rather than adapting to situational demands. Future work should explore fine-grained empathy prompting that separately controls emotional validation, warmth, and perspective-taking; context-adaptive empathy that adjusts based on interaction type; and empathy-focused training objectives beyond standard language modeling.

\section{Data Release and License}
To support transparency and reproducibility, we release the dataset through a Hugging Face repository at \href{https://huggingface.co/datasets/vector-institute/sonic-o1}{https://huggingface.co/datasets/vector-institute/sonic-o1}. During preprocessing, textual personally identifiable information (PII) was removed from the derived annotations and metadata. We do not introduce or link additional identifying information beyond what is present in the original publicly available source videos. The dataset was reviewed and approved through Vector Institute’s internal review process. It is released under the Vector Institute License, subject to the access and use conditions specified on the dataset page. We also provide the code and scripts used for data processing and experimental reproduction through the associated project repository~\href{https://github.com/VectorInstitute/sonic-o1}{https://github.com/VectorInstitute/sonic-o1}. We encourage the community to responsibly use and build on this resource.

\end{document}